\documentclass[journal]{IEEEtran}

\usepackage{graphicx}
\usepackage[latin1]{inputenc}
\usepackage[noadjust]{cite}
\usepackage[T1]{fontenc}
\usepackage{amssymb, amsmath, subfigure, colortbl, color, tabularx}
\usepackage{setspace}
\usepackage{algorithm,algpseudocode}
\usepackage{hyperref}
\definecolor{darkgreen}{rgb}{0.0, 0.5, 0.0}

\definecolor{redmag}{rgb}{0.8, 0.2, 0.2}
\definecolor{darkmag}{rgb}{0.6, 0.2, 0.2}

\newcommand{\mI}{{\mathbf I}}
\newcommand{\vr}{\mathbf{r}}

\newcommand{\vs}{\mathbf{s}}

\newcommand{\ve}{\mathbf{e}}

\newcommand{\vx}{\mathbf{x}}

\newcommand{\vxh}{\hat{\mathbf{x}}}

\newcommand{\vy}{\mathbf{y}}

\newcommand{\vz}{\mathbf{z}}
\newcommand{\vzt}{\mathbf{z}^{\text{\tiny T}}}

\newcommand{\mS}{\mathbf{S}}
\newcommand{\mSt}{\mathbf{S}^{\text{\tiny T}}}
\newcommand{\mG}{\mathbf{G}}

\newcommand{\mD}{\mathbf{D}}
\newcommand{\mDt}{\mathbf{D}^{\text{\tiny T}}}

\newcommand{\mM}{\mathbf{M}}

\newcommand{\mQ}{\mathbf{Q}}
\newcommand{\mQt}{\mathbf{Q}^{\text{\tiny T}}}

\newcommand{\mH}{\mathbf{H}}
\newcommand{\mHt}{\mathbf{H}^{\text{\tiny T}}}

\newcommand{\Ex}{\operatorname{E}}
\newcommand{\eps}{\boldsymbol{\epsilon}}

\newcommand{\veta}{\boldsymbol{\eta}}

\newcommand{\vd}{\mathbf{d}}

\title{A New Adaptive Video Super-Resolution Algorithm With Improved Robustness to Innovations}
\author{Ricardo~Augusto~Borsoi, Guilherme~Holsbach~Costa,~\IEEEmembership{Member,~IEEE,} Jos\'e~Carlos~Moreira~Bermudez,~~\IEEEmembership{Senior~Member,~IEEE}
\thanks{Manuscript received Month day, year; revised Month day, year and Month day, year. Date of publication Month day, year; date of current version Month day, year.
This work has been supported by the National Council for Scientific and Technological Development (CNPq).
The associate editor coordinating the review of this manuscript and approving ti for publication was Prof. Oleg Michailovich.
A preliminary version of this work has been presented in~\cite{borsoi2017newSRRrobustnessInn}.
\textit{(Corresponding author: Ricardo A. Borsoi.)}}
\thanks{R.A. Borsoi and J.C.M. Bermudez are with the Department
of Electrical Engineering, Federal University of Santa Catarina, Florian\'opolis, SC, Brazil. e-mail: raborsoi@gmail.com, j.bermudez@ieee.org.}% <-this % stops a space
\thanks{G.H. Costa is with University of Caxias do Sul, Caxias do Sul, Brazil. e-mail: holsbach@ieee.org.}% <-this % stops a space
\thanks{This paper has supplementary downloadable material available at \url{http://ieeexplore.ieee.org}, provided by the authors. The material includes more extensive experimental results. Contact raborsoi@gmail.com for further questions about this work. This material is 6.1MB in size.}
\thanks{Color versions of one or more of the figures in this paper are available online at \url{http://ieeexplore.ieee.org}.}
\thanks{Digital Object Identifier.}
}

\date{2017}

\begin{document}

\allowdisplaybreaks
% \cblue{say there is the need of expensive training, and does not performs well if test setting is not similar to trainining. and albeit running cost is low, it is still not comparable to rlms's}

% \cblue{cite deep learning here. it achieves good quality, but depends on the training set and might reach results worse than bicubic interpolation, possibly not robust.}

\maketitle

%-------------------------------------------------------------------------------
%-------------------------------------------------------------------------------
\begin{abstract}
In this paper, a new video super-resolution reconstruction (SRR) method with improved robustness to outliers is proposed.
Although the R-LMS is one of the SRR algorithms with the best reconstruction quality for its computational cost, and is naturally robust to registration inaccuracies, its performance is known to degrade severely in the presence of innovation outliers.
By studying the proximal point cost function representation of the R-LMS iterative equation, a better understanding of its performance under different situations is attained. Using statistical properties of typical innovation outliers, a new cost function is then proposed and two new algorithms are derived, which present improved robustness to outliers while maintaining computational costs comparable to that of R-LMS. Monte Carlo simulation results illustrate that the proposed method outperforms the traditional and regularized versions of LMS, and is competitive with state-of-the-art SRR methods at a much smaller computational cost.
\end{abstract}

%---------------------------------------------------------------
%---------------------------------------------------------------
%---------------------------------------------------------------
\section{Introduction}
\label{sec:intro}
Super-resolution reconstruction (SRR) is a well established approach for digital image quality improvement. SRR consists basically in combining multiple low-resolution (LR) images of the same scene or object in order to obtain one or more images of higher resolution (HR), outperforming physical limitations of image sensors.
Applications for SRR are numerous and cover diverse areas, including the reconstruction of satellite images in remote sensing, surveillance videos in forensics and images from low-cost digital sensors in standard end-user systems. References~\cite{Park03,Nasrollahi14} review several important concepts and initial results on SRR.

SRR algorithms usually belong to one of two groups. Image SRR algorithms, which reconstruct one HR image from multiple observations, and video SRR algorithms, which reconstruct an entire HR image sequence.
Video SRR algorithms often include a temporal regularization that constrains the norm of the changes in the solution between adjacent time instants~\cite{Borman99,Tian05,Tian09,Zibetti07,Belekos10,Su11}.
This introduces information about the correlation between adjacent frames, and tends to ensure video consistency over time, improving the quality of the reconstructed sequences~\cite{Choi96}.

% A typical characteristic of SRR algorithms is the high computational cost. Recent developments in both image and video SRR have been mostly directed towards achieving increased reconstruction quality, using more appropriate \textit{a priori} information about the underlying image and the acquisition process.
% %
% This is the case, for example, of the non-parametric methods based on spatial kernel regression~\cite{Takeda09}, non-local methods~\cred{\cite{Protter09,barzigar2016videoSRRscobep}} and variational Bayesian methods~\cred{\cite{Babacan11,liu2014bayesianVideoSRR}}. \cred{cite deep learning here. say that it achieves good quality, but depends on the training set and might reach results worse than bicubic interpolatoin, possibly not robust.}
% %
% Although these techniques led to considerable improvements in the quality of state of the art SRR algorithms, those improvements certainly do not come for free. The computational cost of these algorithms is extremely high, which makes them unsuitable to real-time SRR applications.
% %
% Moreover, registration errors plague most SRR algorithms. This motivates the use of nonlinear cost functions and more complex methodologies, which further contributes to increase the computational cost of the corresponding algorithms~\cite{Farsiu2004challengesSRR,Song2010SRRhybridL1L2}.

A typical characteristic of SRR algorithms is the high computational cost. Recent developments in both image and video SRR have been mostly directed towards achieving increased reconstruction quality, either using more appropriate \textit{a priori} information about the underlying image and the acquisition process or learning the relationship between LR and HR images from a set of training data.
Examples are the non-parametric methods based on spatial kernel regression~\cite{Takeda09}, non-local methods~\cite{Protter09,barzigar2016videoSRRscobep}, variational Bayesian methods~\cite{Babacan11,liu2014bayesianVideoSRR} and, more recently, deep-learning-based methods~\cite{kappeler2016videoSRRneuralNets,liu2017robustSRRneuralNets,tao2017detailRevealingSRRneuralNets}.

Although these techniques have led to considerable improvements in the quality of state of the art SRR algorithms, such improvements did not come for free. The computational cost of these algorithms is very high, which makes them unsuitable for real-time SRR applications. 
While deep-learning methods can be significantly faster than kernel or non-local methods, they rely on extensive training procedures with large amounts of data. Also, the training must be repeated whenever the test conditions change, or their performance may degrade significantly~\cite{bhowmik2017trainingFreeCNN}.
Moreover, registration errors plague most SRR algorithms. This motivates the use of nonlinear cost functions and more complex methodologies, which further contributes to increase the computational cost of the corresponding algorithms~\cite{Farsiu2004challengesSRR,Song2010SRRhybridL1L2}.

\begin{figure}[thb]
\centering
% \begin{minipage}[b]{.35\linewidth}
%   \centering
%   \centerline{\includegraphics[width=0.9\linewidth]{ex_intro/interpnn_News_frame_210_HR.png}}
%   \centerline{(a)}\smallskip
% \end{minipage}
% %\hfill
% \begin{minipage}[b]{.35\linewidth}
%   \centering
%   \centerline{\includegraphics[width=0.9\linewidth]{ex_intro/interpnn_News_210_bicubic.png}}
%   \centerline{(b)}\smallskip
% \end{minipage}\\
% %
% %
% %
% \begin{minipage}[b]{.35\linewidth}
%   \centering
%   \centerline{\includegraphics[width=0.9\linewidth]{ex_intro/interpnn_News_frame_210_reconstr_RLMS.png}}
%   \centerline{(c)}\smallskip
% \end{minipage}
% %\hfill
% %
% \begin{minipage}[b]{.35\linewidth}
%   \centering
%   \centerline{\includegraphics[width=0.9\linewidth]{ex_intro/interpnn_News_210_reconstr_Babacan11_TV.png}}
%   \centerline{(d)}\smallskip
% \end{minipage}

\begin{minipage}[b]{.413\linewidth}
  \centering
  \centerline{\includegraphics[width=0.9\linewidth]{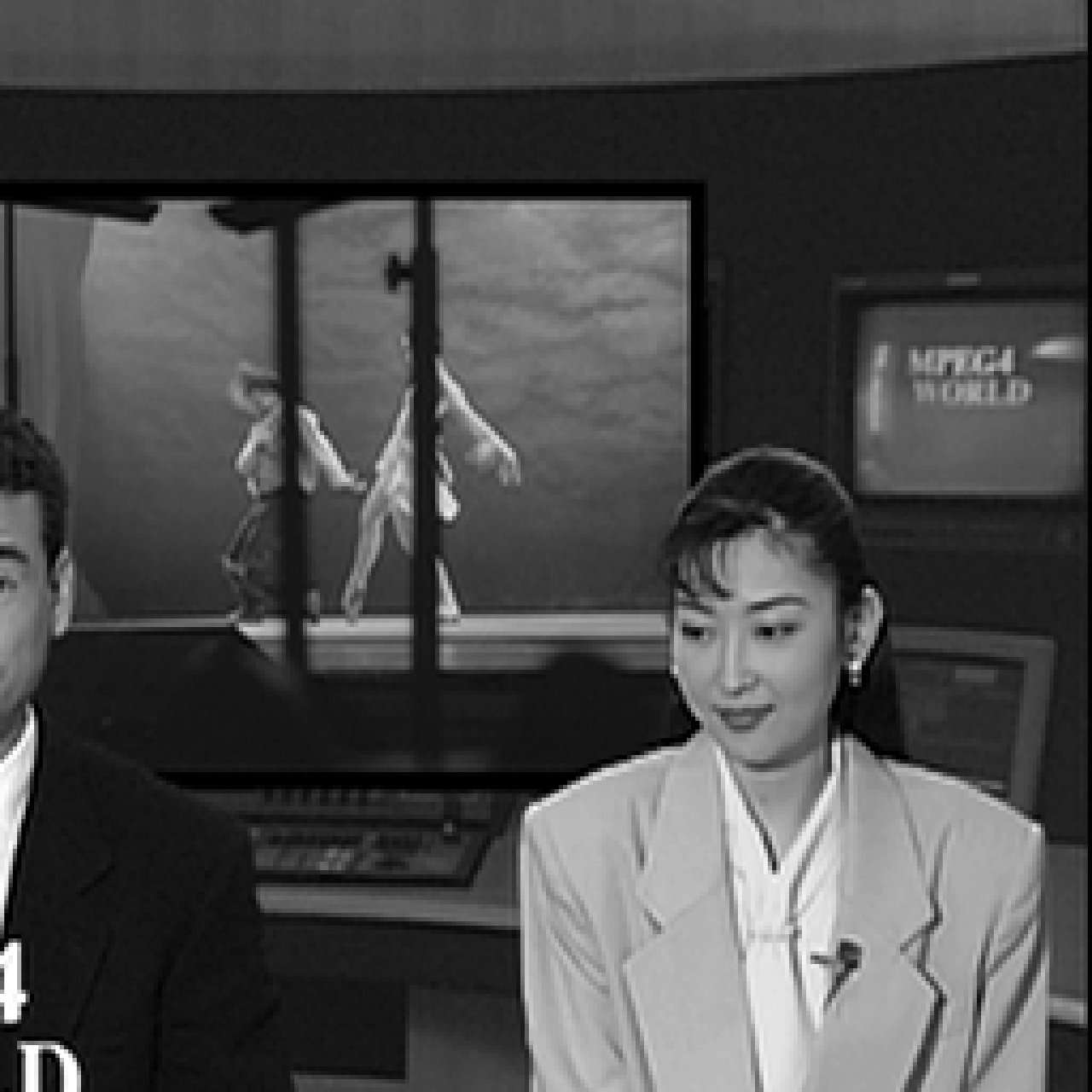}}
  \centerline{(a)}\smallskip
\end{minipage}
%\hfill
\begin{minipage}[b]{.413\linewidth}
  \centering
  \centerline{\includegraphics[width=0.9\linewidth]{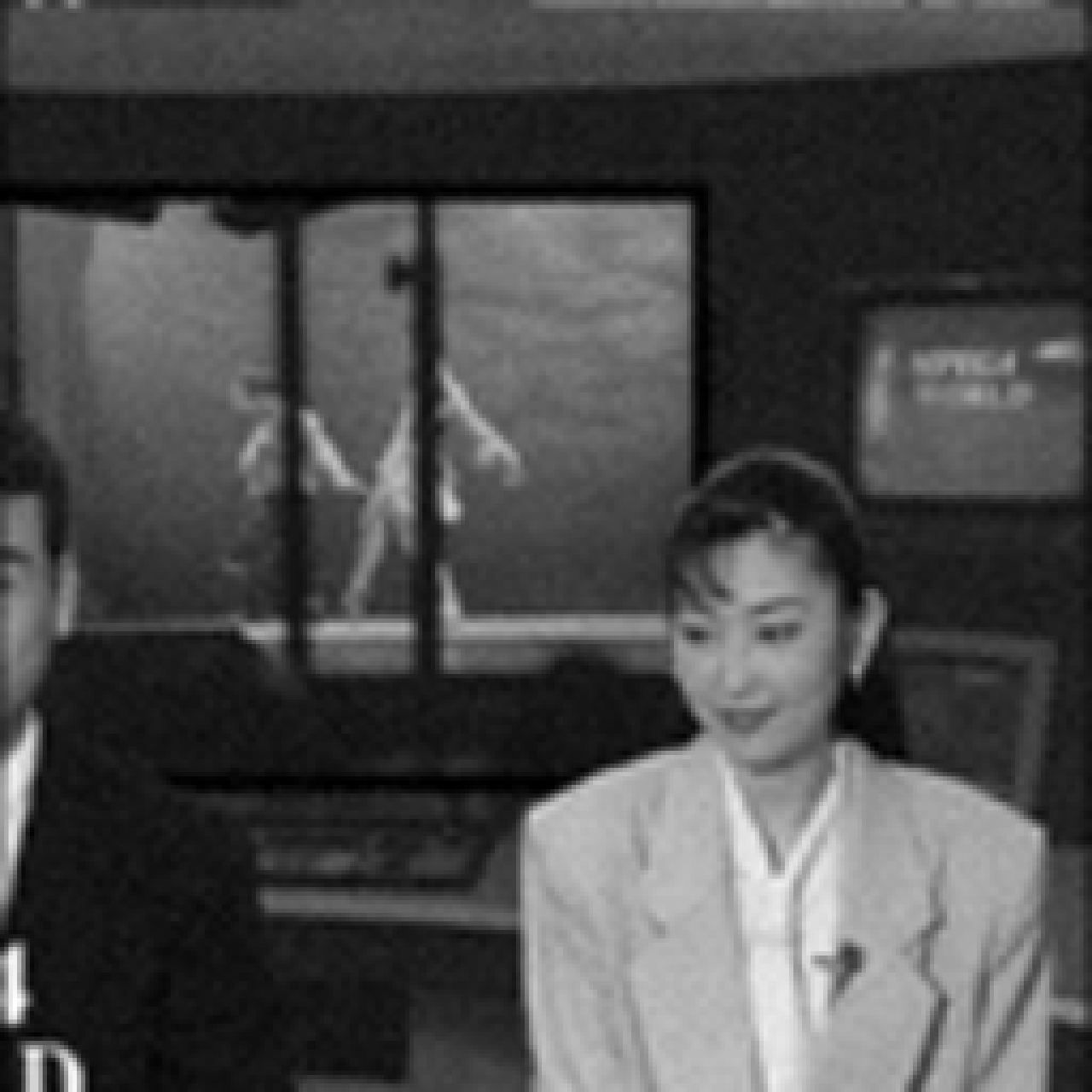}}
  \centerline{(b)}\smallskip
\end{minipage}\\
\begin{minipage}[b]{.413\linewidth}
  \centering
  \centerline{\includegraphics[width=0.9\linewidth]{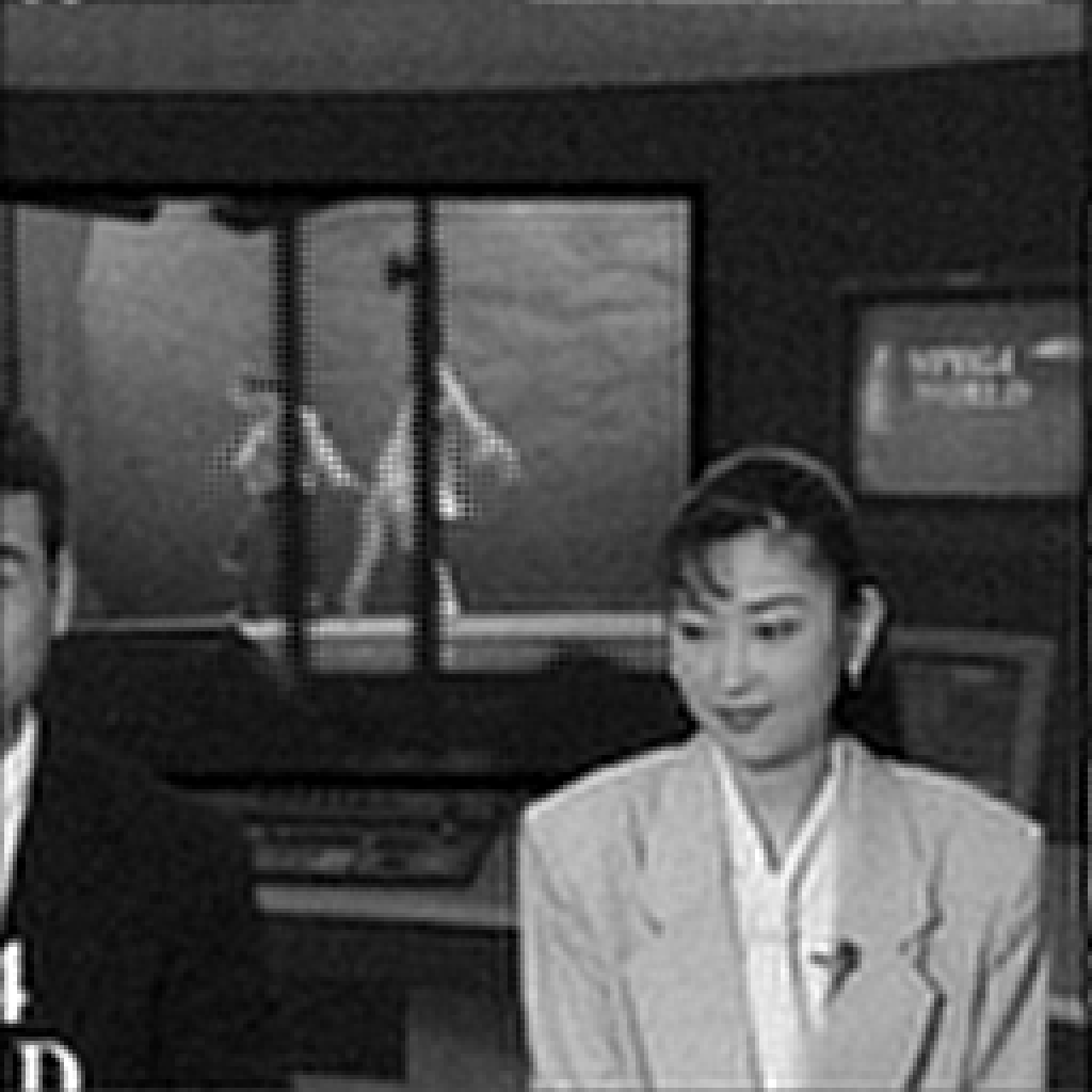}}
  \centerline{(c)}\smallskip
\end{minipage}
%\hfill
%
\begin{minipage}[b]{.413\linewidth}
  \centering
  \centerline{\includegraphics[width=0.9\linewidth]{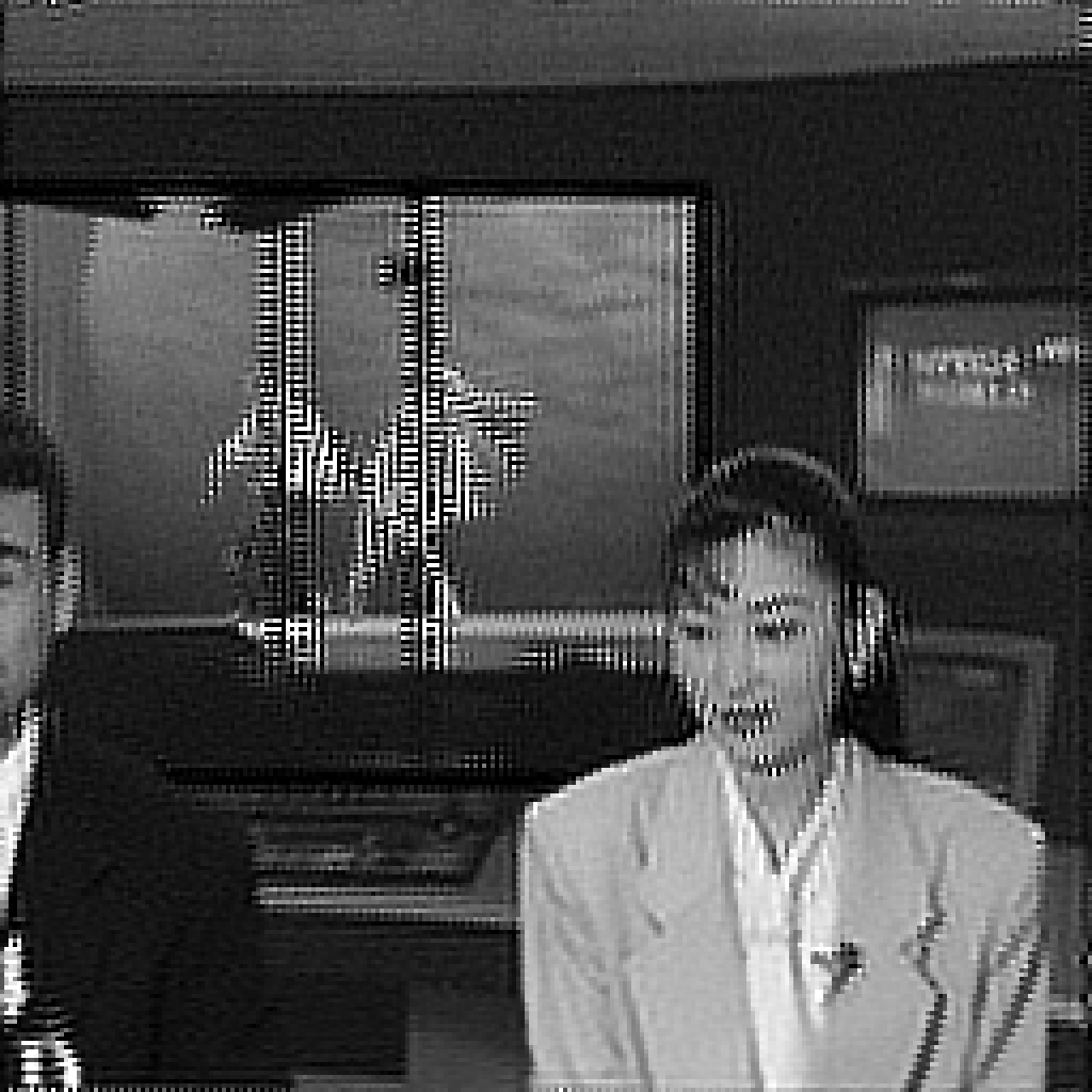}}
  \centerline{(d)}\smallskip
\end{minipage}

% \begin{figure}[thb]
% \begin{minipage}[b]{.475\linewidth}
%   \centering
%   \centerline{\includegraphics[width=0.9\linewidth]{ex_intro/interpnn_News_frame_210_HR.png}}
%   \centerline{(a)}\medskip
% \end{minipage}
% %\hfill
% \begin{minipage}[b]{.475\linewidth}
%   \centering
%   \centerline{\includegraphics[width=0.9\linewidth]{ex_intro/interpnn_News_210_bicubic.png}}
%   \centerline{(b)}\medskip
% \end{minipage}\\
% %
% %
% %
% \begin{minipage}[b]{.475\linewidth}
%   \centering
%   \centerline{\includegraphics[width=0.9\linewidth]{ex_intro/interpnn_News_frame_210_reconstr_RLMS.png}}
%   \centerline{(c)}\medskip
% \end{minipage}
% %\hfill
% %
% \begin{minipage}[b]{.475\linewidth}
%   \centering
%   \centerline{\includegraphics[width=0.9\linewidth]{ex_intro/interpnn_News_210_reconstr_Babacan11_TV.png}}
%   \centerline{(d)}\medskip
% \end{minipage}
%
%
\vspace{-0.35cm}
\caption{Results of the 210th frame for the super-resolution of the \textit{News} sequence by a factor of $2$. (a) Original image (HR). (b) Bicubic spline interpolation. (c) R-LMS using a global translational motion model. (d) SR method of~\cite{Babacan11} using a window of 11 frames.}
\label{fig:example_intro}
\end{figure}

While these traditional SRR methods achieve good reconstruction results~\cite{Ng2003mathematicalSRR}, real-time video SRR applications require simple algorithms. This limitation prompted a significant interest towards developing low complexity SRR algorithms.
The regularized least mean squares (R-LMS)~\cite{Elad99,Elad99pami} is one notable example among the simpler SRR algorithms.
Even though other low-complexity algorithms have been proposed for global translational image motions, such as the $\text{L}_1$ norm estimator in \cite{Farsiu04} and the adaptive Wiener filter of \cite{Hardie07}, their computational complexity is still not competitive with that of the R-LMS.
The R-LMS presents a reasonable reconstruction quality and follows a systematic mathematical derivation. This enables a formal characterization of its behavior~\cite{Costa07} and the specification of well defined design methodologies~\cite{Costa08}.
Furthermore, the R-LMS is also naturally robust to registration errors, which were shown to have a regularizing effect in the algorithm~\cite{Costa09regErrors}.
This makes the quality of the R-LMS algorithm in practical situations to be competitive even with that of costly and elaborated algorithms like~\cite{Babacan11}, as illustrated through an example in Figure~\ref{fig:example_intro}.

% A typical characteristic of SRR algorithms is the high computational cost. While traditional SRR methods achieve good reconstruction results~\cite{Ng2003mathematicalSRR}, real-time video SRR applications require simple algorithms.
%
% This limitation prompted a significant interest towards developing low complexity SRR algorithms. Examples are the implementation of the $\text{L}_1$ norm estimator in \cite{Farsiu04}, and the adaptive Wiener filter of \cite{Hardie07} for global translational image motions.
%
% The least mean squares (R-LMS)~\cite{Elad99,Elad99pami} is one notable example among the simpler SRR algorithms since it presents a reasonable reconstruction quality and follows a systematic mathematical derivation. This enables a formal characterization of its behavior~\cite{Costa07} and the specification of well defined design methodologies~\cite{Costa08}.

Unfortunately, the performances of these simple algorithms, including the R-LMS, tend to be heavily affected by the presence of outliers such as large innovations caused by moving objects and sudden scene changes.
This can lead to reconstructed sequences of worse quality than that of the observations themselves~\cite{Zibetti07}.
Common strategies for obtaining robust algorithms often involve the optimization of nonlinear cost functions~\cite{Borman99,Farsiu04,Zibetti07}, and thus present a computational cost that is not comparable to that of algorithms like the R-LMS.
Interpolation algorithms might seem to be a reasonable option, as their performance is not affected by motion related outliers. However, they do not offer a quality improvement comparable to SRR methods~\cite{blu2001moms,Zibetti07}. Therefore, it is of interest to develop video SRR algorithms that combine good quality, robustness to outliers and a low computational cost.

This paper proposes a new adaptive video SRR algorithm with improved robustness to outliers when compared to the R-LMS algorithm. The contributions of this paper include:
1) A new interpretation of the R-LMS update equation as the proximal regularization of the associated cost function, linearized about the previous estimate, which leads to a better understanding of its quality performance and robustness in different situations.
Using this representation we show that the slow convergence rate of the R-LMS algorithm (typical of stochastic gradient algorithms) establishes a trade-off between its robustness to outliers and the achievable reconstruction quality.
2) A simple model for the statistical properties of typical innovation outliers in natural image sequences is developed, which points to the desirable properties of the proposed technique.
3) A new cost function is then proposed to address the identified problems and two new adaptive algorithms are derived called \textit{Temporally Selective Regularized LMS} (TSR-LMS) and \textit{Linearized Selective Regularized LMS} (LTSR-LMS), which present improved robustness and similar quality at a computational cost comparable to that of the R-LMS algorithm.

% \cblue{[OLD PARAGRAPH:
% This paper proposes a new adaptive video SRR algorithm with improved robustness to outliers when compared to the R-LMS algorithm. The R-LMS update equation is interpreted as proximal regularization of the associated cost function, linearized about the previous estimate. This interpretation leads to a better understanding of its quality performance and robustness in different situations.
% %
% We show that the slow convergence rate of the R-LMS algorithm (typical of stochastic gradient algorithms) establishes a trade-off between its robustness to outliers and the achievable reconstruction quality. Then, using statistical properties of typical innovation outliers in natural image sequences we propose a new cost function that addresses the identified problems.
% %
% Two new adaptive algorithms are derived which present improved robustness and similar quality at a computational cost \cgreen{comparable} to that of the R-LMS algorithm.
% ]}

% ==================================================

The paper is organized as follows. In Section~\ref{sec:notation}, the image acquisition model is defined. In Section~\ref{sec:LMS}, the R-LMS algorithm~\cite{Elad99,Elad99pami} is derived as a stochastic gradient solution to the image estimation problem~\cite{Costa09regErrors}.
In Section~\ref{sec:rlms_with_outliers}, an intuitive interpretation of the R-LMS behavior is presented using the proximal-point cost function representation of the gradient descent iterative equation.
In Section~\ref{sec:proposed}, a new robust cost function is proposed based on a statistical model for the innovations, and two adaptive algorithms are derived.
In chapter~\ref{sec:results}, computer simulations are performed {to assess} the performance of the algorithms. The conclusions are presented in Section~\ref{sec:concl}.

\section{Image Acquisition Model}
\label{sec:notation}

Given the $N \times N$ matrix representation of an LR (observed) digital image $\mathbf{Y}(t)$, and an $M \times M$ ($M > N$) matrix representation of the original HR digital image $\mathbf{X}(t)$, the acquisition process can be modeled as \cite{Park03}
\begin{equation} \label{eq:aquis}
    \vy(t) = \mD\mH\vx(t) + \ve(t)
\end{equation}
where vectors $\vy(t)$ ($N^2 \times 1$) and $\vx(t)$ ($M^2 \times 1$) are the lexicographic representations of the degraded and original images, respectively, at discrete time instant $t$. $\mD$ is an $N^2 \times M^2$ decimation matrix and models the sub-sampling taking place in the sensor. $\mH$ is an $M^2 \times M^2 $ matrix, assumed known, that models the blurring in the acquisition process\footnote{Since $\mH$ is assumed to be estimated independently from the SRR process, the extension of the results in this work to the case of a time-varying $\mH(t)$ is straightforward if an online estimation algorithm for $\mH$ is employed. Here, a possible dependence of $\mH$ on $t$ is omitted for notation simplicity.}. The $N^2 \times 1$ vector $\ve(t)$ models the observation (electronic) noise, whose properties are assumed to be determined from camera tests. The dynamics of the input signal is modeled by~\cite{Elad99}
\begin{equation}\label{eq:dinam}
    \vx(t) = \mG(t) \vx(t-1) + \vs(t)
\end{equation}
where $\mG(t)$ is the warp matrix that describes the relative displacement from $\vx(t-1)$ to $\vx(t)$. Vector $\vs(t)$ models the innovations in $\vx(t)$.

%---------------------------------------------------------------
%---------------------------------------------------------------
%---------------------------------------------------------------
\section{The R-LMS SRR algorithm}
\label{sec:LMS}
Several SRR solutions are based on the minimization of the estimation error (see \cite{Park03} and references therein)
\begin{equation} \label{eq:epsilon}
\eps(t)= \vy(t)-\mD\mH \vxh(t)
\end{equation}
where $\vxh(t)$ is the estimated HR image, and $\eps(t)$ can be interpreted as the estimate of $\ve(t)$ in~\eqref{eq:aquis}. The {LMS} algorithm attempts to minimize the mean-square value of the $\text{L}_2$ norm of~\eqref{eq:epsilon} conditioned on the estimate $\vxh(t)$~\cite{Elad99,Costa07}. Thus, it minimizes the cost function $\mathbf{J}_{\text{MS}}(t) = \Ex\{ \| \eps(t) \|^{2} \, | \, \vxh(t) \}$.

Since natural images are known to be intrinsically smooth, this \textit{a priori} knowledge can be added to the estimation problem in the form of a regularization to the LMS algorithm by constraining the solution that minimizes $\mathbf{J}_{\text{MS}}(t)$. The R-LMS algorithm~\cite{Costa09regErrors,Elad99pami} {is the stochastic gradient version of the gradient descent search for} the solution to the following optimization problem
\begin{equation} \label{eq:RLMS_Lagrangian}
    \mathcal{L}_{\text{R-MS}}(t) = \Ex\{ \| \vy(t)-\mD\mH \vxh(t) \|^{2} \, | \, \vxh(t) \}
						+ \alpha \|\mS\vxh(t)\|^2
\end{equation}
where $\mS$ is the Laplacian operator~\cite[p. 182]{Gonzalez02image}. Note that the performance surface in~\eqref{eq:RLMS_Lagrangian} is defined for each time instant $t$, and the expectation is taken over the ensemble.

Following the steepest descent method, the HR image estimate is updated in the negative direction of the gradient
\begin{align} \label{eq:grad_sup_des}
     \! \! \! \nabla \! \mathcal{L}_{\text{R-MS}}(t) \! = \! - 2 \mHt\mDt \! \{ \Ex[ \vy(t) ] \! - \! \mD\mH \vxh(t) \}
    + 2\alpha \mSt\mS\vxh(t) 
\end{align}
and thus the iterative update of $\vxh(t)$ for a fixed value of $t$ is given by
\begin{align} \label{eq:grad_iterative_eq}
    \vxh_{k+1}(t) = \vxh_{k}(t) - \frac{\mu}{2} \nabla \mathcal{L}_{\text{R-MS}}(t), \quad k=0, 1, \dots, K-1
\end{align}
where $K \in \mathbb{Z^+}$ is the number of iterations of the algorithm, and $\mu$ is the step size used to control the convergence speed. The factor 1/2 is just a convenient scaling.

Using the instantaneous estimate of~\eqref{eq:grad_sup_des} in~\eqref{eq:grad_iterative_eq} yields
\begin{align} 
    \vxh_{k+1}(t) {}={} & \vxh_{k}(t) + \mu
    \mHt\mDt [\vy(t) - \mD\mH
    \vxh_{k}(t)] %-\alpha\mu\mSt\mS\vxh_k(t)\, , 
    \nonumber\\ & \quad
    -\alpha\mu \, \mSt\mS\vxh_k(t)\, \text{,} \,\,\quad k=0, 1, \dots, K-1
    \label{eq:lms}
\end{align}
which is the R-LMS update equation for a fixed value of $t$. The time update of~\eqref{eq:lms} is based on the signal dynamics~\eqref{eq:dinam}, and is performed by the following expression~\cite{Elad99}:
\begin{align} \label{eq:lms_time_update}
    \vxh_{0}(t+1) = \mG(t+1) \vxh_{K}(t) 
    \, \text{.}
\end{align}
Between two time updates,~\eqref{eq:lms} is iterated for $k=0,\ldots,K-1$. The estimate $\vxh(t)$ at a given time instant $t$ is then given by $\vxh(t)=\vxh_K(t)$.

% ------------------------------------------------------------------
% ------------------------------------------------------------------
% ------------------------------------------------------------------
% ------------------------------------------------------------------
\section{R-LMS Performance in the Presence of Outliers}
\label{sec:rlms_with_outliers}

The R-LMS algorithm is computationally efficient when implemented with few stochastic gradient iterations (small $K$) per time instant $t$. {However, the R-LMS algorithm is derived under the assumption that the solution $\vx(t)$ is only slightly perturbed between time instants, a characteristic known as the minimum disturbance principle. This assumption is satisfied in the R-LMS algorithm close to steady-state, when the estimate $\vxh(t)$ has already achieved a reasonable quality (i.e. $\vxh(t)\simeq\vx(t)$). Then, the initialization $\vxh_{0}(t+1)$ determined by~\eqref{eq:lms_time_update} for the next time instant will be already relatively close to the optimal solution, what explains the good steady-state performance of the algorithm even with a small value of $K$ in \eqref{eq:lms}.} 

{The situation is significantly different in the occurrence of innovation outliers. Experience with the R-LMS algorithm shows that the slow convergence of \eqref{eq:lms} as a function of $k$ tends to degrade the quality of the super-resolved images in the presence of innovation outliers. Visible artifacts tend to be created, and the reconstructed images may end up being of inferior quality even when compared to the originally observed LR images. This significantly compromises the quality that can be achieved in real-time super-resolution of video sequences, not just using R-LMS but most of the existing low-complexity algorithms~\cite{Zibetti07}. Super-resolution algorithms designed to be robust under the influence of innovations tend to impose a high computational cost, making them unsuitable for real time applications~\cite{Farsiu04,Costa08}. In the following we examine the R-LMS recursion under a new light, what leads to a mathematically motivated explanation for its lack of robustness to outliers. This explanation will then motivate the proposition of more robust stochastic video SRR algorithms.}

% \cred{In the presence of innovation outliers The lack of robustness to outliers occurs }

% \cred{This limitation compromises the performance of the R-LMS algorithm in the occurrence of outliers\footnote{\cred{The same limitation applies to most low-complexity SRR algorithms.}}. The main reason for this property is that} the R-LMS algorithm is derived under the assumption that the solution $\vx(t)$ is only slightly perturbed between time instants. When the estimate $\vxh(t)$ has already achieved a reasonable quality (i.e. $\vxh(t)\simeq\vx(t)$), the initialization for the next time instant performed according to~\eqref{eq:lms_time_update} will already be relatively close to the optimal solution, what explains the good steady-state performance of the 

% \cred{However, the slow convergence of the R-LMS is known to degrade the quality of the super-resolved images in the presence of innovation outliers. Visible artifacts tend to be created, and the reconstructed images may end up being of quality inferior to that of the originally observed LR images. This significantly compromises the quality that can be achieved with real-time super-resolution of video sequences, not just for R-LMS} but for most existing algorithms~\cite{Zibetti07}. On the other hand, Super-resolution algorithms devised to be robust under the influence of innovations exhibit an increased computational cost, \cred{making them unsuitable} for real time applications~\cite{Farsiu04,Costa08}.

An interesting interpretation of the R-LMS algorithm is possible if we view each iteration of the gradient algorithm \eqref{eq:grad_iterative_eq} (for a fixed value of $t$) as a proximal regularization of the cost function $\mathcal{L}_{\text{R-MS}}(t)$ linearized about the estimation of the previous iteration $\vxh_{k}(t)$.  Proceeding as in \cite[Section~2.2]{Beck09} or \cite[p.~546]{Bertsekas1999}, the gradient iteration \eqref{eq:grad_iterative_eq} can be written as
\begin{align} 
    &\vxh_{k+1}(t) = \underset{\vz}{\arg\min} \Big\{\mathcal{L}_{\tiny{\text{R-MS}}}(\vxh_k(t))
    \nonumber\\ 
    &\hspace{3ex}+ \big(\vz - \vxh_k(t)\big)^{\top}\nabla \mathcal{L}_{\tiny{\text{R-MS}}}(\vxh_k(t)) + \frac{1}{\mu}\left\Vert\vz - \vxh_k(t)\right\Vert^2 \Big\} \label{eq:grad_opt_rlms}
\end{align}
    %
%    =&\, \cblue{\underset{\vz}{\arg\min} 
%    \Big\{\vzt\Big( 2\alpha\mSt\mS\vxh_k(t) 
%    - 2\mHt\mDt\!\big\{\Ex[\vy(t)]\!-\!\mD\mH\vxh_k(t)\big\} \Big)}
%    \nonumber\\ &
%    \cblue{+ \frac{1}{\mu}\left\Vert\vz - \vxh_k(t)\right\Vert^2\Big\}}
    % \, \text{,}
    %
    
\noindent which, using the previous expressions, yields    
\begin{align}   
    \vxh_{k+1}(t) = &\, \underset{\vz}{\arg\min} 
    \Big\{2 \alpha \vzt\mSt\mS\vxh_k(t) 
    - 2 \vzt \mHt\mDt \Ex[\eps_k(t)]
    \nonumber\\ &
    + \frac{1}{\mu}\left\Vert\vz - \vxh_k(t)\right\Vert^2\Big\}
    \label{eq:grad_opt_rlms2}
\end{align}
where $\Ex[\eps_k(t)]$ is the expected value of the observation error~\eqref{eq:epsilon} conditioned on $\vxh(t)=\vxh_k(t)$. 
% \cblue{This equivalence is verified as follows. Differentiating the expression within the external brackets in~\eqref{eq:grad_opt_rlms} with respect to $\vz$ and setting it to zero yields
% %
% \begin{align}
%     \nabla \mathcal{L}_{\tiny{\text{R-MS}}}(\vxh_k(t))+\frac{2}{\mu}(\vz-\vxh_k(t)) = \textbf{0}
% \end{align}
% which solved for $\vz= \vxh_{k+1}(t)$ leads to
% \begin{align}
%     \vxh_{k+1}(t) = \vxh_k(t) - \frac{\mu}{2} \nabla \mathcal{L}_{\tiny{\text{R-MS}}}(\vxh_k(t))
% \end{align}
% which is the R-LMS gradient descent update equation~\eqref{eq:grad_iterative_eq}.}
{This equivalence can be verified by differentiating the expression in the curl brackets in \eqref{eq:grad_opt_rlms} with respect to $\vz$, setting it equal to zero and solving for $\vz= \vxh_{k+1}(t)$.}

Now, the presence of the squared norm within the {curl brackets in~\eqref{eq:grad_opt_rlms} and ~\eqref{eq:grad_opt_rlms2} means} that the optimization algorithm seeks $\vxh_{k+1}(t)$ that minimizes the perturbation $\vxh_{k+1}(t)-\vxh_k(t)$ at each iteration. Evidencing this property leads to a more detailed understanding of the dynamical behavior of the algorithm, its robustness properties and the reconstruction quality it provides. For instance, this constraint on the perturbation of the solution explains how the algorithm tends to preserve details in $\vxh(t)$ that have been estimated in previous time iterations and that are present in $\vxh(t-1)$. However, this same term also opposes changes from $\vxh_k(t)$ to $\vxh_{k+1}(t)$, slowing down the reduction of the observation error from $\eps_k(t)$ to $\eps_{k+1}(t)$ since changes in $\eps_k(t)$ require changes in $\vxh_k(t)$. Therefore, this algorithm cannot simultaneously achieve a fast convergence rate and preserve the super-resolved details for the practically important case of a small number of iterations per time instant (small $K$). The time sequence of reconstructed images will either converge fast and yield low temporal correlation between time estimates (leading to a solution that approaches an interpolation of $\vy(t)$), or will converge slowly and yield a highly correlated image sequence with better quality in the absence of innovation outliers. The occurrence of outliers will result in a significant deviation from the desired signal.

To illustrate this behavior, consider for instance that the reconstructed image sequence at time instant $t-1$ is reasonably close to the real (desired) sequence, i.e. $\vxh(t-1)\simeq\vx(t-1)$. If we consider the video sequence to be only slightly perturbed at the next time instant such that $\|\vs(t)\|\approx 0$ in \eqref{eq:dinam}, the first iteration of \eqref{eq:grad_opt_rlms2} ($k=0$ at time $t$) is given by
\begin{align}
    \vxh_{1}(t) {}={} \underset{\vz}{\arg\min} 
    &\Big\{2 \alpha \vzt\mSt\mS\vxh_0(t) 
    \nonumber\\
   &- 2 \vzt \mHt\mDt \eps_0(t) + \frac{1}{\mu}\left\Vert\vz - \vxh_0(t)\right\Vert^2\Big\}
\end{align}    
which, using \eqref{eq:lms_time_update} with $\vxh(t-1)\simeq\vx(t-1)$, yields 
\begin{align}
    \vxh_{1}(t){}\approx{} \underset{\vz}{\arg\min} 
    \Big\{&2 \alpha \vzt\mSt\mS\vxh_0(t) 
    - 2 \vzt \mHt\mDt \eps_0(t)
    \nonumber\\
    &+ \frac{1}{\mu}\left\Vert\vz - \mG(t)\vx(t-1)\right\Vert^2\Big\}.
    \label{eq:r_lms_srr_proximal_approx_partial}
\end{align}
Using \eqref{eq:dinam} in \eqref{eq:r_lms_srr_proximal_approx_partial},
\begin{align}    
    \vxh_{1}(t){}\approx{} \underset{\vz}{\arg\min} 
    \Big\{&2 \alpha \vzt\mSt\mS\vxh_0(t) 
    - 2 \vzt \mHt\mDt \eps_0(t) 
    \nonumber\\
    &+ \frac{1}{\mu}\left\Vert\vz - \vx(t) + \vs(t)\right\Vert^2\Big\}.
\end{align}
Finally, assuming $\|\vs(t)\|\approx0$ (no outlier in $\vx(t)$)
\begin{align}    
    \vxh_{1}(t){}\approx{} \underset{\vz}{\arg\min} 
    \Big\{&2 \alpha \vzt\mSt\mS\vxh_0(t) 
    - 2 \vzt \mHt\mDt \eps_0(t)
    \nonumber\\
    &+ \frac{1}{\mu}\left\Vert\vz - \vx(t)\right\Vert^2\Big\}.
    \label{eq:r_lms_srr_proximal_approx_noinovations}
\end{align}
{Now, using $\|\vs(t)\|\approx0$, $\vxh(t-1)\simeq\vx(t-1)$, \eqref{eq:aquis} and \eqref{eq:epsilon}, the norm of $\eps_0(t)$ can be approximated by}
\begin{align}
    \|\eps_0(t)\| \simeq & \|\mD\mH\vx(t)+\ve(t) \,-\, \mD\mH\big(\vx(t)-\vs(t)\big)\|
    \nonumber\\	
    \approx & \| \ve(t) \|
\end{align}
which is small since the energy of the observation noise is much smaller than that of registration errors and outliers in most practical applications~\cite{Park03,Costa08}. The first term in the r.h.s. of~\eqref{eq:r_lms_srr_proximal_approx_noinovations} is due to the regularization an promotes the smoothness of the solution. Hence, $\alpha$ should be small to avoid compromising the estimation of the details of $\vx(t)$. The last term will promote a solution close to $\vx(t)$, especially for small values of $\mu$. Then, for reasonably small values of $\alpha$ and $\mu$, $\|\vs(t)\|\approx0$ and $\vxh(t-1)\simeq\vx(t-1)$, the first and second terms in~\eqref{eq:r_lms_srr_proximal_approx_noinovations} can be neglected (i.e. $|2\alpha \vzt\mSt\mS\vxh_0(t) - 2\vzt\mHt\mDt\eps_0(t)| \ll \frac{1}{\mu}\left\Vert\vz - \vx(t)\right\Vert^2$). Then, the solution of the optimization problem \eqref{eq:r_lms_srr_proximal_approx_noinovations} will converge to a vector $\vxh_1(t) \approx \vx(t)$. The same reasoning can be extended to the remaining iterations for $k=2, \ldots, K-1$, which shows that, for $\|\vs(t)\| \approx 0$, the algorithm will lead to a reconstructed image of good quality $\vxh_K(t)\simeq\vx(t)$. This explains how the R-LMS algorithm preserves the reconstructed content in time and extracts information from the different observations, attaining good reconstruction results for well behaved sequences, i.e. in the absence of large innovations.

% The solution $\vxh_1(t)$ will only be slightly perturbed from the initialization $\vxh_0(t)$ due to the first term in~\eqref{eq:r_lms_srr_proximal_approx_noinovations}. Hence, the dominance of the term $\frac{1}{\mu}\left\Vert\vz - \vx(t)\right\Vert^2$ will lead to a solution $\vxh_{1}(t) \approx \vx(t)$. 
% %

Now, let's consider the presence of a significant innovation outlier at time $t$, while still assuming a good reconstruction result at time $t-1$ (i.e. $\vxh(t-1)\simeq\vx(t-1)$). Due to the outlier at time instant $t$, $\vs(t)$ in~\eqref{eq:dinam} will have a significant energy. Then, repeating~\eqref{eq:r_lms_srr_proximal_approx_noinovations} without the assumption $\|\vs(t)\|\approx 0$ yields
\begin{align}
    \vxh_{1}(t) {}\approx{} \underset{\vz}{\arg\min} \,\,
    \Big\{&2 \alpha \vzt\mSt\mS\vxh_0(t) 
    - 2 \vzt \mHt\mDt \eps_0(t)
    \nonumber \\%\displaybreak\\
    &+ \frac{1}{\mu}\left\Vert\vz - \big(\vx(t) - \vs(t)\big)\right\Vert^2\Big\}
    \label{eq:r_lms_srr_proximal_approx_largeinovations}
\end{align}
% \begin{align}
%     \vxh_{1}(t) {}\approx{} \underset{\vz}{\arg\min} \,\,
%     \Big\{&2 \alpha \vzt\mSt\mS\vxh_0(t) 
%     - 2 \vzt \mHt\mDt \eps_0(t)
%     \nonumber\\
%     &+ \frac{1}{\mu}\left\Vert\vz - \big(\vx(t) - \vs(t)\big)\right\Vert^2\Big\}
%     \label{eq:r_lms_srr_proximal_approx_largeinovations}
% \end{align}

\noindent where now the observation error is given by
\begin{align} \label{eq:MagEps0}
    \|\eps_0(t)\| \simeq & \|\mD\mH\vx(t)+\ve(t) \,-\, \mD\mH\big(\vx(t)-\vs(t)\big)\|
    \nonumber\\
    = & \|\ve(t) \,+\, \mD\mH\vs(t)\|.
\end{align}

This result clearly shows that, for large $\| \vs(t) \|$, the solution $\vxh_1(t)$ of \eqref{eq:r_lms_srr_proximal_approx_largeinovations} can be considerably away from $\vx(t)$, as it should contain information introduced by $\vs(t)$. For the desirable case of small $K$, this estimation error will hardly be significantly reduced from $\vxh_0(t)$ to $\vxh_K(t)$ due to the slow convergence of the gradient based recursion. This explains the poor transient performance of the algorithm in the presence of outliers.

% For a fast convergence of the algorithm for a fixed value of $t$ and $k=1,\ldots,K$, meaning that one could choose $K$ small, the cost function should allow for a considerable change of the estimate towards $\vx(t)$ when going from $\vxh_l(t)$ to $\vxh_{l+1}(t)$, $l=1,\ldots,K-1$. We discuss the case $l=0$ (first iteration) and then extend the conclusions to other values of $l$. Now, for large values of $\vs(t)$, $\|\eps_0(t)\|$ in~\eqref{eq:MagEps0} will be large and dominated by the term $\mD\mH\vs(t)$. Moreover, for values of $\alpha$ and $\mu$ typically chosen for an outlier-free situation, the term $\frac{1}{\mu}\|\vz-\big(\vx(t)-\vs(t)\big)\|^2$ will be very large if $\vz\simeq\vx(t)$. Hence, the estimation $\vxh_1(t)$, while being still close to the initialization (which does not contain the outlier $\vs(t)$), will be far from the desired solution $\vx(t)$ (which should contain $\vs(t)$). This explains the poor transient performance of the algorithm in the presence of outliers.

Performance improvement in the presence of outliers could be sought by increasing the value of $\mu$ to reduce the influence of the term $\frac{1}{\mu}\|\vz-\big(\vx(t)-\vs(t)\big)\|^2$ in~\eqref{eq:r_lms_srr_proximal_approx_largeinovations}. However, $\mu$ cannot be made arbitrarily large for stability reasons. Hence, improvement has to come from increasing the importance of the first term in~\eqref{eq:r_lms_srr_proximal_approx_largeinovations}. 

Neglecting the contribution of $\ve(t)$ in \eqref{eq:MagEps0}, expression~\eqref{eq:r_lms_srr_proximal_approx_largeinovations} can be written as
\begin{align}
\vxh_{1}(t) {}\approx{} 2\,\, \underset{\vz}{\arg\min} \Big\{&\alpha \left(\mS \vz \right)^\top\left(\mS \vxh_0(t) \right) - \left( \mD\mH \vz \right)^\top \left( \mD\mH\vs(t)\right) \nonumber\\ 
    &+\frac{1}{2\mu}\left\Vert\vz - \vxh_0(t)\right\Vert^2 \Big\}
    \,\text{.}
    \label{eq:outlier_contribution}
\end{align}
% \begin{equation*}
% \begin{split}
% \vxh_{1}(t) {}&\approx{} \underset{\vz}{\arg\min} \,\,
%     \Big\{2 \alpha \vz^\top\mS^\top\mS\vxh_0(t) 
%     - 2 \vz^\top \mH^\top\mD^\top \eps_0(t)
%     \nonumber\\ & \hspace{4cm}
%     + \frac{1}{\mu}\left\Vert\vz - \vxh_0(t)\right\Vert^2\Big\} \\
%     %
%     &= \cblue{2\,\, \underset{\vz}{\arg\min} \Big\{\alpha \left(\mS \vz \right)^\top\left(\mS \vxh_0(t) \right) - \left( \mD\mH \vz \right)^\top \eps_0(t)}
%     \nonumber\\ & \hspace{4cm}
%     \cblue{+ \frac{1}{2\mu}\left\Vert\vz - \vxh_0(t)\right\Vert^2 \Big\}}\\
%     %
%     &\approx 2\,\, \underset{\vz}{\arg\min} \Big\{\alpha \left(\mS \vz \right)^\top\left(\mS \vxh_0(t) \right) - \left( \mD\mH \vz \right)^\top \left( \mD\mH\vs(t)\right) 
%     \nonumber\\ & \hspace{4cm}
%     +\frac{1}{2\mu}\left\Vert\vz - \vxh_0(t)\right\Vert^2 \Big\}
% \end{split}
% \end{equation*}

First, note that $\vxh_0(t)=\mG(t)\vxh_K(t-1)$ does not include information on the outlier $\vs(t)$, as it is not present in $\vx(t-1)$. Moreover, we have already seen that the first term promotes the smoothness of the solution. Thus, increasing the value of $\alpha$ in an attempt to speed up convergence in the presence of large innovations by reducing the influence of the last term in~\eqref{eq:r_lms_srr_proximal_approx_largeinovations} will also reduce the temporal correlation of the estimated image sequence, resulting in an overly blurred solution with lower quality in the absence of outliers. Hence, the solution $\vxh(t)$ can hardly approach the desired solution $\vx(t)$ in few iterations (small $K$). If, however, $K$ is made sufficiently large, the solution can adapt to track the innovations even with a large weighting for the term $\frac{1}{\mu}\|\vz-\big(\vx(t)-\vs(t)\big)\|^2$. The algorithm could then achieve and maintain a good reconstruction quality both with and without the presence of outliers, but at a prohibitive computational cost.
\subsection{Illustrative example}
\label{sec:r_lms_example}

The behavior of the R-LMS algorithm is illustrated in the following example, where we consider the reconstruction of a synthetic video sequence generated through small translational displacements of a $32\times32$ window over a larger natural image. At a specific time instant during the video (the 32nd frame), an outlier is introduced by adding a black square of size $16\times16$ to the scene. The square remain in the scene for $3$ frames, before disappearing again. The HR sequence is convolved with a $3\times3$ uniform blurring mask and down-sampled by a factor of $2$. Finally, a white Gaussian noise with variance $10$ is added to generate the low resolution video.

\begin{figure}[thb]
\hspace{-0.3cm}
\begin{minipage}[b]{.55\linewidth}
  \centering
  \centerline{\includegraphics[width=1\linewidth]{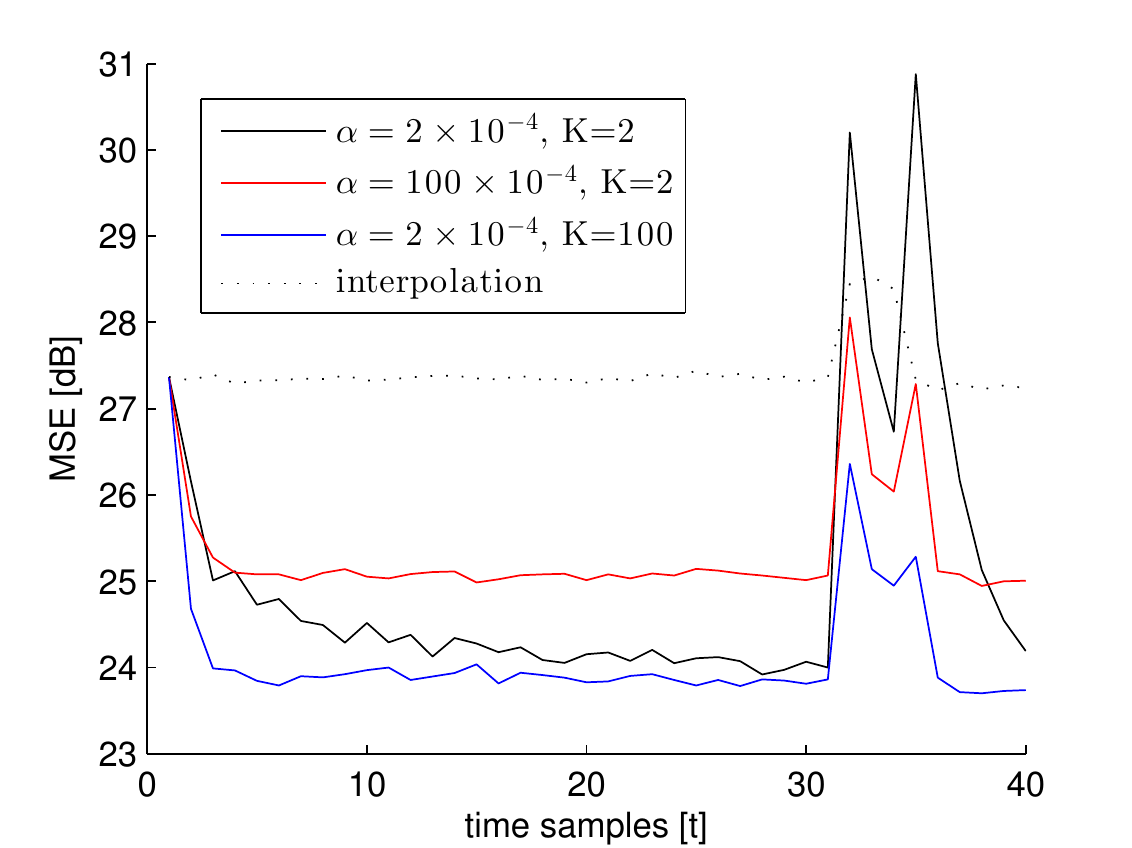}}
  \centerline{(a)}\smallskip
\end{minipage}
%\hfill
\hspace{-0.85cm}
\begin{minipage}[b]{.55\linewidth}
  \centering
  \centerline{\includegraphics[width=1\linewidth]{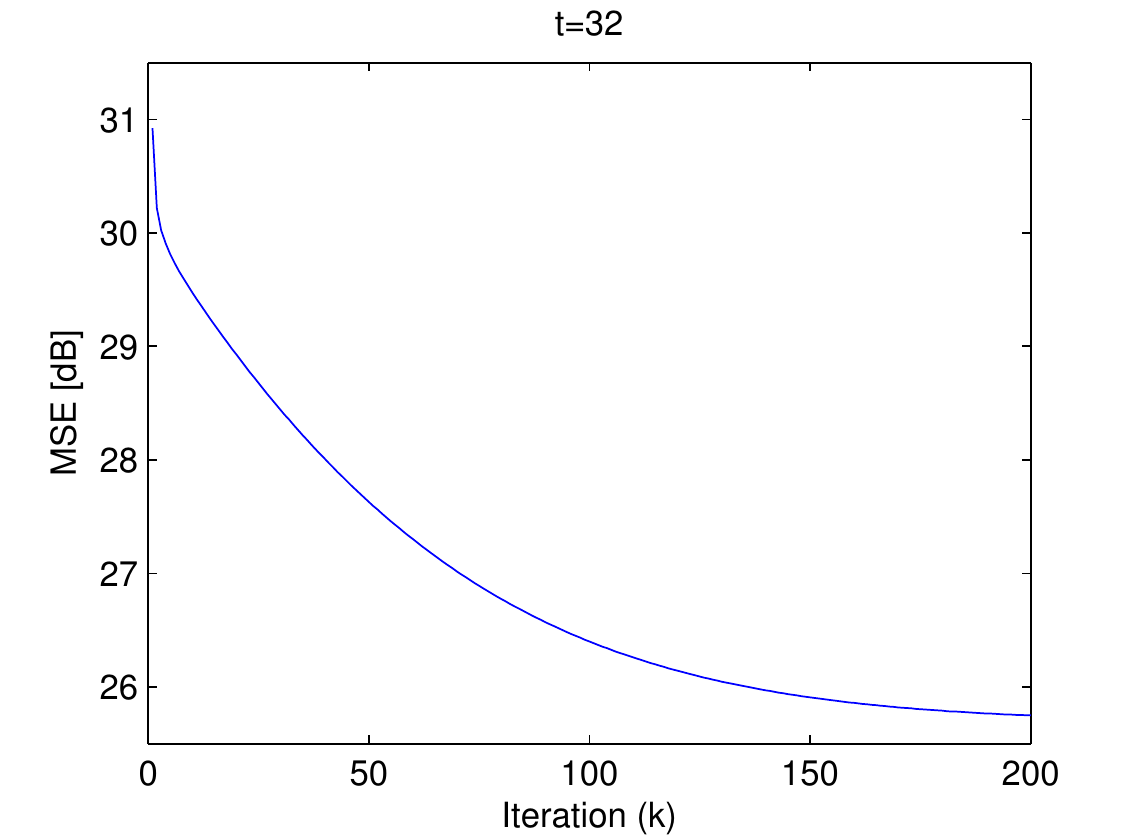}}
  \centerline{(b)}\smallskip
\end{minipage}
\hspace{-1cm}
\vspace{-0.35cm}
\caption{(a) R-LMS algorithm MSE for different values of $\alpha$ and $K$. (b) MSE evolution per iteration during a single time instant $t=32$.}
\label{fig:ex_lms_prob_error_iter}
\end{figure}

The R-LMS algorithm is applied to super-resolve the synthetic LR videos generated. The mean square error (MSE) between the original and reconstructed sequences is estimated by averaging the results from 50 realizations. To illustrate the trade-offs between the effects of different values of the step size and regularization parameter in the cost function, we reconstructed the sequence with both $\alpha=2\times10^{-4}$ and $\alpha=100\times10^{-4}$, for $\mu=4$. To evaluate the effect of using different numbers of iterations per time interval, we ran the algorithm with $K=2$ and $K=100$. The MSE is depicted in Figure~\ref{fig:ex_lms_prob_error_iter}-(a). From the two curves for $K=2$ one can verify that a large value for $\alpha$ (red curve) reduces the MSE in the presence of the outlier, while the greater temporal correlation induced by a small value of $\alpha$ (black curve) tends to reduce the error for small innovations and to increase it in the presence of an outlier. Comparing the blue ($K=100$) and the black ($K=2$) curves, both for $\alpha=2\times10^{-4}$ and $\mu=4$, one verifies that the MSE can be substantially decreased by employing the R-LMS algorithm with a large $K$. The MSE is smaller than that obtained for $K=2$ both for small and for large innovations. This performance improvement is because the algorithm is allowed to converge slowly for each time interval. Figure~\ref{fig:ex_lms_prob_error_iter}-(b) shows the MSE as a function of $k$ for time instant $t=32$, when the outlier is present. These results illustrate the property that a large value of $K$ is necessary to achieve a significant MSE reduction for a fixed value of $t$.

In the light of the aforementioned limitations of the R-LMS algorithm, it is of interest to devise an algorithm that performs better both in terms of robustness and quality at a reasonable computational cost.

% \begin{figure}
%     \centering
%     \includegraphics[width=6cm]{figures/ex_r_lms_prob/MSE_vs_alpha_vs_K2}
%     \caption{R-LMS algorithm MSE for different values of $\alpha$ and $K$.}
%     \label{fig:ex_lms_prob_alphas_K}
% \end{figure}
% \begin{figure}
%     \centering
%     \includegraphics[width=6cm]{figures/ex_r_lms_prob/MSE_per_iter2}
%     \caption{MSE evolution per iteration during a single time instant $t=32$.}
%     \label{fig:ex_lms_prob_error_iter}
% \end{figure}

% ----------------------------------------------------
% ----------------------------------------------------
% ----------------------------------------------------
% ----------------------------------------------------
% ----------------------------------------------------
% ----------------------------------------------------
\section{Improving the Robustness to Innovations}
\label{sec:proposed}

A temporal regularization of adaptive algorithms such as the R-LMS that constrains the value of $\|\vxh(t)-\mG(t)\vxh(t-1)\|$ in the SRR cost function~\cite{Choi96,Borman99,Zibetti07} can be interpreted as the application of the well known least perturbation or minimum disturbance principle. This principle states that \emph{the parameters of an adaptive system should only be disturbed in a minimal fashion in the light of new data}~\cite[p. 355]{Haykin91}. Using this principle, the one-dimensional LMS algorithm can be shown to correspond not to an approximate solution of a gradient-based optimization problem, but to the exact solution of a constrained optimization problem \cite[p. 216]{Sayed03}.

Differently from simultaneous video SRR methods, the cost function~\eqref{eq:RLMS_Lagrangian} of the R-LMS algorithm is defined for a single time instant. Thus, the proximal regularization described in Section~\ref{sec:rlms_with_outliers} only guarantees consistence between consecutive iterations in $k$.
As the solution $\vxh(t-1)$ at the previous time instant is only introduced during the initialization in~\eqref{eq:lms_time_update}, 
consistence between consecutive time instants is only obtained if the parameters of the R-LMS algorithm are selected such that the solution is not disturbed during all iterations $k=1,\ldots,K$ (i.e. $\vxh_K(t)\simeq\vxh_0(t)$). However, as illustrated in the example in section~\ref{sec:r_lms_example}, this makes the R-LMS algorithm very sensitive to outliers. A choice of parameters leading to a superior robustness, on the other hand, compromises the estimation of the details in $\vx(t)$.

% \cblue{To preserve the super-resolved} details between consecutive time instants regardless of the choice \cblue{of the R-LMS parameters}, one might be tempted to \cblue{introduce an additional temporal term to the optimization problem~\eqref{eq:grad_opt_rlms2} to prevent the loss of  content estimated in~$\vxh(t-1)$ when reconstructing~$\vxh(t)$. This  results in the following optimization problem: }
To preserve the super-resolved details between consecutive time instants regardless of the choice of the R-LMS parameters, one might be tempted to introduce an additional temporal term to the optimization problem~\eqref{eq:grad_opt_rlms2} to prevent the loss of  content estimated in~$\vxh(t-1)$ when reconstructing~$\vxh(t)$. This  results in the following optimization problem:
%
% To alleviate this limitation, one might be tempted to modify the optimization problem~\eqref{eq:grad_opt_rlms2} by introducing an additional temporal regularization term as follows:
\begin{align} 
     &\vxh_{k+1}(t) {}={}  \underset{\vz}{\arg\min} \big\{\mathcal{L}_{\tiny{\text{R-MS}}}(\vxh_k(t)) \nonumber\\
    &\hspace{2ex}+ \big(\vz - \vxh_k(t)\big)^{\!\top}\nabla \mathcal{L}_{\tiny{\text{R-MS}}}(\vxh_k(t))+ \frac{1}{\mu}\left\Vert\vz - \vxh_k(t)\right\Vert^2
    \nonumber\\ 
    &\hspace{2ex}+ \frac{1}{\alpha_{\tiny{\text{T}}}\mu}\left\Vert\vz - \mG(t)\vxh(t-1)\right\Vert^2\big\}
    \label{eq:grad_opt_classical_lp}
\end{align}
where $\alpha_{\tiny{\text{T}}}$ is a weighting factor controlling the temporal disturbance.
Albeit removing the dependence of its solution on the time initialization~\eqref{eq:lms_time_update}, the algorithm in~\eqref{eq:grad_opt_classical_lp} fails to achieve good results. Instead, this new regularization term makes the algorithm less robust since it prevents convergence to the desired solution $\vx(t)$ in the presence of large innovations even for a large number of iterations (large $K$). This is clearly perceived by assuming again that $\|\vs(t)\|$ is large and $\vxh(t-1)\simeq\vx(t-1)$, and examining the norm of the last term in~\eqref{eq:grad_opt_classical_lp} for $\vz = \vxh_{k+1}(t)$
\begin{align}
    \|\vxh_{k+1}(t)-\mG(t)\vxh(t-1)\|\approx \|\vxh_{k+1}(t)-\big(\vx(t)-\vs(t)\big)\|
    \nonumber
    %\,\text{,}
\end{align}
which will be large if $\vxh_{k+1}(t)\simeq\vx(t)$ not only for $k=1$, but for all iterations.
Furthermore, this term would be unnecessary for small innovations. In this case the R-LMS can retain the temporal consistency even for a large number of iterations ($K$), as {illustrated} in the example of section~\ref{sec:r_lms_example} for $K=100$. Hence,  algorithm robustness and quality must be addressed using other approaches.

Most works in single-frame or video SRR seek robustness by considering cost functions including non-quadratic (e.g. $\text{L}_1$) error norms~\cite{Farsiu04,Borman99,Zibetti07} or signal dependent regularizations~\cite{Su11,Richter11}, which result in non-linear algorithms. Although these techniques achieve good reconstruction results, their increased computational cost makes real-time operation unfeasible even for the fastest algorithms.
Differently from the simultaneous SRR methods, the robustness problem of the R-LMS is related with its slow convergence, since a good result is achieved for large $K$. A different approach is therefore required to adequately handle the innovations in the R-LMS algorithm.

In the following, we propose to use meaningful \textit{a priori} information about the statistical nature of the innovations in deriving a new stochastic SRR method using the least perturbation principle. The proposed approach can improve the robustness of the R-LMS algorithm while retaining a reduced computational cost. By employing statistical information about $\vs(t)$, which has been overlooked in the design of simple SRR algorithms, it becomes possible to provide robustness to the innovations while maintaining a good reconstruction quality.

%%%%%%%%%%%%%%%%%%%%%%%%%%%%%%%%%%
%%%%%%%%%%%%%%%%%%%%%%%%%%%%%%%%%%
%%%%%%%%%%%%%%%%%%%%%%%%%%%%%%%%%%
%%%%%%%%%%%%%%%%%%%%%%%%%%%%%%%%%%

% --------------------------
\subsection{Constructing an Innovation-Robust Regularization}

{To achieve the desired effect}, we propose to modify the norm being minimized in the last term of~\eqref{eq:grad_opt_classical_lp} through the inclusion of a weighting matrix $\mQ$ {properly} designed to emphasize the image details in the regularization term. This will allow the resulting algorithm to attain a faster speed of convergence with a good quality, while at the same time {reducing the influence of the innovations on the solution of the optimization problem.}
The new constraint is then given by
\begin{align}
    \label{eq:modified_proximal_least_perturbation}
    \big\|\mQ \big( \vxh_{k+1}(t) - \mG(t) \vxh(t-1)\big) \big\|
    %\,\text{,}
\end{align}
{where} $\mQ$ must be designed to preserve only the details of the estimated images, {so that innovations} will have a minimal effect upon the regularization term.  Thus, it is desired that
\begin{align} \label{eq:intuitive_task_of_Q}
    \begin{split}
        \mQ \,\, \vx(t) \sim & \, \text{details}
        % \nonumber
        \\
        \mQ \,\, \vs(t) \sim & \, \mathbf{0}
    \end{split}
\end{align}
which means that the image details must lie in the column space of $\mQ$, while the innovations lie in its nullspace.
Therefore, if we assume the reconstructed image in time instant~$t-1$ to be reasonably close to the real (desired) image, (i.e. $\vxh_K(t-1)\simeq\vx(t-1)$), we can write the modified restriction as
\begin{align}
    \|\mQ\vxh_{k+1}(t) - \mQ\vx(t) + \mQ\vs(t)\|
    \label{eq:modified_restriction_Q_i}
    \,\text{.}
\end{align}
If $\mQ$ {satisfies~\eqref{eq:intuitive_task_of_Q}, $\|\mQ\vs(t)\|\approx0$ even in the presence of an outlier, and~\eqref{eq:modified_restriction_Q_i} can be approximated by}
\begin{align} \nonumber
    \|\mQ\vxh_{k+1}(t) - \mQ\vx(t) + \mQ\vs(t)\|
    \,\, \approx \,\,
    \|\mQ\vxh_{k+1}(t)-\mQ\vx(t)\|
    %\,\text{}
\end{align}
{enabling} the preservation of the image details even in the presence of large innovations.

The question that arises is how to design the transformation matrix $\mQ$ to achieve the required properties. We propose to base the design of $\mQ$ on a stochastic model for the innovations.

% --------------------------
% --------------------------
\subsection{Statistical Properties of {Innovations} in Natural Image Sequences}

% In this section we study the behavior of innovations in natural image sequences with the objective of determining a meaningful definition for the transformation matrix $\mQ$.

The statistical properties of natural images have been thoroughly studied in the literature. A largely employed probabilistic model for natural images is characterized by a zero-mean and highly leptokurtotic, fat-tailed distribution, with its power spectral density remarkably close to a $1/f_r^{\rho}$ function, where $f_r$ is the absolute spatial frequency and $\rho$ is close to $2$~\cite{Schaaf96}. This characterization has led, for example, to the development of sparse derivative prior models for natural images~\cite{Tappen03} that have been widely employed in image processing algorithms.

{When it comes to obtaining accurate probabilistic models for the signals in the dynamic evolution of a video sequence, particularly the innovations, the task becomes more challenging. This is due to the dependence of the signal statistics on the generally unknown movement in the scene. With the motion frequently estimated from a low-resolution observed video sequence, the employed model must distinguish between errors originating from the image registration and errors caused by true changes in the scene, the latter often labeled as outliers.}

% When it comes to representing video sequences, however, obtaining accurate probabilistic models for the signals in the dynamic evolution of the sequence, particularly the innovations, is a more challenging task. This is due to the dependence of the signal statistics on the generally unknown movement in the scene. With the motion estimated from the observed and frequently low-resolution video sequence, the employed model must distinguish between errors originating from the image registration and those caused by true changes in the scene, which are often labeled as outliers in the community.

The modeling of large magnitude changes in a scene has already been considered for the image matching problem. Hasler et. al.~\cite{Hasler03} proposed to consider the error patterns generated by non-coinciding regions of an aligned image pair to be similar to the error generated by comparing two random regions of the underlying scene.
This relationship clearly arises in a dynamical model for a video sequence when the motion model fails to account for unpredictable changes between two adjacent images, generating an error signal that will consist of the difference between the new image and a misaligned part of the previous image.
Considering the case of one dimensional signals for simplicity, the auto-correlation function of the difference between two patches of an image separated by $\Delta$ samples can be computed as
\begin{align}
    \vr_{\Delta}(l) & = \Ex[ \{I(p)-I(p-\Delta)\} \{ I(p-l)-I(p-\Delta-l)\} ]
                    \nonumber\\
                    & = 2\vr_I(l) - \vr_I(l-\Delta) - \vr_I(l+\Delta) 
\end{align}
where $\Ex[\,\cdot\,]$ denotes statistical expectation, $I(p)$ is a point in the one dimensional image, {and $\vr_I(l)$ is the image auto-correlation function. Thus, $\vr_{\Delta}(l)$ is the auto-correlation of the simulated outlier.}
If the covariance between the image pixels diminishes with their distance, for a sufficiently large value of $\Delta$ the terms $\vr_I(l \pm \Delta)$ will become approximately equal to the square of the mean image value.
Therefore, the auto-correlation function of the simulated outlier will be similar to that of a natural image.

% =================
This interpretation can be more intuitively achieved by considering a different {approach, which models the innovations assuming} a scene model composed by the interactions of objects in an occlusive environment~\cite{Lee01deadLeaves}.
Innovations in a video sequence can be broadly described as pixels in $\vx(t)$ that cannot be described as a linear combination of the pixels in $\vx(t-1)$ (i.e. are statistically orthogonal). These pixels will be here divided as
\begin{align}
    \vs(t)=\vd(t)+\veta(t)
    \label{eq:innovation_lowlevel_model_separation}
\end{align}
where $\veta(t)$ consists of small changes {in the scene, originated, for instance} from specular surfaces. $\veta(t)$ can be modeled as a low power high frequency noise. $\vd(t)$ represents large magnitude changes (outliers) arising due to occlusions or due to objects suddenly appearing in the scene (such as image borders). $\vd(t)$ is usually sparse and compact~\cite{Baker11databaseOF}
\footnote{Note that $\vs(t)$ is not to be confused with registration errors due to the ill-posed nature of the motion estimation process. The latter can be shown to originate from a random linear combination of the pixels in $\vx(t-1)$~\cite{Costa07}.}.

% An important property of $\vd(t)$ is it's spatial correlation.
A region of the scene corresponding to a dis-occluded area typically reveals part of a background or object at a different depth from the camera. Hence, the nonzero pixels in $\vd(t)$ will consist of highly correlated compact regions. Furthermore, the joint pixel statistics at these locations should actually be similar to that of natural scenes. This conclusion becomes straightforward if we consider, for instance, the Dead Leaves image formation model~\cite{Lee01deadLeaves}, which characterizes a natural scene by a superposition of opaque objects of random sizes and positions occluding each other. Here, a dis-occluded area would correspond to the removal of an object (or a ``leaf") at random from the topmost of the z-axis. The corresponding region in the new image will therefore be composed of the next objects present on the z-axis. Since the area behind the view plane is completely filled with objects (superimposed "leaves") in this model, there is no difference between the statistical properties of a region in the foremost-top image and those of a region behind an object. This reinforces the notion of correlation obtained by considering the more generic outlier model of Hasler et al.~\cite{Hasler03}.

% Having a statistical characterization of the properties of $\vs(t)$, the important question is how to translate it into an operator $\mQ$ such that the objectives depicted in \eqref{eq:intuitive_task_of_Q} are satisfied.

% => present innovations model 2 (d+eta)
% === Sub Chapter: constructing Q

\begin{figure}
    \centering
    \includegraphics[width=6cm]{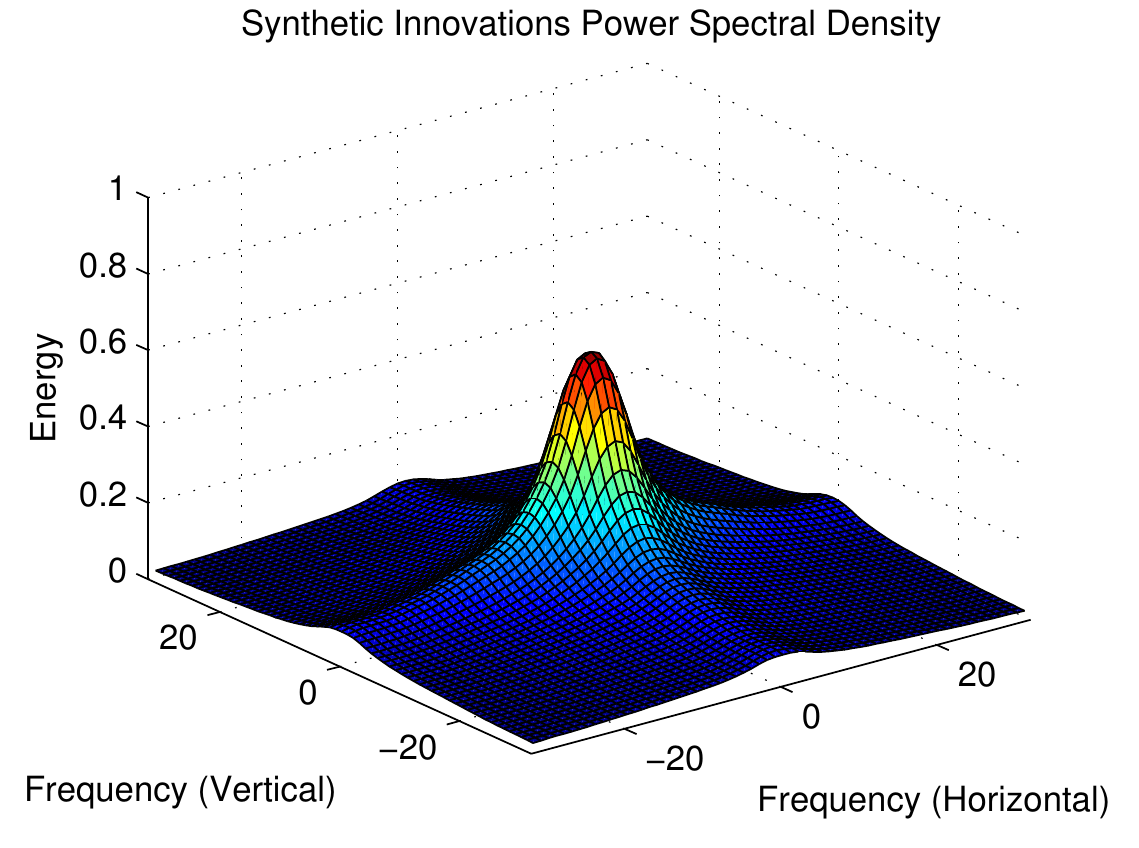}
    % \vspace{-0.3cm}
    \caption{Power spectral density of synthetically generated innovations.}
    \label{fig:inn_energy_synthetic_simulations}
\end{figure}

To verify the proposed innovations model, we have determined the power spectral density (PSD) of synthetic images representing the innovations. These images were generated by pasting small pieces of the difference between two independent natural images with sizes ranging from $5\times5$ to $15\times15$ in random positions of a $64\times64$ background.  We have extracted the small pieces from 20 different natural images, so that they emulate small regions appearing in the occluded regions of a video sequence. The PSD is computed by averaging 200 realizations of a Monte Carlo (MC) simulation. Figure~\ref{fig:inn_energy_synthetic_simulations} shows the obtained result. It can be clearly seen that the energy is concentrated in the lower frequencies of the spectrum, resulting in a highly correlated signal.

% --------------------------
\subsubsection{Choosing the Operator $\mQ$}

Natural scene innovations tend to be highly correlated in space. Thus, their energy tends to be primarily concentrated in the low spatial frequencies. Hence, the operator $\mQ$ should in general emphasize the high frequency components to accomplish the design objectives in~\eqref{eq:intuitive_task_of_Q}. Unfortunately, the specific scenes to be processed are not known in advance, what hinders the {accurate} determination of the statistical properties of the innovations, and thus of the optimal operator $\mQ$.  A simple solution with reduced computational complexity is to use a {basic} high-pass filter with small support, such as a differentiator or a Laplacian. For simplicity, the Laplacian filter mask will be employed during the remaining of this work. Thus, we shall use $\mQ=\mS$, leaving the search for an optimal operator for a future work.

%---------------------------------------------------------------
%---------------------------------------------------------------
%\section{{A New Fast Adaptive SRR Algorithm Robust to Innovations}}
\section{The Temporally Selective Regularized LMS (TSR-LMS) Algorithm}

To derive the new algorithm, we propose a new cost function that minimizes the perturbation only on the details of the reconstructed image, while at the same time observing the objectives of the R-LMS algorithm. Differently from~\eqref{eq:grad_opt_classical_lp}, the new cost function allows for more flexibility for the component of the solution in the subspace corresponding to the outlier while retaining its quality. Such strategy leads to an increased algorithm robustness. 

We propose to solve the following optimization problem:
\begin{align}
    \vxh_{k+1}(t){}={}&\underset{\vz}{\arg\min} \,\, \Big\{\mathcal{L}_{\tiny{\text{R-MS}}}(\vxh_k(t)) \nonumber\\
    &+ \big(\vz - \vxh_k(t)\big)^{\top}\nabla \mathcal{L}_{\tiny{\text{R-MS}}}(\vxh_k(t))+ \frac{1}{\mu}\left\Vert\vz - \vxh_k(t)\right\Vert^2
    \nonumber\\ &
    + \frac{1}{\alpha_{\tiny{\text{T}}}\mu}\left\Vert  \mQ\vz - \mQ\mG(t)\vxh(t-1)\right\Vert^2\Big\}
    \,\text{.}
    \label{eq:grad_opt_lp_1}
\end{align}

Calculating the gradient of the cost function with respect to $\vz$, setting it equal to $\textbf{0}$, solving for $\vz=\vxh_{k+1}(t)$ and approximating the statistical expectations by their instantaneous values yields the iterative equation for the TSR-LMS algorithm:
\begin{align}
    & \vxh_{k+1}(t)  = \Big(\mI+\!\frac{1}{\alpha_{\tiny{\text{T}}}}\mQt\mQ\Big)^{\!-1} \! \Big\{
    \vxh_k(t) + \frac{1}{\alpha_{\tiny{\text{T}}}}\mQt\mQ\mG(t)\vxh(t-1)
    \nonumber \\ & \hspace{1cm}
    -\mu\mHt\mDt\big[\mD\mH\vxh_k(t)-\vy(t)\big]
    -\mu\alpha\mSt\mS\vxh_k(t) \Big\}
    \label{eq:new_lmslp_iterative_1}
\end{align}
where the time update is based on the signal dynamics~\eqref{eq:dinam} and performed by $\vxh_{0}(t+1)=\mG(t+1)\vxh_{K}(t)$~\cite{Elad99}.

Algorithm~\eqref{eq:new_lmslp_iterative_1} generalizes R-LMS and the least perturbation approach~\eqref{eq:grad_opt_classical_lp}. It collapses to these solutions if $\alpha_{\tiny{\text{T}}}\rightarrow\infty$ or $\mQ=\mI$, respectively.
%
% It is clear that the proposed algorithm is a generalization of the R-LMS algorithm and of the classic least perturbation approach~\eqref{eq:grad_opt_classical_lp}, as it collapses to these solutions if $\alpha_{\tiny{\text{T}}}\rightarrow\infty$ or $\mQ=\mI$, respectively.
%
It should perform well both with and without outliers, at the cost of little extra computational effort. Though the matrix inversion can be made \textit{a priori} and the resulting inverse might be sparse, its storage is still rather costly. If $\mQ$ is chosen to be block circulant (BC) (such as a Laplacian), then $\big(\mI+\frac{1}{\alpha_{\tiny{\text{T}}}}\mQt\mQ\big)^{-1}$ is block circulant~\cite{Mazancourt83inverse} and can be computed as a convolution, leading to important memory savings.
% Algorithm~\eqref{eq:new_lmslp_iterative_1} should have a good performance both with and without the presence of the outliers, at the cost of little additional computational effort. A detailed comparison is present at the end of the present section. Although the matrix inversion can be made \textit{a priori} and the resulting inverse might be sparse, its storage is still rather costly. If the operator $\mQ$ is chosen to be a block circulant (BC) matrix (such as a Laplacian), then $\big(\mI+\frac{1}{\alpha_{\tiny{\text{T}}}}\mQt\mQ\big)^{-1}$ is known to be block circulant as well~\cite{Mazancourt83inverse}, and can therefore be computed as a convolution, leading to important memory savings.

Although~\eqref{eq:new_lmslp_iterative_1} may resemble the Gradient Projection Method (GPM)~\cite{Bertsekas76gradProj}, this is not generally true, as {$\mM = \big(\mI+\frac{1}{\alpha_{\tiny{\text{T}}}}\mQt\mQ\big)^{-1}$} is not necessarily a projection matrix (i.e. $\mM^2\neq\mM$). Hence, the convergence and stability properties of these algorithms are not the same in general.
%results obtained for the GPM algorithm cannot be directly applied to~\eqref{eq:new_lmslp_iterative_1}, making the performance analysis and parameter design more difficult.

% -------------------------------------------------
%\subsection{{A Simplified Algorithm}}
\subsection{The Linearized Temporally Selective Regularized LMS (LTSR-LMS)}

Whereas algorithm~\eqref{eq:new_lmslp_iterative_1} should present a good cost-benefit ratio, the aforementioned limitations motivates the pursuit of another algorithm that trades a small performance loss for a simpler implementation and a more predictable performance. This section describes one possible modification.

Since the details of the solution are minimally disturbed between iterations, we can safely assume that $\mQ\vxh_{k+1}(t)\approx\mQ\vxh_k(t)$. Therefore, we can employ a linear approximation for the quadratic regularization introduced in the last term of~\eqref{eq:grad_opt_lp_1} using a first-order Taylor series expansion of this norm with respect to the transformed variable $\mQ\vz$ about the point $\mQ\vz=\mQ\vxh_k(t)$. The resulting cost function can be written as:
\begin{align}
    \vxh_{k+1} & (t) {}={} \underset{\vz}{\arg\min}  \Big\{\, \mathcal{L}_{\tiny{\text{R-MS}}}(\vxh_k(t))
    \nonumber\\ &
    + \big(\vz - \vxh_k(t)\big)^{\top}\nabla \mathcal{L}_{\tiny{\text{R-MS}}}(\vxh_k(t)) + \frac{1}{\mu}\left\Vert\vz - \vxh_k(t)\right\Vert^2 \nonumber
% \end{align}
% \begin{align}   
\nonumber \\ 
    &+ \frac{1}{\tilde{\alpha_{\tiny{\text{T}}}}\mu}
    \big\{\mQ\vxh_k(t) - \mQ\mG(t)\vxh(t-1)\big\}^{\!\top} \!
    \big\{\mQ\vz - \mQ\vxh_k(t)\big\}
    \nonumber\\ &
    + \frac{1}{2\tilde{\alpha_{\tiny{\text{T}}}}\mu}
    \|\mQ\vxh_k(t)-\mQ\mG(t)\vxh(t-1)\|^2
     \Big\}
    \text{.}
    \label{eq:grad_opt_lp_2}
%     \\&
%     %
%     \cblue{=\underset{\vz}{\arg\min}  \Big\{\, \mathcal{L}_{\tiny{\text{R-MS}}}(\vxh_k(t))
%     + \big(\vz - \vxh_k(t)\big)^{\top}\nabla \mathcal{L}_{\tiny{\text{R-MS}}}(\vxh_k(t))}
%     \nonumber\\ & 
%     \cblue{+ \frac{1}{\mu}\left\Vert\vz - \vxh_k(t)\right\Vert^2
%     % \nonumber\\&%\hspace{1cm}
%     + \frac{1}{\tilde{\alpha_{\tiny{\text{T}}}}\mu}
%     \big\{ \mQ\big(\vz-\mG(t)\vxh(t-1)\big) \big\}^{\top}}
%     \nonumber\\ &
%     \cblue{\cdot \big\{ \mQ\big(\vxh_k(t) - \mG(t)\vxh(t-1)\big) \big\} \Big\}}
%     \, \text{.}
%     \label{eq:grad_opt_lp_2_alternate}
\end{align}

% \begin{align}
%     \vxh_{k+1} & (t) {}={} \underset{\vz}{\arg\min}  \Big\{\, \mathcal{L}_{\tiny{\text{R-MS}}}(\vxh_k(t))
%     \nonumber\\ &
%     + \big(\vz - \vxh_k(t)\big)^{\top}\nabla \mathcal{L}_{\tiny{\text{R-MS}}}(\vxh_k(t))
%     % \nonumber\\ &
%     + \frac{1}{\mu}\left\Vert\vz - \vxh_k(t)\right\Vert^2
%     \nonumber\\ & 
%     + \frac{1}{\tilde{\alpha_{\tiny{\text{T}}}}\mu}
%     \big\{\mQ\vxh_k(t) - \mQ\mG(t)\vxh(t-1)\big\}^{\!\top} \!
%     \big\{\mQ\vz - \mQ\vxh_k(t)\big\}
%     \nonumber\\ &
%     + \frac{1}{2\tilde{\alpha_{\tiny{\text{T}}}}\mu}
%     \|\mQ\vxh_k(t)-\mQ\mG(t)\vxh(t-1)\|^2
%      \Big\}
%     %
%     \text{.}
%     \label{eq:grad_opt_lp_2}
% %     \\&
% %     %
% %     \cblue{=\underset{\vz}{\arg\min}  \Big\{\, \mathcal{L}_{\tiny{\text{R-MS}}}(\vxh_k(t))
% %     + \big(\vz - \vxh_k(t)\big)^{\top}\nabla \mathcal{L}_{\tiny{\text{R-MS}}}(\vxh_k(t))}
% %     \nonumber\\ & 
% %     \cblue{+ \frac{1}{\mu}\left\Vert\vz - \vxh_k(t)\right\Vert^2
% %     % \nonumber\\&%\hspace{1cm}
% %     + \frac{1}{\tilde{\alpha_{\tiny{\text{T}}}}\mu}
% %     \big\{ \mQ\big(\vz-\mG(t)\vxh(t-1)\big) \big\}^{\top}}
% %     \nonumber\\ &
% %     \cblue{\cdot \big\{ \mQ\big(\vxh_k(t) - \mG(t)\vxh(t-1)\big) \big\} \Big\}}
% %     \, \text{.}
% %     \label{eq:grad_opt_lp_2_alternate}
% \end{align}

Note that if the algorithm initialization is selected as $\vxh_{0}(t)=\mG(t)\vxh_{K}(t-1)$~\cite{Elad99}, the linearized regularization introduced in the last term of~\eqref{eq:grad_opt_lp_2} is equal to zero for the first iteration ($k=1$). Therefore, $K\geq2$ iterations per time instant are necessary in order to have an improvement over the R-LMS algorithm. This is not the case for the algorithm proposed in~\eqref{eq:new_lmslp_iterative_1}, where an improvement can be obtained even for $K=1$.

{By ignoring the constant terms in the optimization problem~\eqref{eq:grad_opt_lp_2} and using~\eqref{eq:RLMS_Lagrangian} and~\eqref{eq:grad_sup_des}, it can be shown that~\eqref{eq:grad_opt_lp_2} corresponds to the proximal point cost function of the following Lagrangian for a single time time~$t$:}
\begin{align} \label{eq:lmslp_lagrangian}
    &\mathcal{L}(\vxh(t))=\Ex\big\{\,\|\mD\mH\vxh(t)-\vy(t)\|^2
    \big|\,\vxh(t)\big\} + \alpha\|\mS\vxh(t)\|^2
    \nonumber\\ & \hspace{2.5cm}
    +\alpha_{\tiny{\text{T}}} \|\mQ\vxh(t)-\mQ\mG(t)\vxh(t-1)\|^2
    %\,\text{,}
\end{align}
%where $\alpha_{\tiny{\text{T}}}=\frac{1}{2\tilde{\alpha_{\tiny{\text{T}}}}\mu}$.
where $\alpha_{\tiny{\text{T}}}=1/(2\tilde{\alpha_{\tiny{\text{T}}}}\mu)$.
Note that by using $\mQ=\mI$ on~\eqref{eq:lmslp_lagrangian}, the algorithm particularizes to the well known temporal regularization, commonly employed in simultaneous video SRR in order to achieve temporal consistency~\cite{Choi96,Zibetti07}. In this case, the algorithm is not expected to be robust since the innovations are not accounted for.

Calculating the gradient of the cost function in~\eqref{eq:grad_opt_lp_2} with respect to $\vz$, setting it equal to $\textbf{0}$, solving for $\vz=\vxh_{k+1}(t)$ and approximating the statistical expectations by their instantaneous values yields the iterative equation for the LTSR-LMS algorithm based on the linearized version of the proposed regularization:
\begin{align} \label{eq:new_lmslp_iterative_2}
    &\vxh_{k+1}(t)=\vxh_k(t)-\mu\alpha_{\tiny{\text{T}}}\mQt\big[\mQ\vxh_k(t)-\mQ\mG(t)\vxh(t-1)\big]
    \nonumber\\&\hspace{1cm} 
    -\mu\mHt\mDt\big[\mD\mH\vxh_k(t)-\vy(t)\big]-\mu\alpha\mSt\mS\vxh_k(t)
    %\, \text{,}
\end{align}
where the update is performed for a fixed $t$ and for $k=1,$ $\ldots,K$. Like the traditional R-LMS, the time update of~\eqref{eq:new_lmslp_iterative_2} is based on the signal dynamics~\eqref{eq:dinam}, and performed by $\vxh_{0}(t+1)=\mG(t+1)\vxh_{K}(t)$~\cite{Elad99}.

% =================================================
\subsection{Computational Cost of the Proposed Solution}

The computational and memory costs of the (L)TSR-LMS algorithms are still comparable to those of the (R)-LMS algorithm.
An important characteristic of the problem that allows a fast implementation of both the (R)-LMS and the (L)TSR-LMS methods is the spatial invariance assumption of the operators $\mM=(\mI+\frac{1}{\alpha_{\text{\tiny T}}}\mQt\mQ)^{-1}$, $\mH$, $\mS$ and $\mQ$, which results in them being block-circulant or block-Toeplitz matrices.
In this case, the number of nonzero elements of the matrices (denoted by $|\,\cdot\,|$) scales linearly with the number of HR image pixels (i.e. $\propto M^2$), and so does the number of operations of the algorithms.
In this case, the computational and memory costs for the algorithms considered can be seen in Tables~\ref{tab:alg_mem_cost} and~\ref{tab:alg_comp_cost}.

\begin{table} [htb]
% \small
\footnotesize
% \scriptsize
\caption{Memory cost of the algorithms.}
\vspace{-0.2cm}
\centering
% \begin{center}
\renewcommand{\arraystretch}{1.2}
\begin{tabular}{c|cccc}
\hline
& \textbf{Memory} \\
\hline
LMS & $M^2+|\mathbf{H}|/M^2$ \\
R-LMS & $M^2+\big(|\mathbf{H}|+|\mathbf{S}|\big)/M^2$ \\
TSR-LMS  & $2M^2+\big(|\mathbf{H}|+|\mathbf{S}|+|\mathbf{M}|+|\mathbf{Q}|\big)/M^2$ \\
LTSR-LMS  & $2M^2+\big(|\mathbf{H}|+|\mathbf{S}|+|\mathbf{Q}|\big)/M^2$ \\
\hline
\end{tabular}
% \end{center}
\label{tab:alg_mem_cost}
\end{table}

\begin{table}[htb]
% \small
\footnotesize
% \scriptsize
\caption{Computational cost per iteration of the algorithms (additions and multiplications, surplus additions, and re-samplings were considered).}
\vspace{-0.2cm}
\centering
% \begin{center}
\renewcommand{\arraystretch}{1.2}
\begin{tabular}{c|cccc}
\hline
& \textbf{Operations} \\
\hline
LMS & $3|\mathbf{H}|+2M^2$\\
R-LMS & $3|\mathbf{H}|+2|\mathbf{S}|+2M^2$ \\
TSR-LMS  &  $3|\mathbf{H}|+2|\mathbf{S}|+2|\mathbf{Q}|+|\mathbf{M}|+2M^2$\\
LTSR-LMS & $3|\mathbf{H}|+2|\mathbf{S}|+2|\mathbf{Q}|+3M^2$  \\
\hline
\end{tabular}
% \end{center}
\label{tab:alg_comp_cost}
\end{table}

% \label{sec:examples_real_i}

\section{Results}
\label{sec:results}
% exemplo1: MC com movimento e sequencia sinteticos, com e sem algoritmo de registro
% exemplo2: MC com flying bird, inserting a square on a frame. MSE and perceptual results
% example3: SR of real video sequences, one or two videos, OF registration, show frames with high innovation outliers
% possible 4th example: SR of real video sequences considering translational motion model, shot that it is more robust to outliers.

% \eqref{eq:new_lmslp_iterative_1}
% \eqref{eq:new_lmslp_iterative_2}

We now present four examples to illustrate the performace of the TSR-LMS and LTSR-LMS methods. Examples~1 and~2 use a controlled environment to assess differences in quality (section~\ref{sec:examples_synth_i}) and robustness (section~\ref{sec:examples_synth_ii}) when compared to interpolation, LMS and R-LMS without the influence of unaccountable effects.
Example~3 (section~\ref{sec:examples_real_i}) compares the performances of the proposed methods to those of LMS, R-LMS and interpolation algorithms using real video sequences. Example~4 (section~\ref{sec:examples_real_ii}) evaluates the TSR-LMS and LTSR-LMS methods against state-of-the-art SRR algorithms in practical applications.

Example~1 evaluates the average performance of the algorithms without outliers, in a close-to-ideal environment. We used synthetic video sequences with small translational motion to enable Monte Carlo simulations and to be able to control the occurrence of modeling errors. The motion between frames was assumed known \textit{a priori}, and the mean squared reconstruction error could be evaluated because we had access to the desired HR images.
The simulation was also performed using a typical registration algorithm to evaluate the influence of motion estimation on the performances of algorithms~\eqref{eq:new_lmslp_iterative_1} and~\eqref{eq:new_lmslp_iterative_2}. A decline in performance was expected, as reported in~\cite{Choi96} for the classical temporal regularization algorithms.

Example~2 evaluates the proposed algorithms in the presence of innovation outliers. A synthetic simulation emulates the case of a \textit{flying bird} when an object suddenly appears in a frame or moves independently of the background, generating occlusions and leading to a high level of innovations in some specific frames of the video sequence.

Example~3 evaluates the performances of the algorithms when super-resolving real video sequences. We super-resolved a set of video sequences containing complex motion patterns and frames with large levels of innovations and registration errors.
Finally, Example~4 extends Example~3 by comparing the TSR-LMS and LTSR-LMS methods with state-of-the-art SRR algorithms, namely, a Bayesian method~\cite{liu2014bayesianVideoSRR} and a Convolutional Neural Network~\cite{tao2017detailRevealingSRRneuralNets}. We employ recent and challenging video sequences and compare the results for robustness, quality and computational cost.

For $\mQ=\mI$, the LTSR-LMS algorithm particularizes to the popular classical temporal regularization~\cite{Choi96,Zibetti07}. We do not report this case here, as we could not obtain any improvement (quantitative or perceptual) when compared to the R-LMS ($\mQ=0$) performance. The matrix $\mQ$ employed in both algorithms for all examples shown here was a Laplacian filter.
The SRR algorithms were compared to the bicubic interpolation algorithm proposed in~\cite{blu2001moms}. Both the MSE and the structural similarity index (SSIM) were considered in the evaluation. The obtained SSIM results were qualitatively similar to the MSE results. Hence, the SSIM values are provided only for the displayed images.

The boundary condition for the convolution matrices was chosen to be circulant. This simplifies the implementation and results in the inversion of a circulant matrix in~\eqref{eq:new_lmslp_iterative_1}~\cite{Mazancourt83inverse}.
We selected the boundary condition for $\mG(t)$ in the global translational case to be circulant as well to simplify implementation. For the case of a dense motion field, the warped images were computed through the bilinear interpolation of the original image pixels.

% ------------------------------------------------------------------
\subsection{Example 1} \label{sec:examples_synth_i}
For a Monte Carlo simulation, each HR video sequence was created based on the translation of an $256\times256$ window over a static image, resulting in whole-image translational movements. The window displacements consisted in a random walk process (i.i.d. unitary steps) on both horizontal and vertical directions. The still images used to generate each video sequence were natural scenes such as \textit{Lena}, \textit{Cameraman}, \textit{Baboon} and others, and were totally distinct from each other. The resulting sequence was then blurred with a uniform unitary gain $3\times3$ mask and decimated by a factor of $2$, resulting in LR images of dimension $N=128$. Finally, white Gaussian noise with variance $\sigma^2=10$ was added to the decimated images. The algorithm performances were evaluated by averaging 50 realizations, one for each input image.

% In this example, the performance of the proposed method is evaluated through a Monte Carlo (MC) simulation with 50 realizations. In order to enable the evaluation of the mean square reconstruction error (MSE), each HR video sequence is created based on the translation of an $256\times256$ window over a static image, resulting in whole-image translational movements. 

The performances of the different algorithms are highly dependent on the  parameters selected, as verified in~section~\ref{sec:r_lms_example}). Hence, the parameters were carefully selected to yield an honest comparison.
We selected the parameters for each method to achieve the minimum steady-state MSE (i.e. for large $t$). The steady-state MSE for each set of parameters was estimated by running an exhaustive search over a small, independent set of images and averaging the squared errors for the last 5 frames. Table~\ref{tab:parametersKnown} shows the parameter values that resulted in the best performance for each algorithm.

We applied both standard and regularized versions of LMS and the proposed methods to super resolve the synthetic sequences, all initialized with $\vxh(1)$ as a bicubic (spline) interpolation of the first LR image.
First, we considered the motion vectors to be known \textit{a priori} to avoid the influence of a registration algorithm, and used $K=2$ iterations per time instant.
%
%(notice that due to the time update in~\eqref{eq:lms_time_update}, when $K=1$ the method proposed in~\eqref{eq:new_lmslp_iterative_2} particularizes to the R-LMS algorithm in~\eqref{eq:lms}).
%
The super-resolved sequences were compared to the original HR sequence and the MSE was computed across all realizations.

\begin{table}
% \small
\footnotesize
% \scriptsize
\caption{Parameter values used on the simulations with the with outlier-free sequences.}
\vspace{-0.2cm}
\centering
% \begin{center}
\renewcommand{\arraystretch}{1.2}
\begin{tabular}{lcccc}
\hline
& \textbf{LMS} & \textbf{R-LMS} & \textbf{TSR-LMS} & \textbf{LTSR-LMS} \\
\hline
$\mu$      &  2  &  2.75  &  1.15  &  3  \\
$\alpha$   &  --  &  5$\times10^{-4}$  &  1.5$\times10^{-4}$  &  1$\times10^{-4}$\\
$\alpha_{\text{T}}$ &  --  &  --  &  82  &  0.02  \\
\hline
\end{tabular}
% \end{center}
\label{tab:parametersKnown}
\end{table}

\begin{figure}
\centering
\includegraphics[width=6cm,trim={0 0 0 0.4cm},clip]{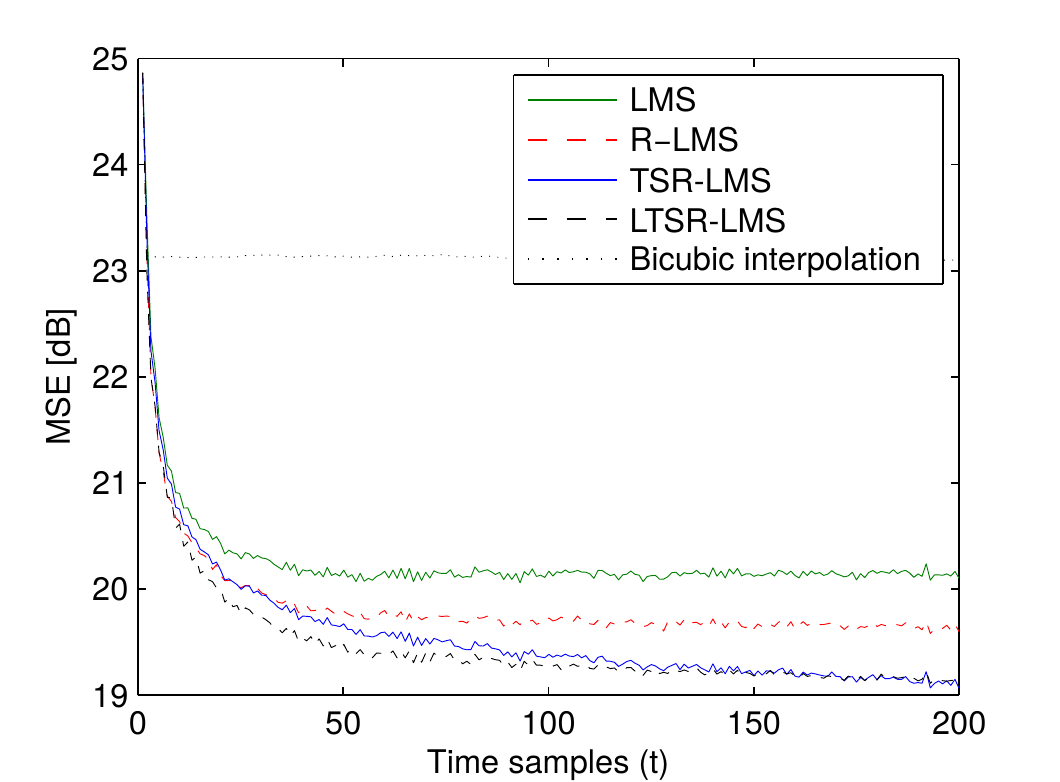}
\vspace{-0.2cm}
\caption{Average MSE per pixel for the known motion case.}
\label{fig:MSEknown}
\end{figure}

The MSE performance is depicted in Figure~\ref{fig:MSEknown}. It can be seen that the proposed methods outperform LMS and R-LMS. 
Moreover, both proposed algorithms achieve the same MSE given enough time. Finally, the LTSR-LMS algorithm reaches the minimum MSE faster than the original TSR-LMS algorithm in the absence of outliers.

\begin{figure}
\centering
\includegraphics[width=6cm,trim={0 0 0 0.4cm},clip]{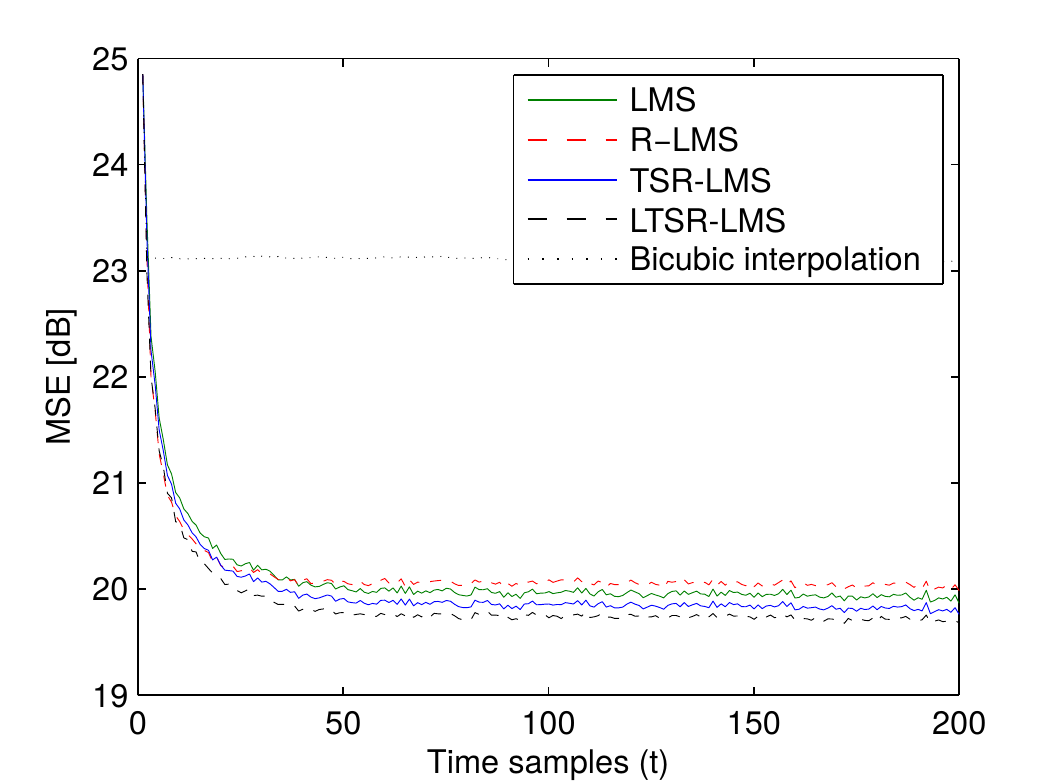}
\vspace{-0.2cm}
\caption{Average MSE per pixel in the presence of registration errors.}
\label{fig:MSEreg}
\end{figure}

\begin{figure*}[htb]
\begin{minipage}[b]{.16\textwidth}
  \centering
  \centerline{\includegraphics[width=0.9\linewidth]{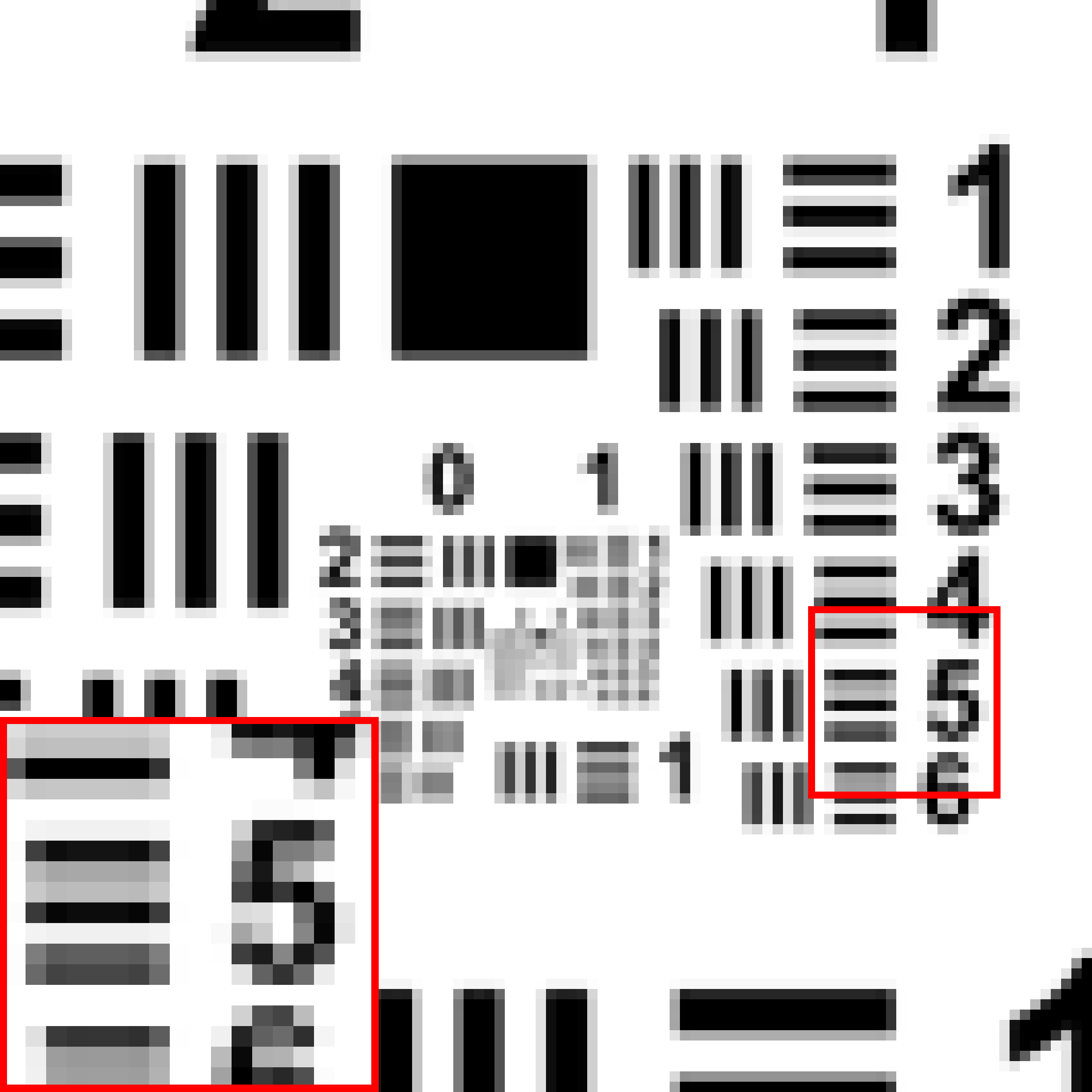}}
  \centerline{(a)}\medskip
\end{minipage}
%\hfill
%
\begin{minipage}[b]{.16\textwidth}
  \centering
  \centerline{\includegraphics[width=0.9\linewidth]{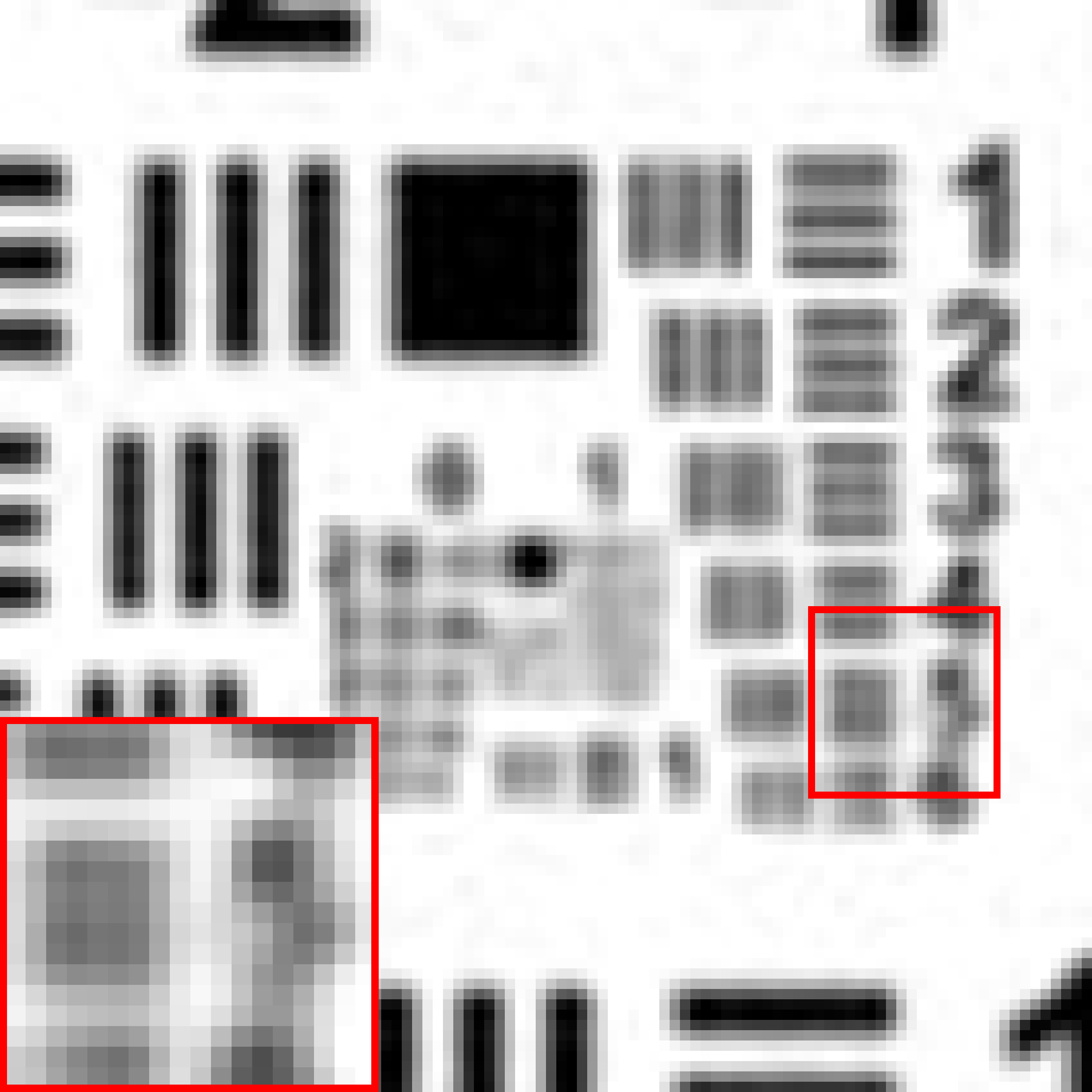}}
  \centerline{(b)}\medskip
\end{minipage}
\begin{minipage}[b]{.16\textwidth}
  \centering
  \centerline{\includegraphics[width=0.9\linewidth]{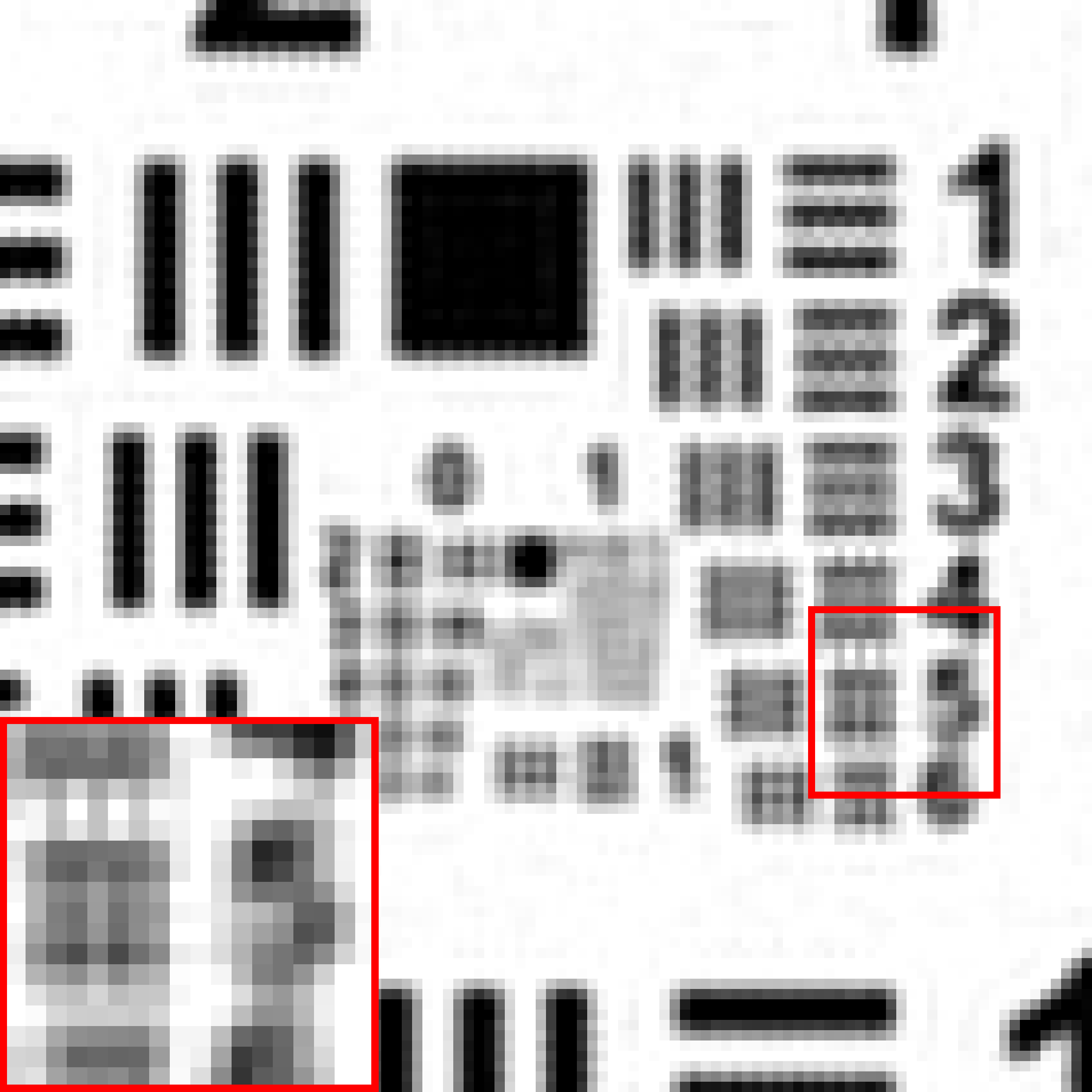}}
  \centerline{(c)}\medskip
\end{minipage}
%\hfill
%
\begin{minipage}[b]{.16\textwidth}
  \centering
  \centerline{\includegraphics[width=0.9\linewidth]{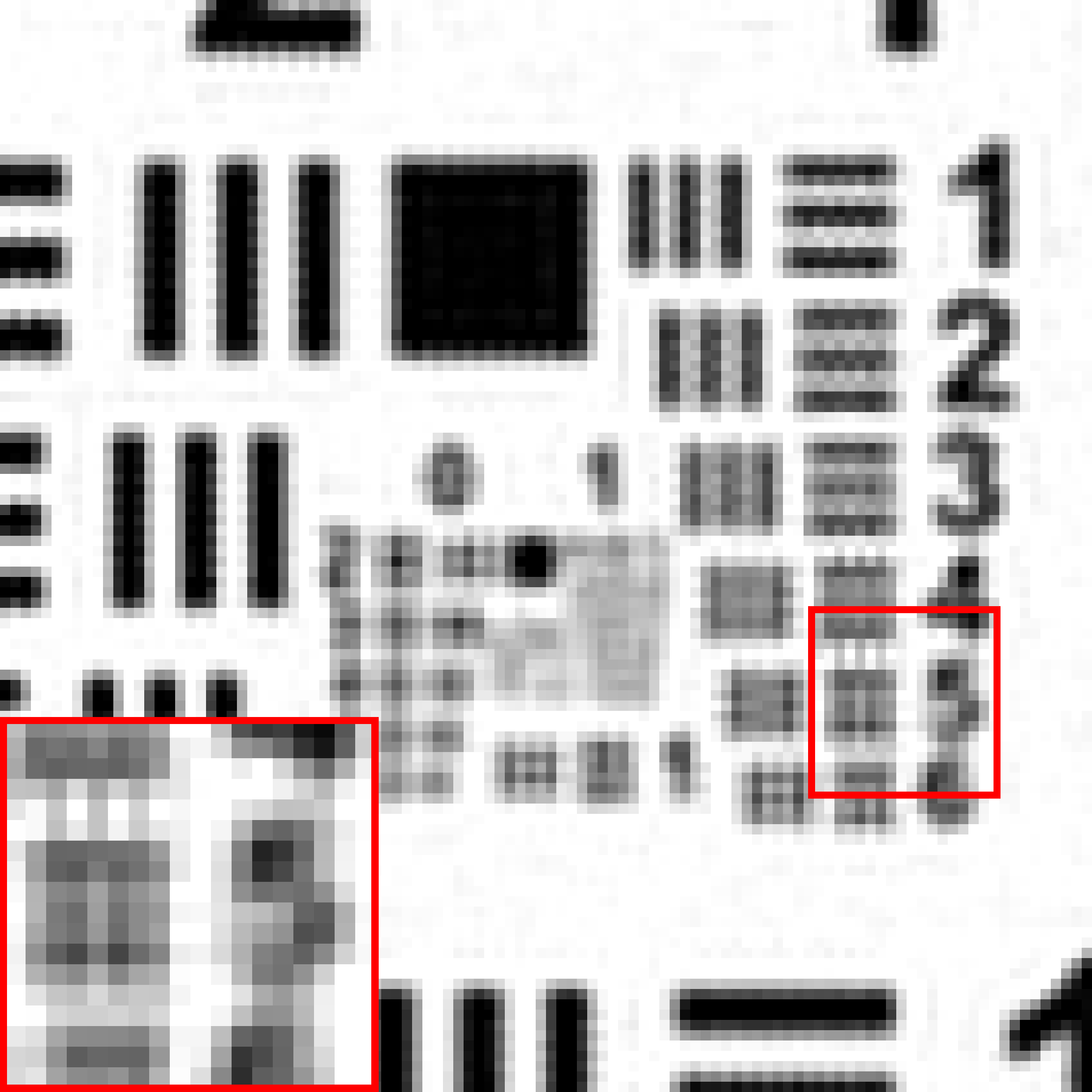}}
  \centerline{(d)}\medskip
\end{minipage}
\begin{minipage}[b]{.16\textwidth}
  \centering
  \centerline{\includegraphics[width=0.9\linewidth]{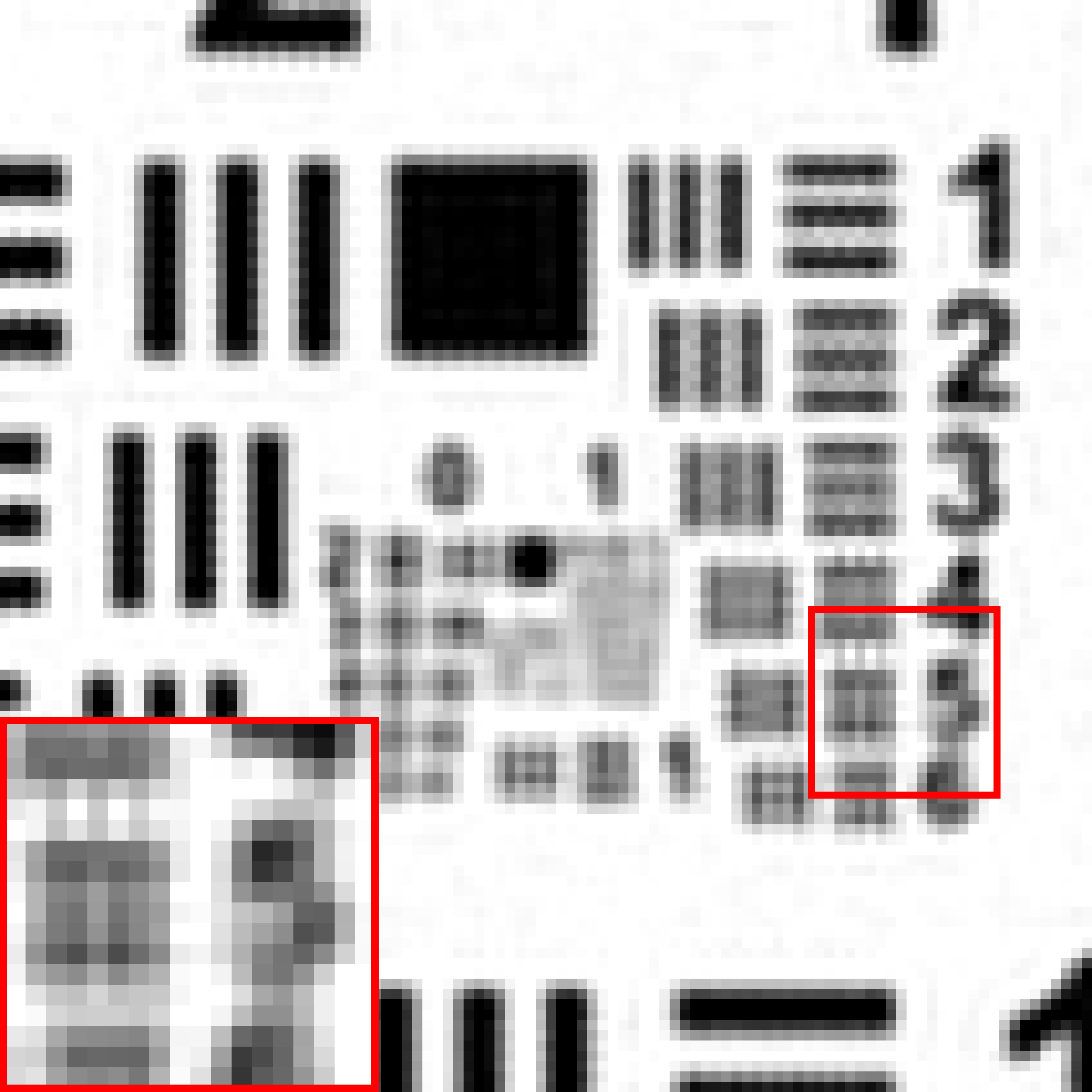}}
  \centerline{(e)}\medskip
\end{minipage}
\hfill
\begin{minipage}[b]{.16\textwidth}
  \centering
  \centerline{\includegraphics[width=0.9\linewidth]{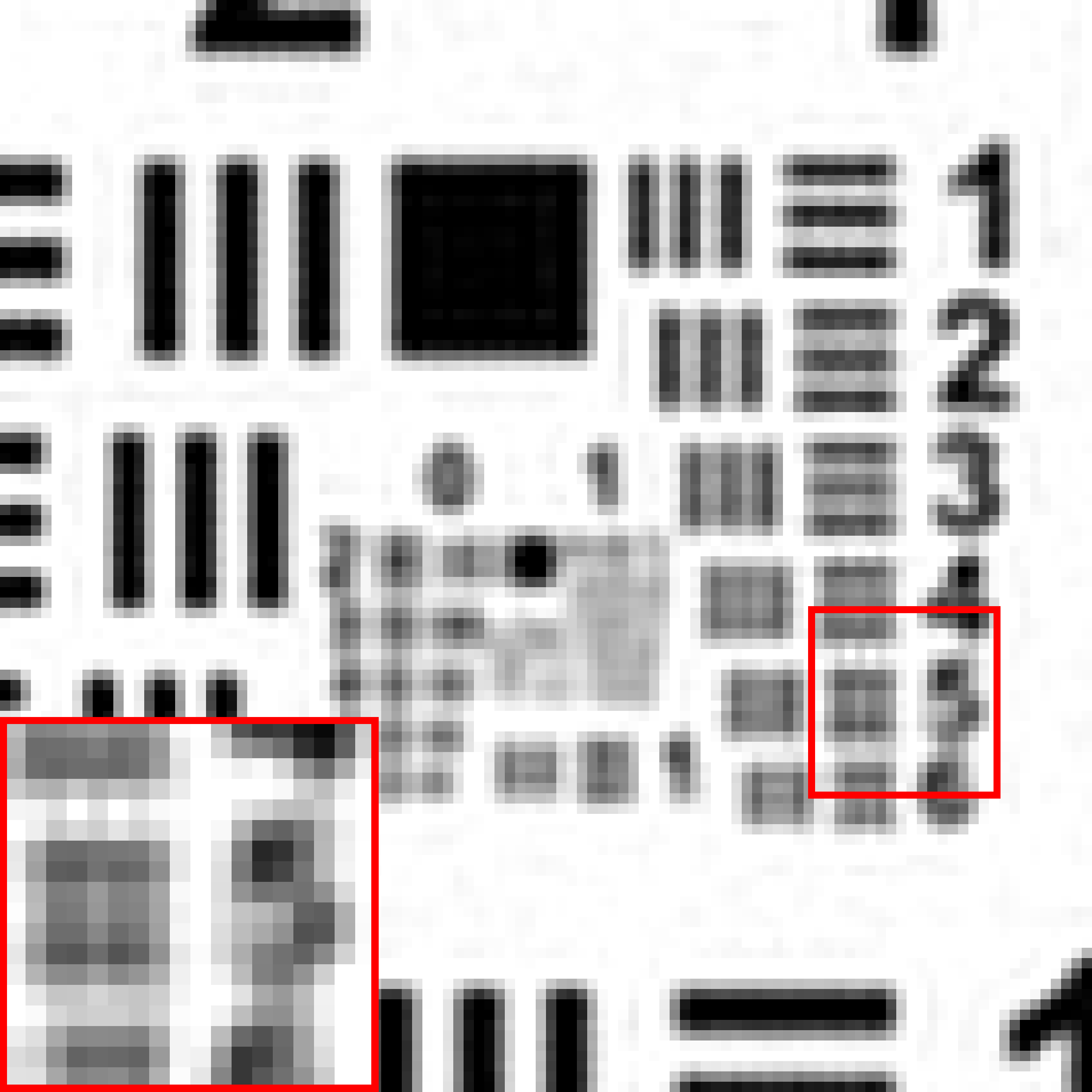}}
  \centerline{(f)}\medskip
\end{minipage}
\vspace{-0.4cm}
\caption{Sample of the 200th reconstructed frame. (a) Original image. (b) Bicubic interpolation (MSE=27.73dB, SSIM=0.846). (c) LMS (MSE=25.39dB, SSIM=0.890). (d) R-LMS algorithm (MSE=25.00dB, SSIM=0.872). (e) TSR-LMS (MSE=25.23dB, SSIM=0.889). (f) LTSR-LMS (MSE=25.06dB, SSIM=0.884).}
\label{fig:resIdeal}
\end{figure*}

For a more realistic evaluation, we repeated the MC simulation considering the influence of registration errors. The \textit{Horn \& Schunck} registration algorithm~\cite{Horn1981determiningOF,Sun10} was employed\footnote{The parameters were set as: \texttt{lambda=1$\times$10$^3$}, \texttt{pyramid\_levels=4}, \texttt{pyramid\_spacing=2}.}, with the velocity fields averaged across the entire image to compute the global displacements. The algorithm parameters were the same used in the previous simulation. The resulting MSE results are depicted in Figure~\ref{fig:MSEreg}, and an example of a reconstructed image of a resolution test chart is shown in Figure~\ref{fig:resIdeal}.

%%%%

It can be verified that the proposed methods still outperform the conventional (R-)LMS algorithms, though by a smaller margin due to the effect of registration errors.
It should be noted that large levels of registration errors tend to reduce the effectiveness of the TSR-LMS and LTSR-LMS methods, as we are basically preserving information (details) from previous frames that must be registered.
Furthermore, the performance of the TSR-LMS algorithm showed greater sensitivity to unknown registration, as its performance degraded more when compared to the LTSR-LMS algorithm.
The images reconstructed by the four evaluated algorithms were found to be perceptually similar, although a careful inspection reveals a slight improvement in the reconstruction result using the LTSR-LMS algorithm.
Nevertheless, the following examples will illustrate that the proposed methods perform considerably better than the others in the presence of outliers.

% ------------------------------------------------------------------
\subsection{Example 2} \label{sec:examples_synth_ii}

This example evaluates the robustness to innovation outliers. This was done by super-resolving synthetic video sequences containing a suddenly appearing object, which is independent of the background. The first MC simulation of \textit{Example 1} was repeated, this time with the inclusion of an $N\times N$ black square appearing in the middle of the~32nd frame and disappearing in the~35th frame of every sequence, thus emulating the behavior of a \textit{flying bird} outlier on the scene.

\begin{figure}[t]
\centering
\begin{minipage}[b]{0.45\linewidth}
  \centering
  \centerline{\includegraphics[width=\linewidth]{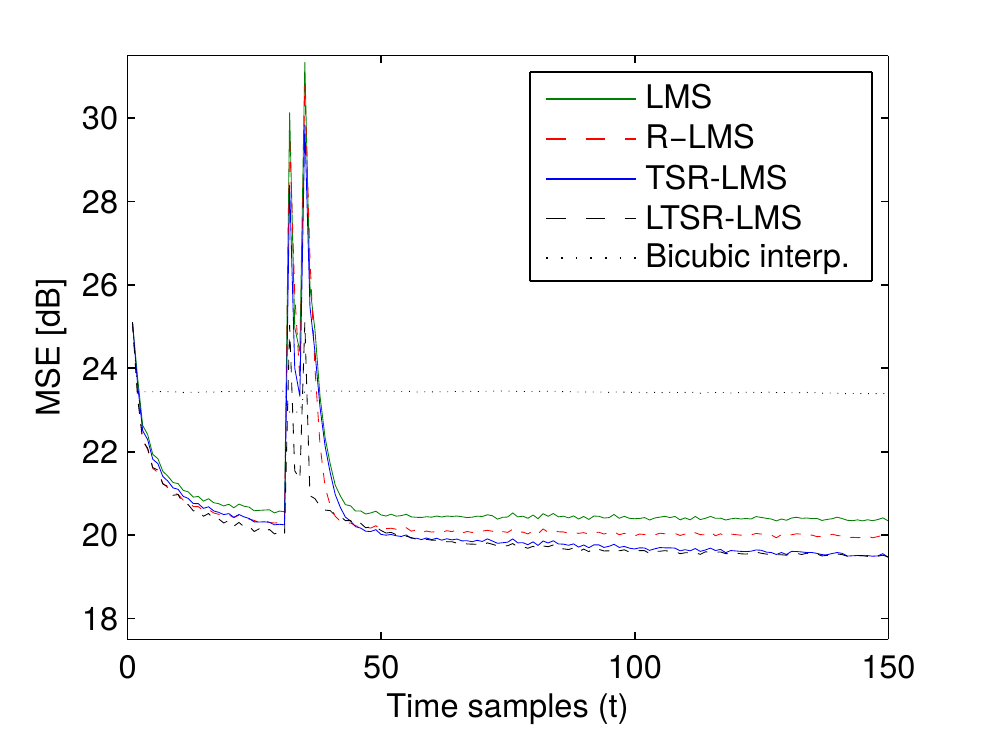}}
  \centerline{(a)}%\medskip
\end{minipage}
\hspace{0.3cm}
\begin{minipage}[b]{0.45\linewidth}
  \centering
  \centerline{\includegraphics[width=\linewidth]{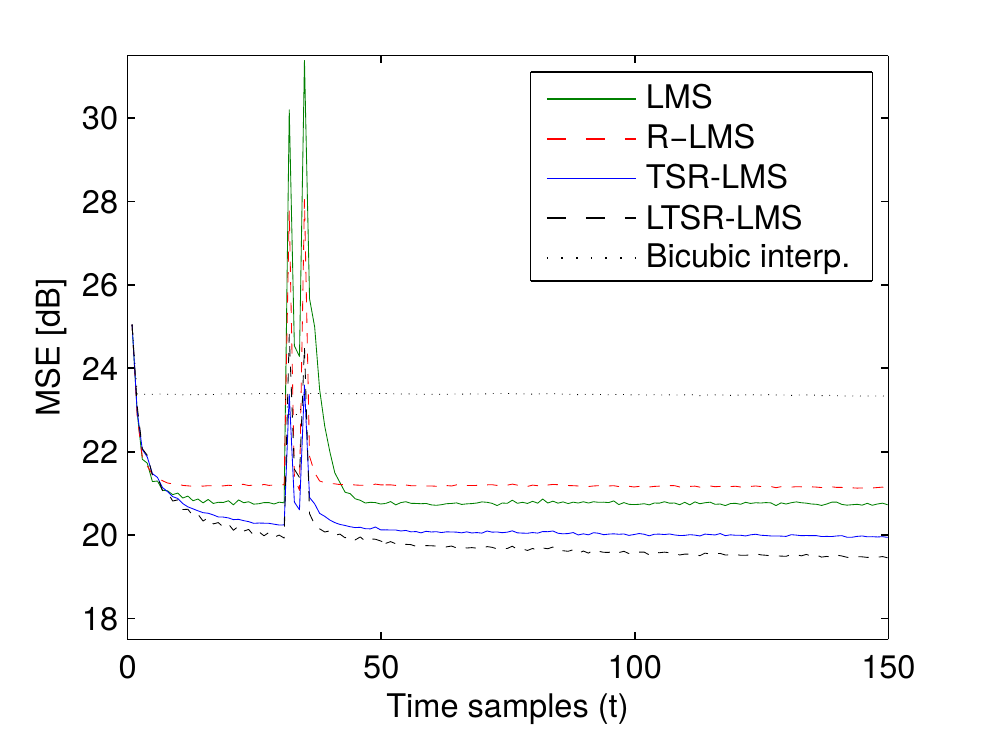}}
  \centerline{(b)}%\medskip
\end{minipage}
\begin{minipage}[b]{0.45\linewidth}
  \centering
  \centerline{\includegraphics[width=\linewidth]{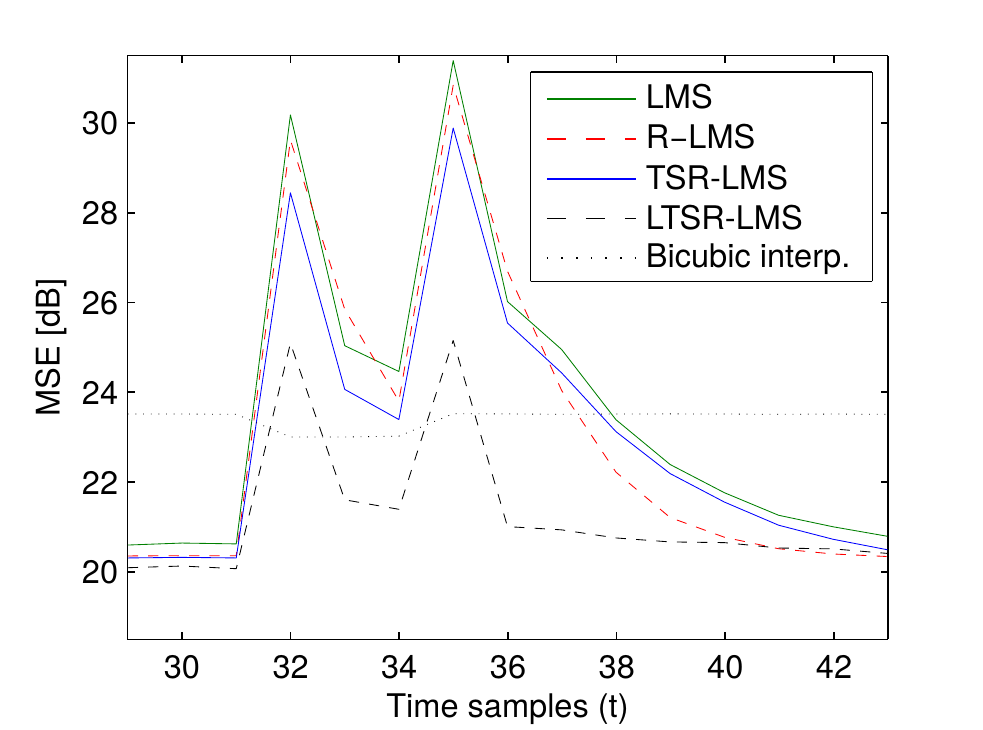}}
  \centerline{(c)}%\medskip
\end{minipage}
\hspace{0.3cm}
\begin{minipage}[b]{0.45\linewidth}
  \centering
  \centerline{\includegraphics[width=\linewidth]{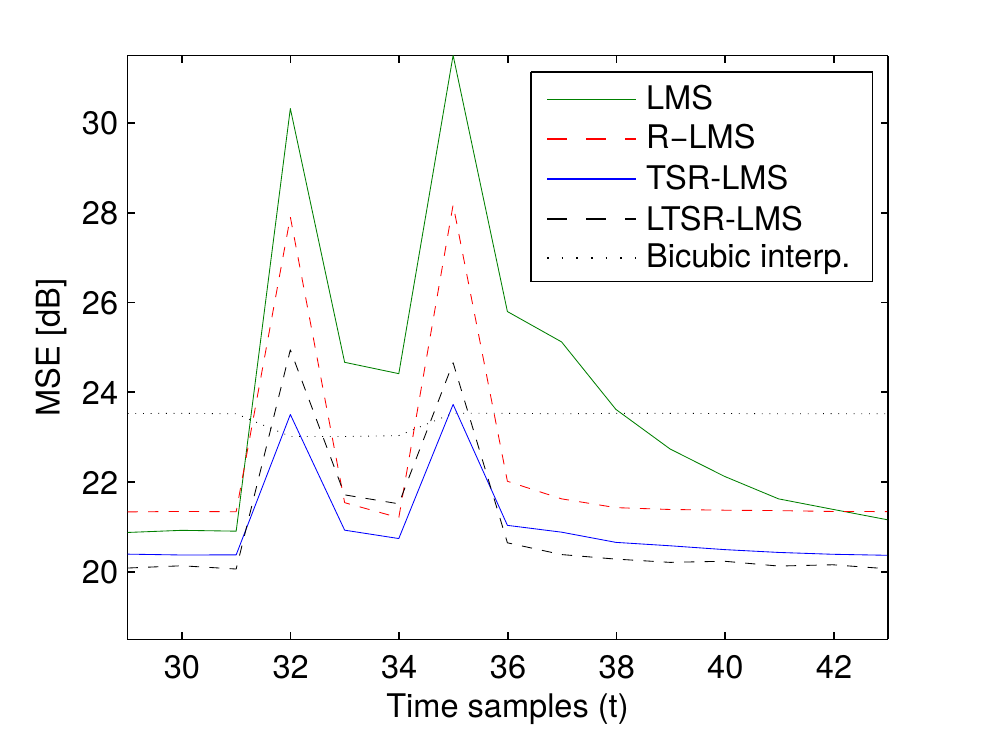}}
  \centerline{(d)}%\medskip
\end{minipage}
\vspace{-0.1cm}
\caption{MSE per pixel with an outlier at frames~32-35. (a) and (c): Full sequence and zoom with for reconstruction with parameters of Table~\ref{tab:parametersKnown}. (b) and (d): Full sequence and zoom with for reconstruction with parameters of Table~~\ref{tab:parametersOutlier}.}
\label{fig:MSEbird}
\end{figure}

The MSE evolution for the parameters shown in Table~\ref{tab:parametersKnown} is shown in Figures~\ref{fig:MSEbird}-(a) and~\ref{fig:MSEbird}-(c) (zoomed). The improvement provided by the LTSR-LMS algorithm is clearly significant, suggesting its greater robustness to outliers.

%The simulation is first performed for the set of parameters shown in Table~\ref{tab:parametersKnown}, which have been used in the previous example. The MSE evolution is depicted in Figure~\ref{fig:MSEbird}-(a) and~\ref{fig:MSEbird}-(c). 
%
%It can be seen that the estimated solutions deviate significantly from the desired image in the presence of an outlier. Comparing the different algorithms, it can be noticed that the method proposed in~\eqref{eq:new_lmslp_iterative_1} offer a slight improvement in comparison with the LMS and R-LMS algorithms, whereas the improvement achieved by the method presented in~\eqref{eq:new_lmslp_iterative_2} is much more significant.
%
% This comes due to the fact that the parameters portrayed in Table~\ref{tab:parametersKnown}, although allowing the algorithms to achieve their best steady-state MSE (i.e. large $t$ and small innovations), do not necessarily result in a good performance in the presence of outliers (e.g. the MSE in the frames with large innovations).
%
%These results suggest that the algorithm~\eqref{eq:new_lmslp_iterative_2} is the most robust to outliers among all tested algorithms.

To improve the design of the proposed algorithms, we performed exhaustive searches in the parameter space of all algorithms to determine good sets of parameters for reconstructing a small independent set of images. The parameters in Table~\ref{tab:parametersOutlier} yielded the minimum MSE averaged between frames $30$ and $40$ for each algorithm. The MSE evolutions for these parameters are shown in Figures~\ref{fig:MSEbird}-(b) and~\ref{fig:MSEbird}-(d).

%A performance improvement may be achieved at the cost of some loss in steady-state performance. To illustrate this point, we have determined an alternative set of parameters to obtain the minimum MSE averaged between frames $30$ and $40$. We did this by performing an exhaustive search for reconstructing a small independent set of images. The resulting parameters are presented in Table~\ref{tab:parametersOutlier}. The MSE evolution is depicted in Figures~\ref{fig:MSEbird}-(b) and~\ref{fig:MSEbird}-(d).

\begin{table}
% \small
\footnotesize
% \scriptsize
\caption{Parameter values used on the simulations considering the presence of outliers.}
\vspace{-0.2cm}
\centering
% \begin{center}
\renewcommand{\arraystretch}{1.2}
\begin{tabular}{lcccc}
\hline
& \textbf{LMS} & \textbf{R-LMS} & \textbf{TSR-LMS} & \textbf{LTSR-LMS}  \\
\hline
$\mu$      &  4.7  &  4.2  &  2.2  &  3.4  \\
$\alpha$   &  --  &  40$\times10^{-4}$  &  18$\times10^{-4}$  &  1$\times10^{-4}$\\
$\alpha_{\text{T}}$ &  --  &  --  &  16  &  0.017  \\
\hline
\end{tabular}
% \end{center}
\label{tab:parametersOutlier}
\end{table}

% \begin{figure}[t]
% \centering
% \begin{minipage}[b]{1\linewidth}
%   \centering
%   \centerline{\includegraphics[width=10cm]{figures/MSE_Bird_3x3_KnownMotion_outleirparams_full}}
%   \centerline{(a)}\medskip
% \end{minipage}
% %
% \begin{minipage}[b]{1\linewidth}
%   \centering
%   \centerline{\includegraphics[width=10cm]{figures/MSE_Bird_3x3_KnownMotion_outlierparams_crop}}
%   \centerline{(b)}\medskip
% \end{minipage}
% %
% \caption{Average MSE per pixel with an outlier at frame 32. (a) Full sequence. (b) Zoom into region of interest.}
% \label{fig:MSEbird}
% \end{figure}
%

The proposed methods led to a significant performance gain compared to the other algorithms when in the presence of outliers in frames~32 and~35, with a slightly better results verified for the TSR-LMS method.

%It can be noticed that both proposed methods provide a significant performance gain when compared to the remaining algorithms in the presence of outliers in frames~32 and~35, where the method of~\eqref{eq:new_lmslp_iterative_1} performed slightly better than that of~\eqref{eq:new_lmslp_iterative_2}.
%

\begin{figure*}[htb]
\begin{minipage}[b]{.16\textwidth}
  \centering
  \centerline{\includegraphics[width=0.9\linewidth]{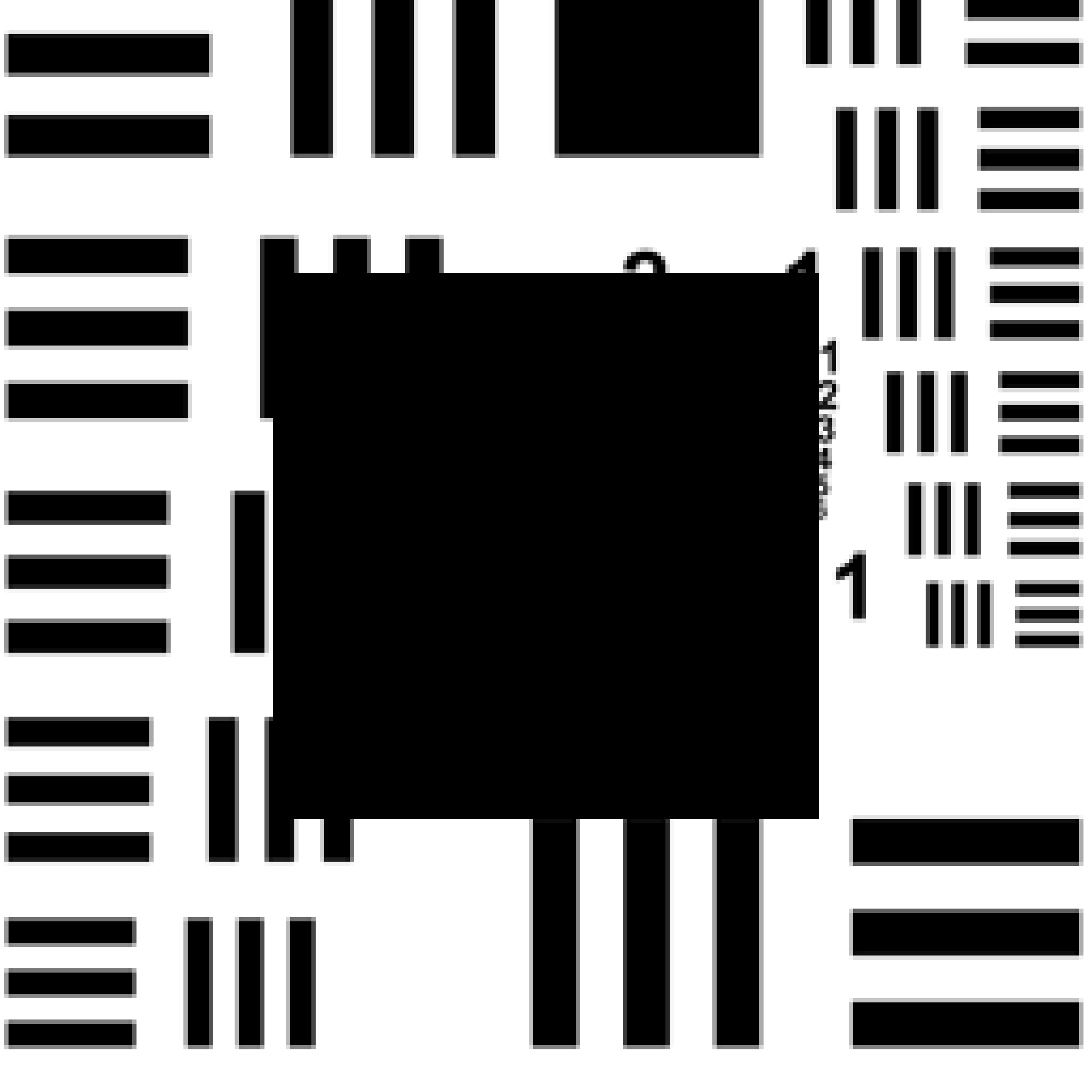}}
  \centerline{(a)}\smallskip
\end{minipage}
%\hfill
\begin{minipage}[b]{.16\textwidth}
  \centering
  \centerline{\includegraphics[width=0.9\linewidth]{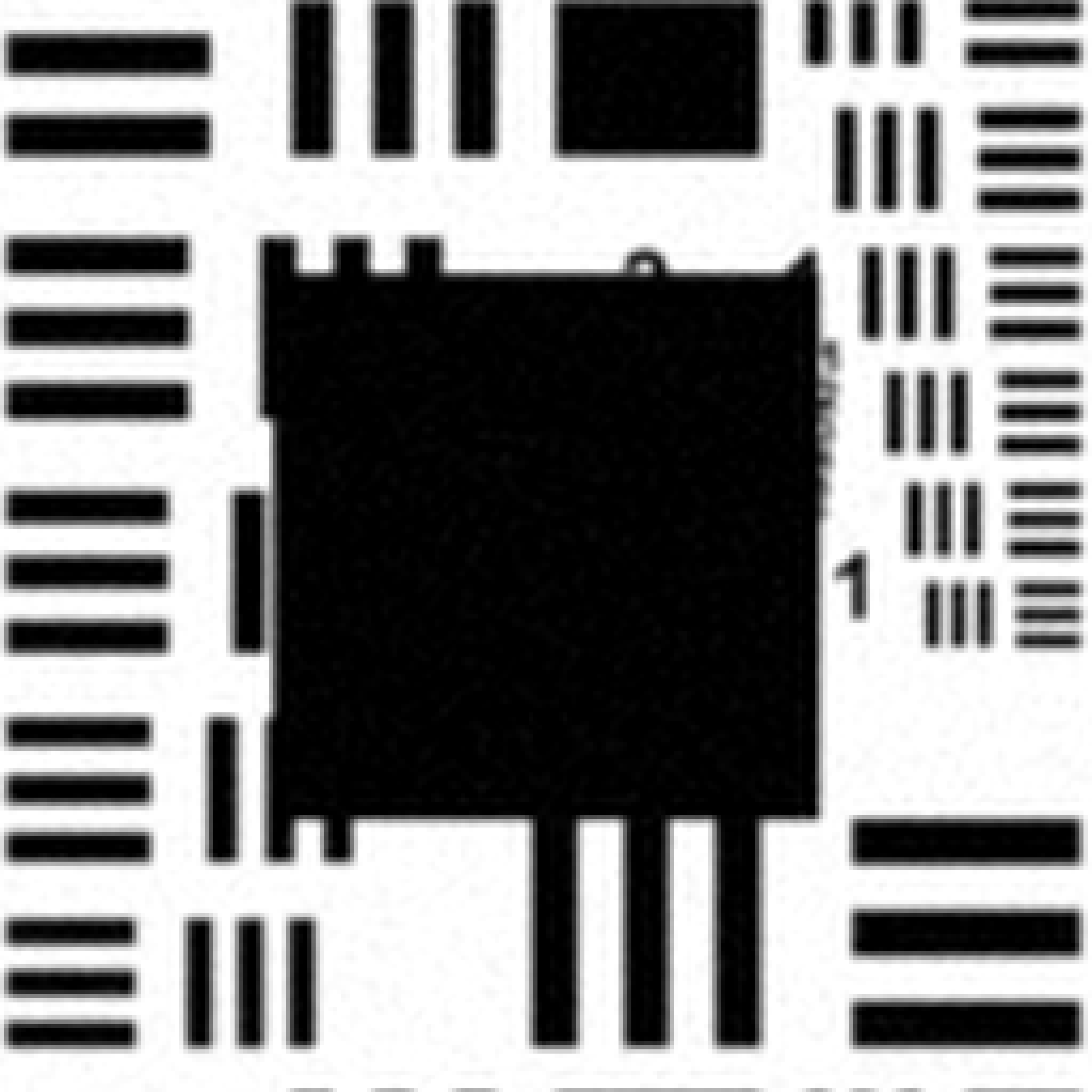}}
  \centerline{(b)}\smallskip
\end{minipage}
\begin{minipage}[b]{.16\textwidth}
  \centering
  \centerline{\includegraphics[width=0.9\linewidth]{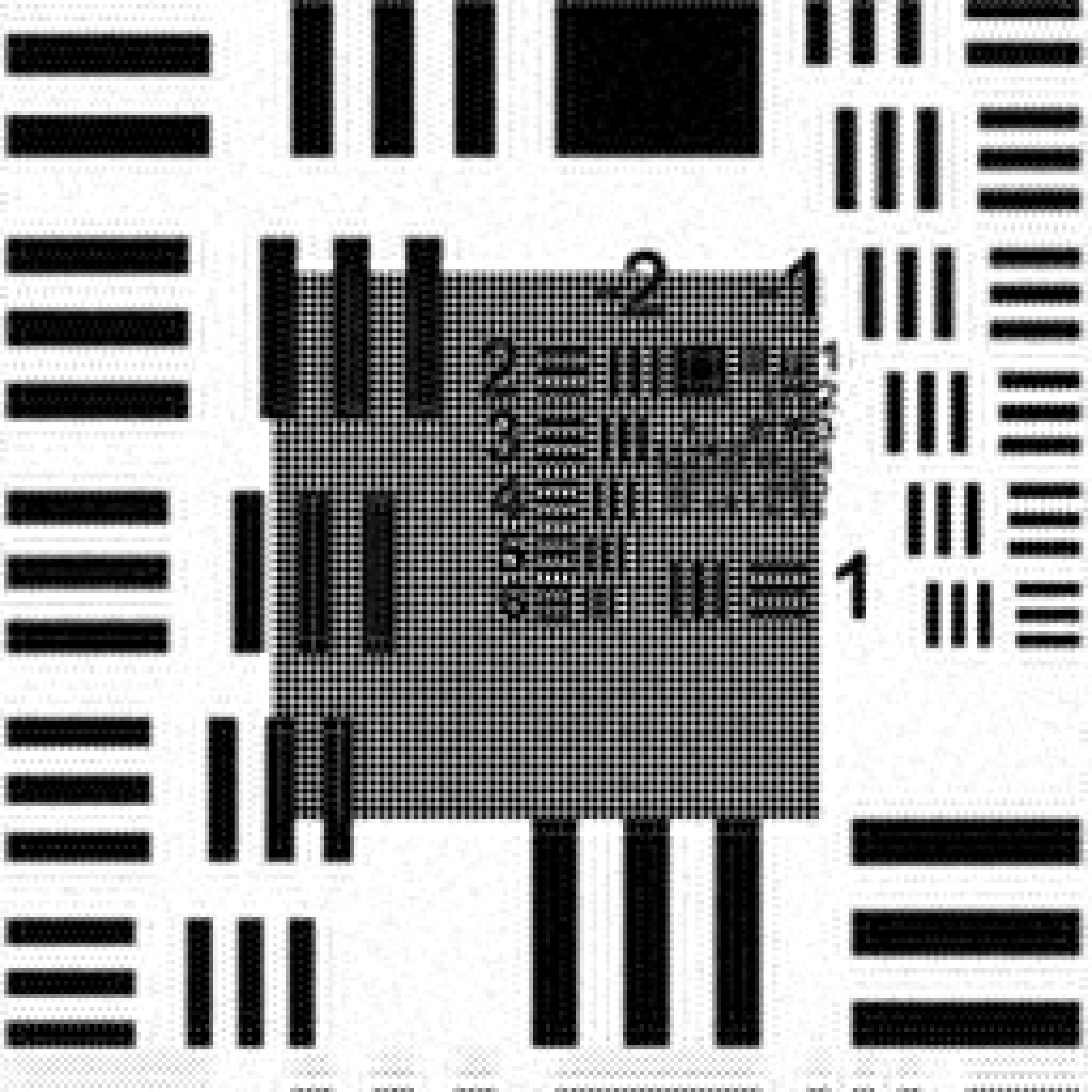}}
  \centerline{(c)}\smallskip
\end{minipage}
%\hfill
%
\begin{minipage}[b]{.16\textwidth}
  \centering
  \centerline{\includegraphics[width=0.9\linewidth]{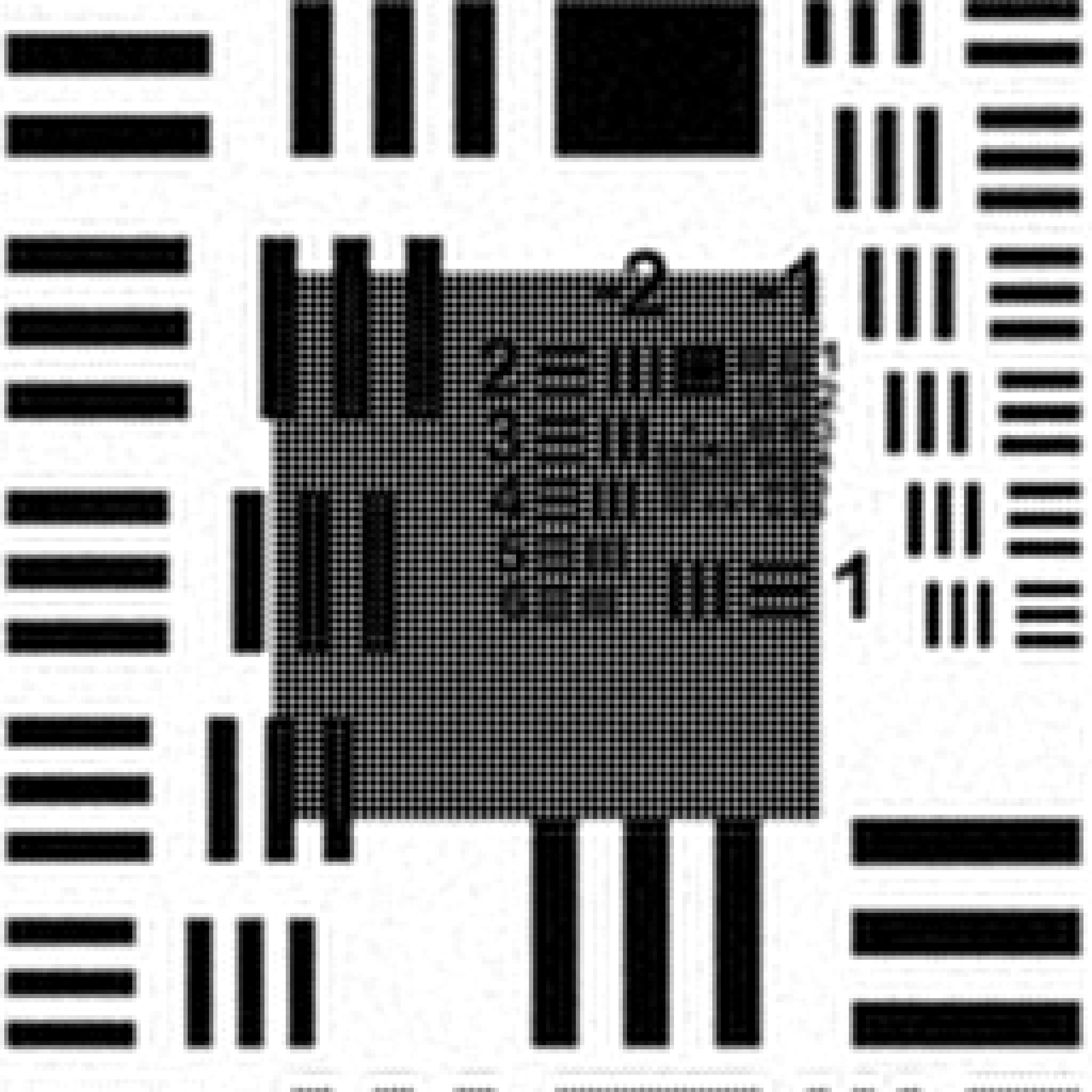}}
  \centerline{(d)}\smallskip
\end{minipage}
\begin{minipage}[b]{.16\textwidth}
  \centering
  \centerline{\includegraphics[width=0.9\linewidth]{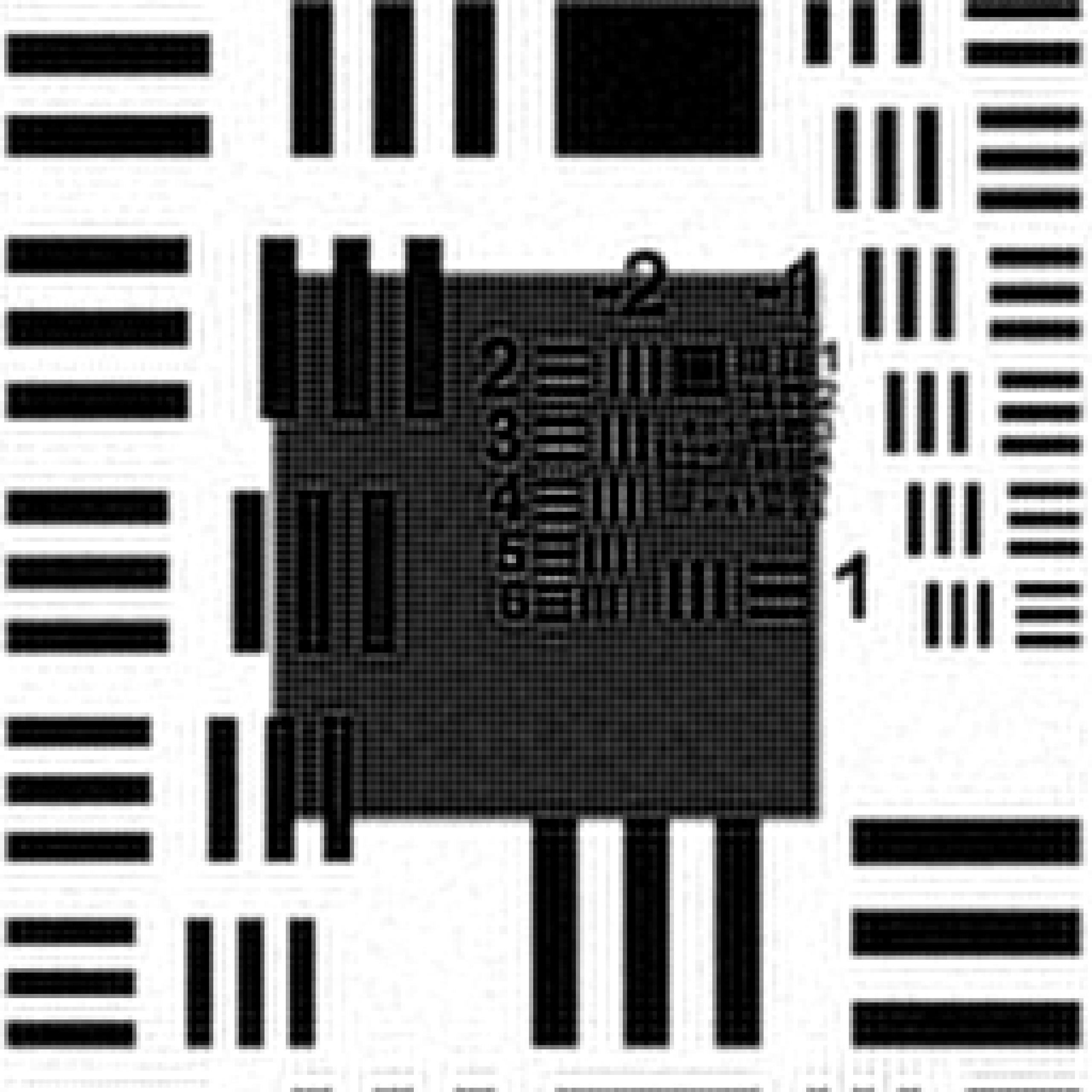}}
  \centerline{(e)}\smallskip
\end{minipage}
\hfill
\begin{minipage}[b]{.16\textwidth}
  \centering
  \centerline{\includegraphics[width=0.9\linewidth]{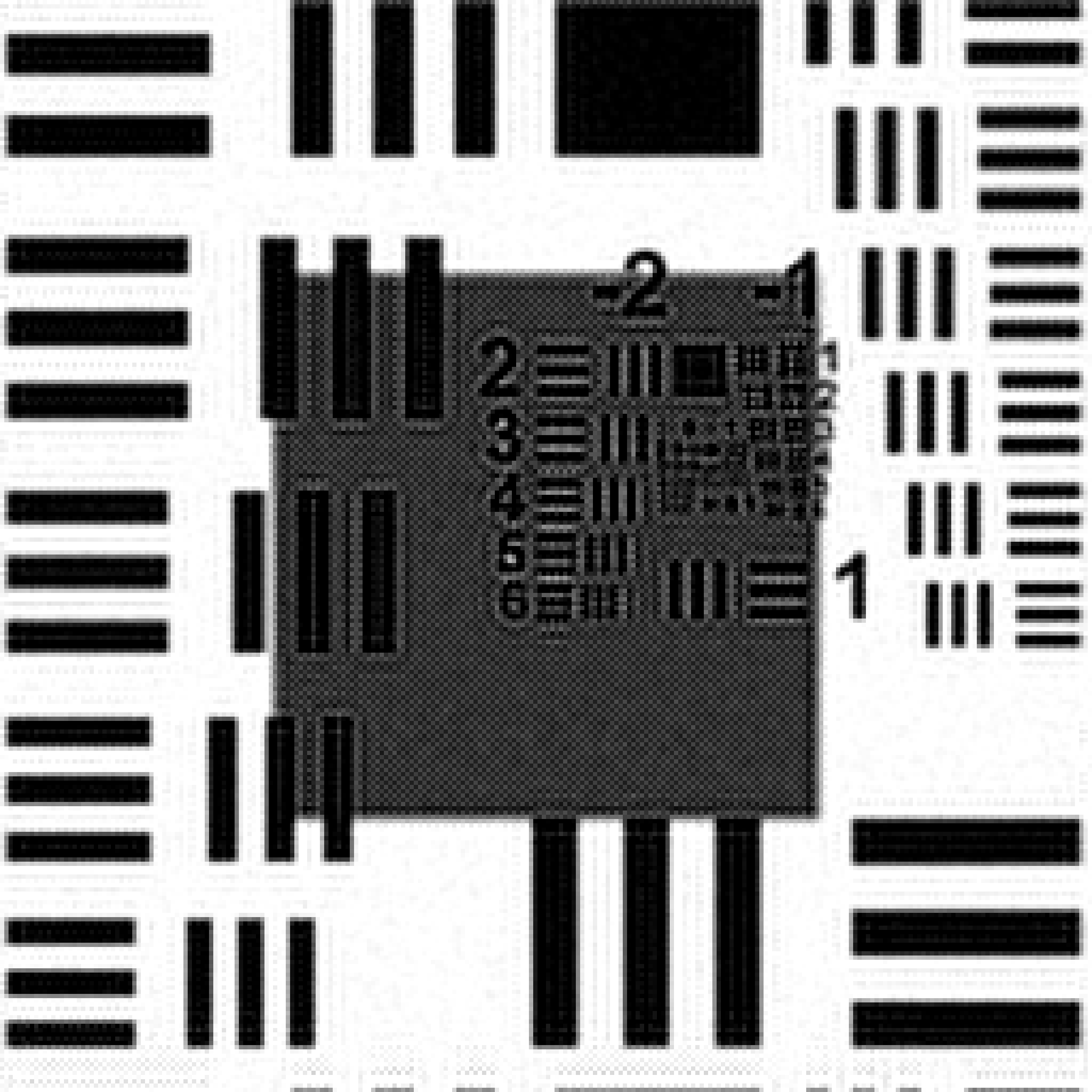}}
  \centerline{(f)}\smallskip
\end{minipage}
\vspace{-0.4cm}
\caption{Sample of 32th frame of a reconstructed sequence (the black square is present in the desired image). (a) Original image. (b) Bicubic interpolation (MSE=27.31dB, SSIM=0.826). (c) LMS (MSE=34.83dB, SSIM=0.595). (d) R-LMS algorithm (MSE=32.35dB, SSIM=0.649). (e) TSR-LMS (MSE=28.17dB, SSIM=0.649). (f) LTSR-LMS (MSE=29.71dB, SSIM=0.639).}
\label{fig:resBird}
\end{figure*}

While R-LMS led to a MSE similar to that achieved by the TSR-LMS and LTSR-LMS algorithms for frames~33 and~34 (when the black square remained in the scene), its performance degraded considerably for larger $t$. Note also that the loss in steady-state performance as a result of the optimization to handle outliers was higher for LMS and R-LMS.  Finally, the LTSR-LMS algorithm showed to be less sensitive to the parameter selection, performing reasonably well in both simulations. Figure~\ref{fig:resBird} shows the reconstructed images for frame~32, which confirm the quantitative results. The black square introduced in the sequence is significantly better represented for the proposed methods (when it is indeed present in the HR image), and a slight improvement can be noticed in the result of the TSR-LMS algorithm when compared to that of the LTSR-LMS.

%The foreman sequence contains a monologue of a man mo ving his head dynamically and t the end of the sequence there is a contiguous scene change.
% showing a closeup of a talking man followed by a camera panning and view of a building being constructed

% ------------------------------------------------------------------
\subsection{Example 3}
\label{sec:examples_real_i}

This example illustrates the effectiveness of the proposed methods when super-resolving real video sequences.
We performed a Monte Carlo simulation with 15 realizations consisting of natural video sequences, like \textit{Foreman}, \textit{Harbour}, \textit{News} and others.
In this case, the true motions of the objects and camera are unknown. We used \textit{Horn \& Shunck} algorithm with the same parameters shown in Table~\ref{tab:parametersOutlier} to estimate the dense velocity field, but now considering the displacement to be unique for each image pixel.

For a quantitative evaluation, the original videos were used as available HR image sequences. For simplicity, only the $256\times256$ upper-right region of the original sequence was considered so that the resulting images were square. Like in \textit{Example 1}, the HR sequences were blurred with an uniform unitary gain $3\times3$ mask, decimated by a factor of $2$ {and contaminated with white Gaussian noise with variance $\sigma^2=10$ to form} the LR images. The standard LMS versions and the proposed methods were used to super-resolve the LR sequences with $K=2$ iterations per time sample and the parameters set at the values in Table~\ref{tab:parametersOutlier}. Hence, they were not guaranteed to be optimal, as the amount of innovations is different and the motion is not known in advance.
%
% It is important to notice that, since this example contains a significant level of innovations, the parameters were chosen based on the context of \textit{Example 2}. Nevertheless, they were not guaranteed to be optimal, as we were working with a single video sequence, with unknown motion and in the presence of registration errors.
%
%It is important to notice that, since the videos considered in this example contains a significant level of innovations, the parameters were chosen based on the context of \textit{Example 2}. Nevertheless, they were not guaranteed to be optimal, since the amount of innovations and is different and the motion is not known in advance.

\begin{figure}
\centering
\includegraphics[width=6cm,trim={0 0 0 0.5cm},clip]{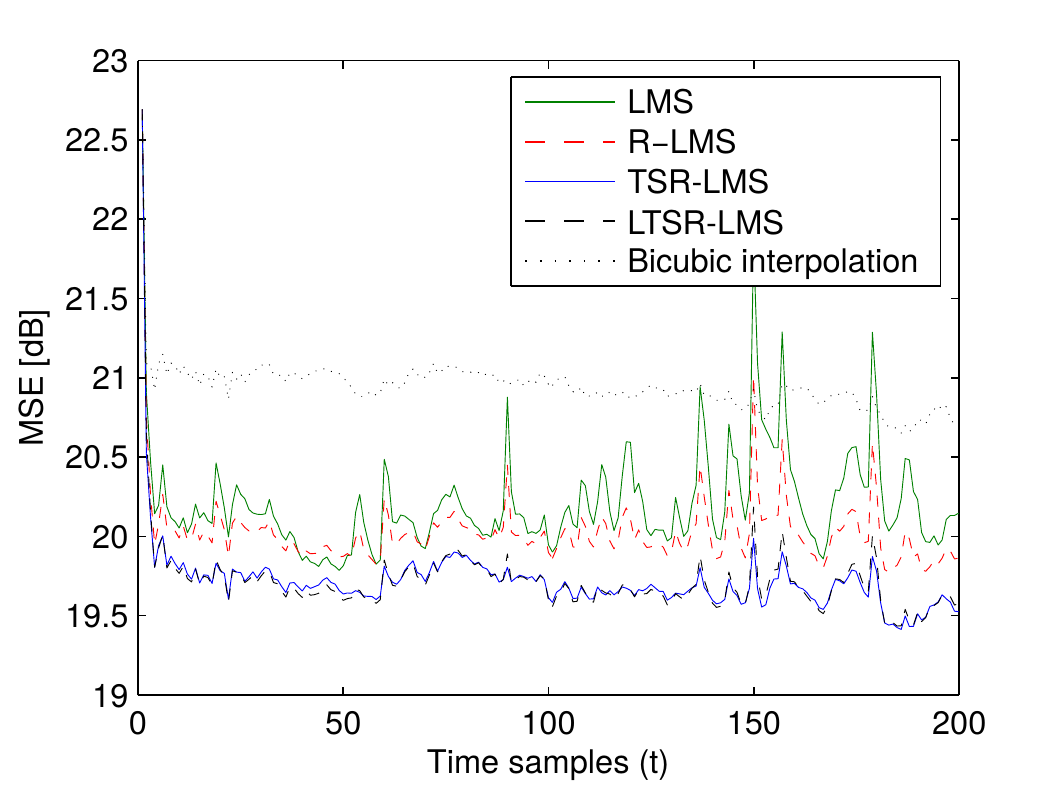}
\vspace{-0.25cm}
\caption{Average MSE per pixel for natural video sequences.}
\label{fig:MSEreal}
\end{figure}

\begin{figure}[htb]
\begin{minipage}[b]{.49\linewidth}
  \centering
  \centerline{\includegraphics[width=0.9\linewidth]{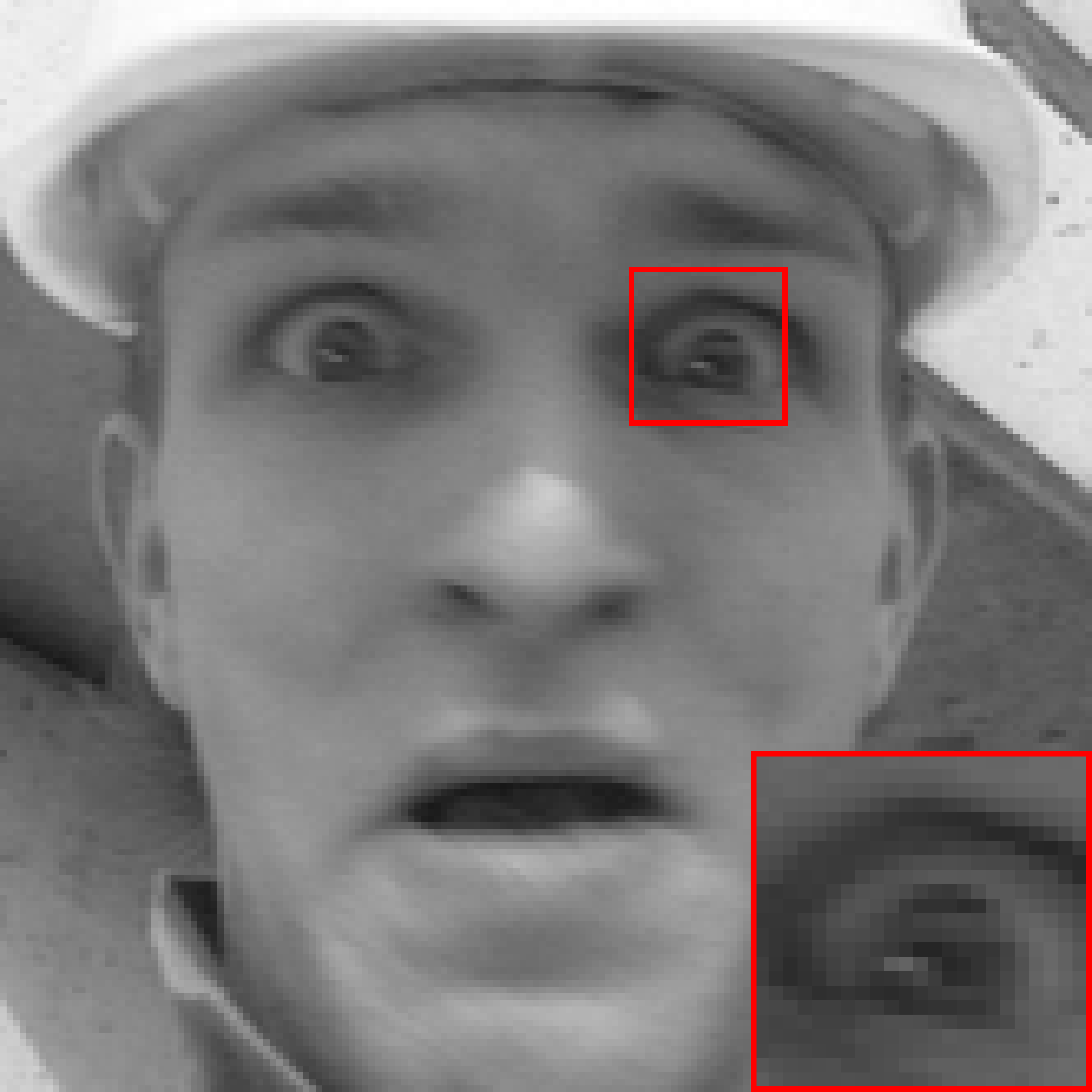}}
  \centerline{(a)}\smallskip
\end{minipage}
%\hfill
\begin{minipage}[b]{0.49\linewidth}
  \centering
  \centerline{\includegraphics[width=0.9\linewidth]{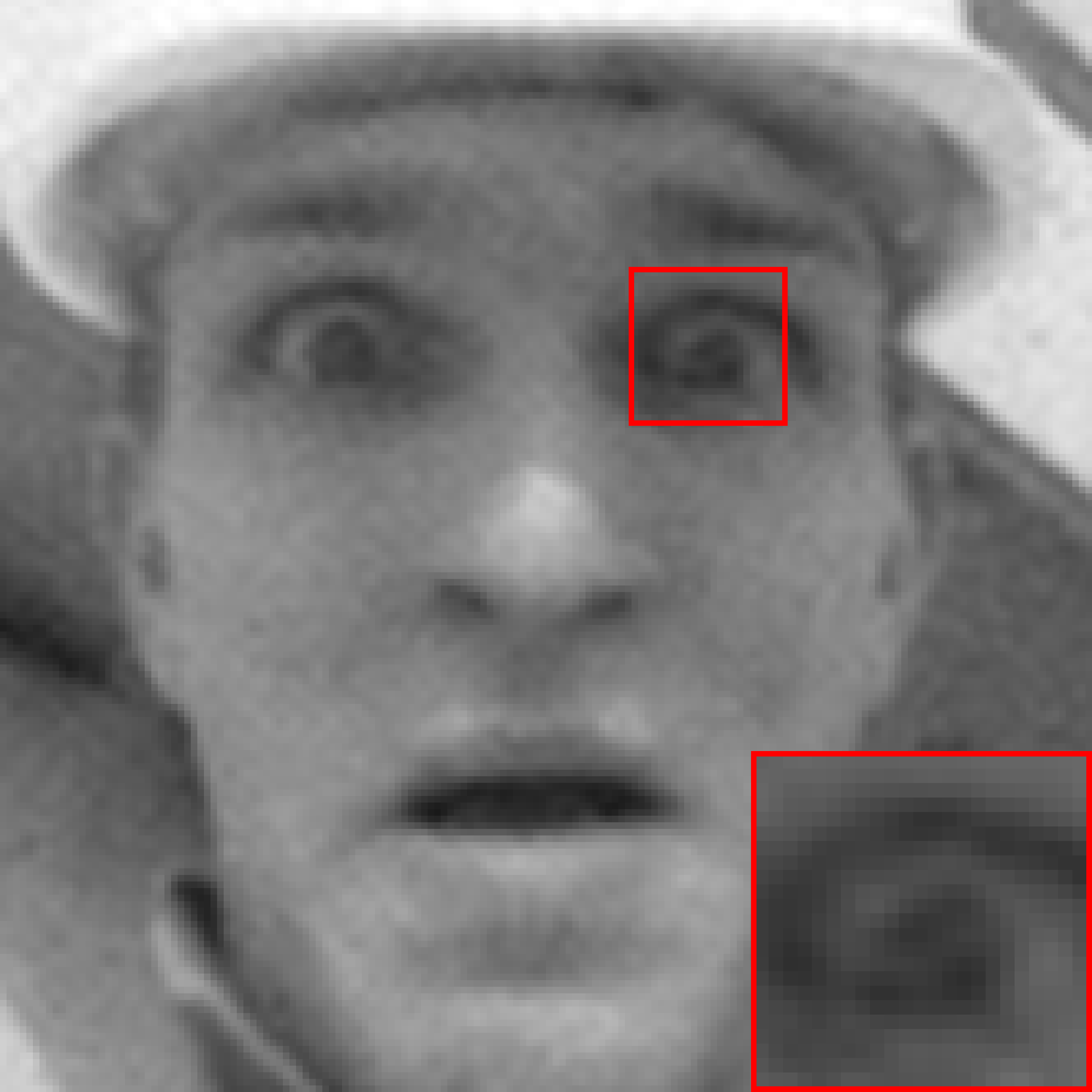}}
  \centerline{(b)}\smallskip
\end{minipage}
\begin{minipage}[b]{.49\linewidth}
  \centering
  \centerline{\includegraphics[width=0.9\linewidth]{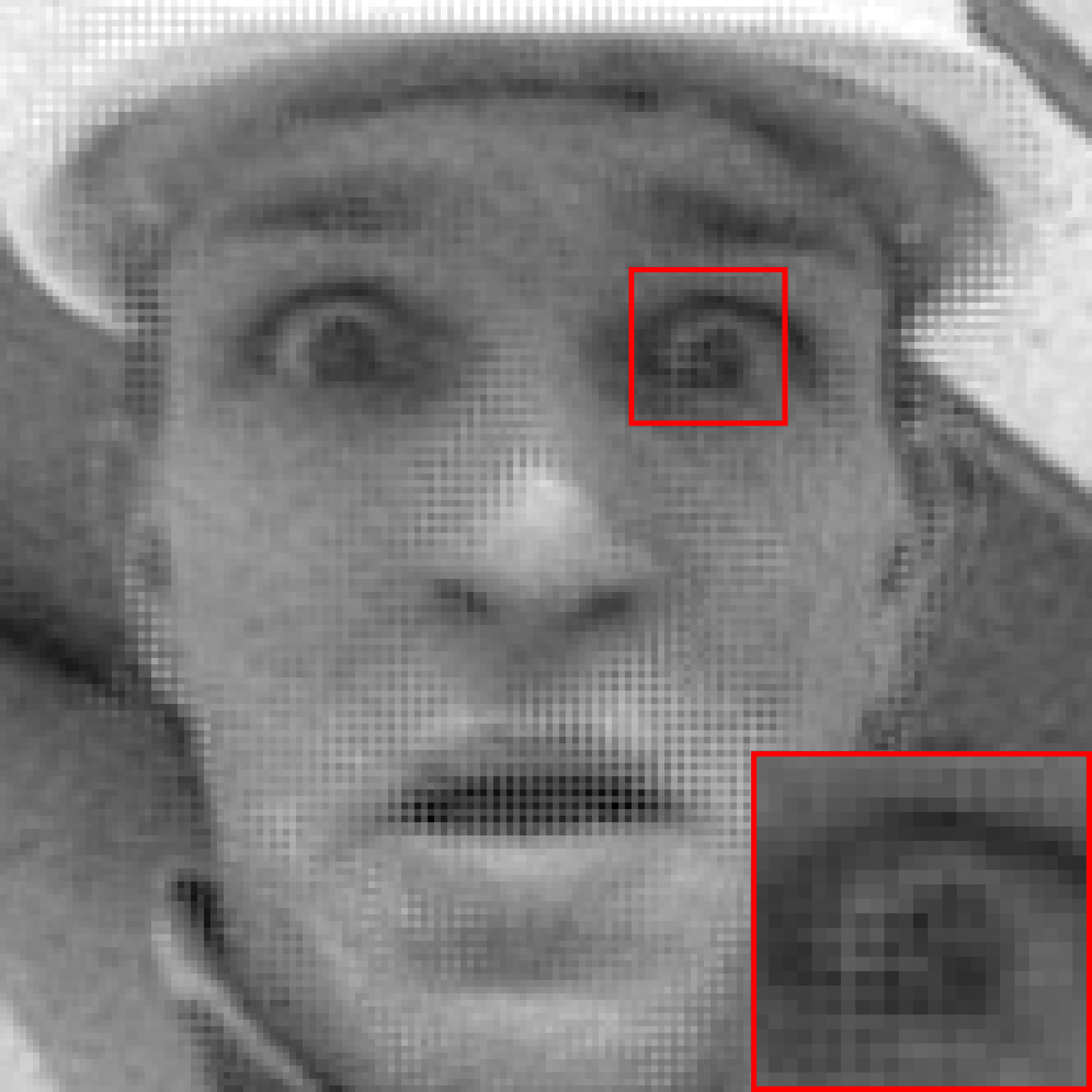}}
  \centerline{(c)}\smallskip
\end{minipage}
%\hfill
\begin{minipage}[b]{0.49\linewidth}
  \centering
  \centerline{\includegraphics[width=0.9\linewidth]{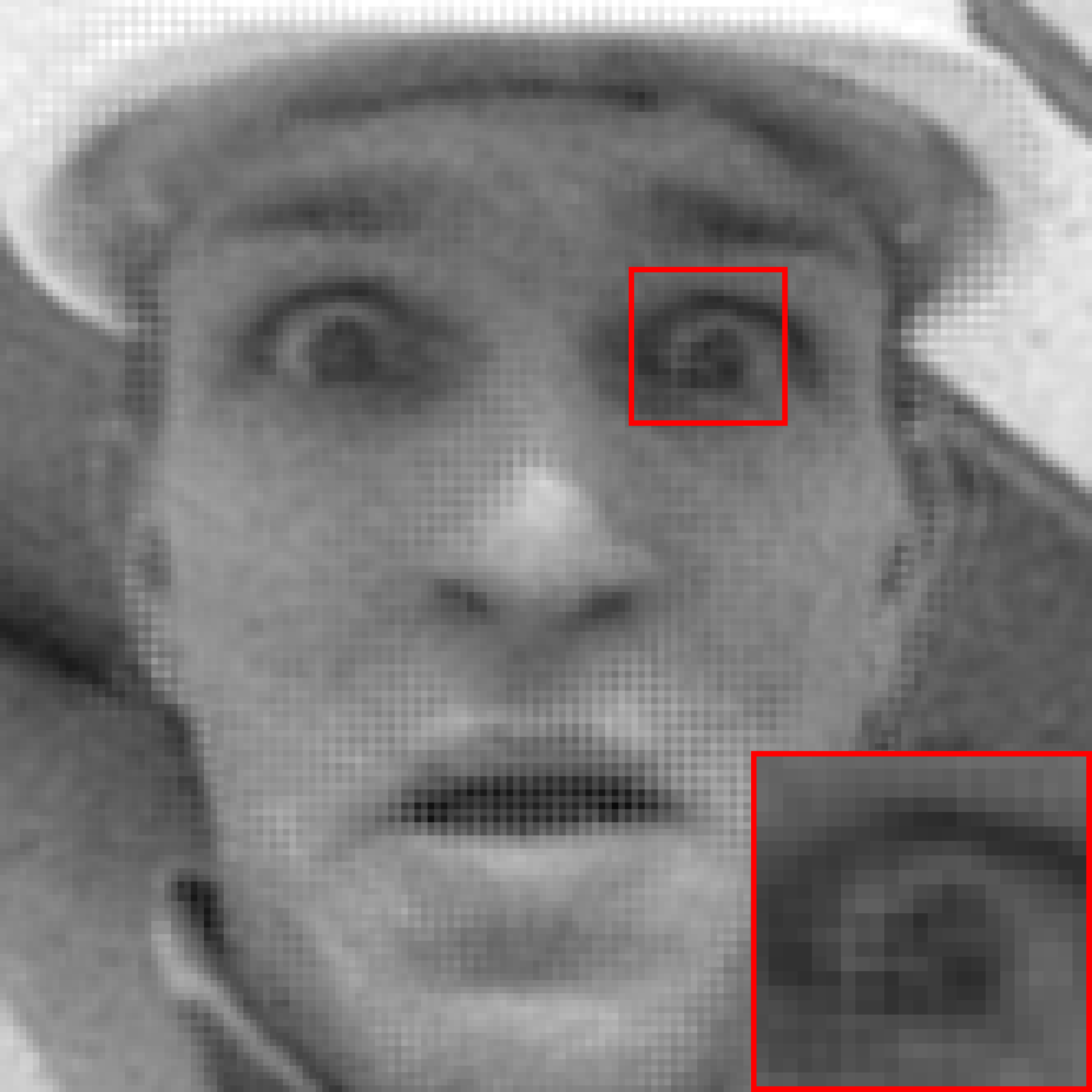}}
  \centerline{(d)}\smallskip
\end{minipage}
\begin{minipage}[b]{.49\linewidth}
  \centering
  \centerline{\includegraphics[width=0.9\linewidth]{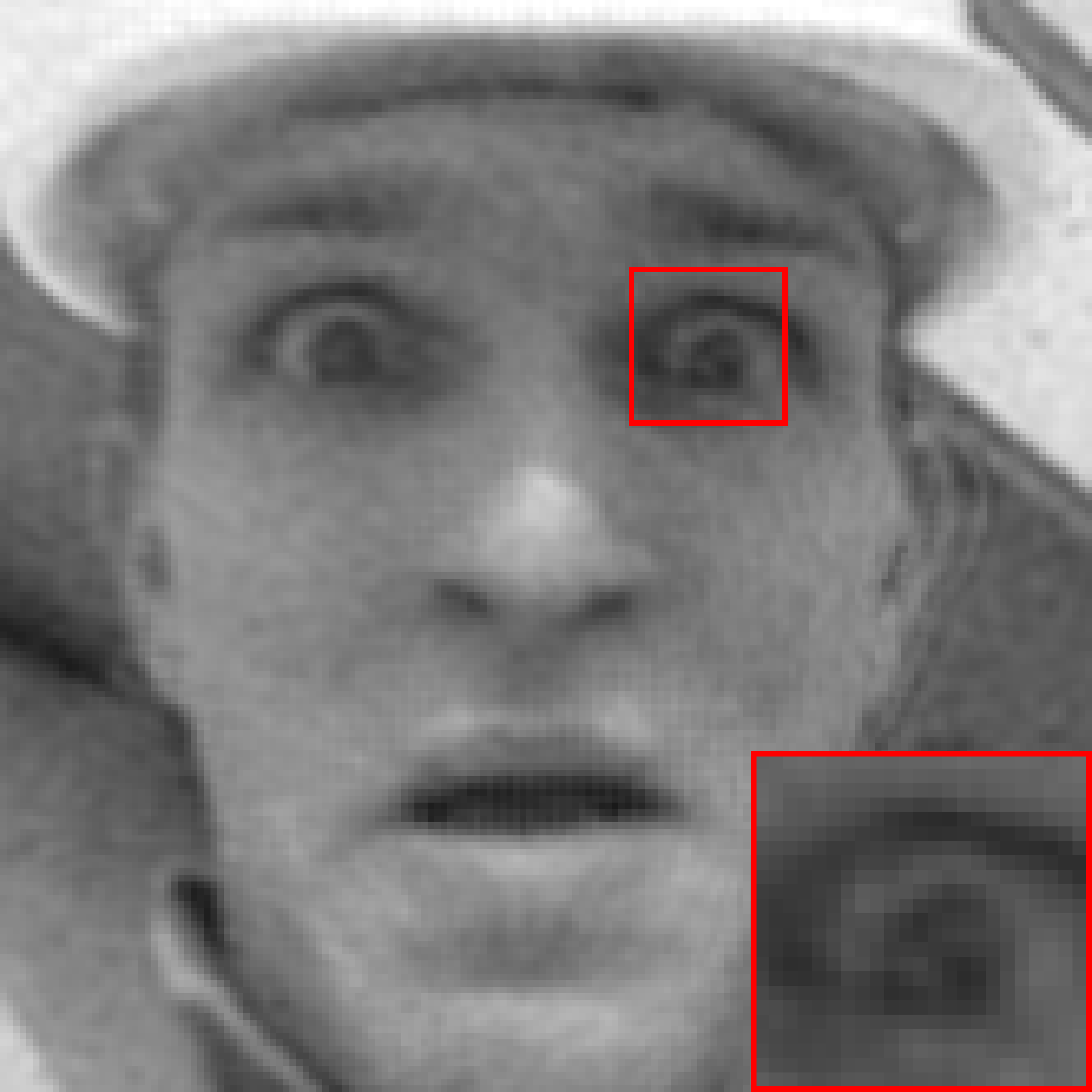}}
  \centerline{(e)}\smallskip
\end{minipage}
\hfill
\begin{minipage}[b]{0.49\linewidth}
  \centering
  \centerline{\includegraphics[width=0.9\linewidth]{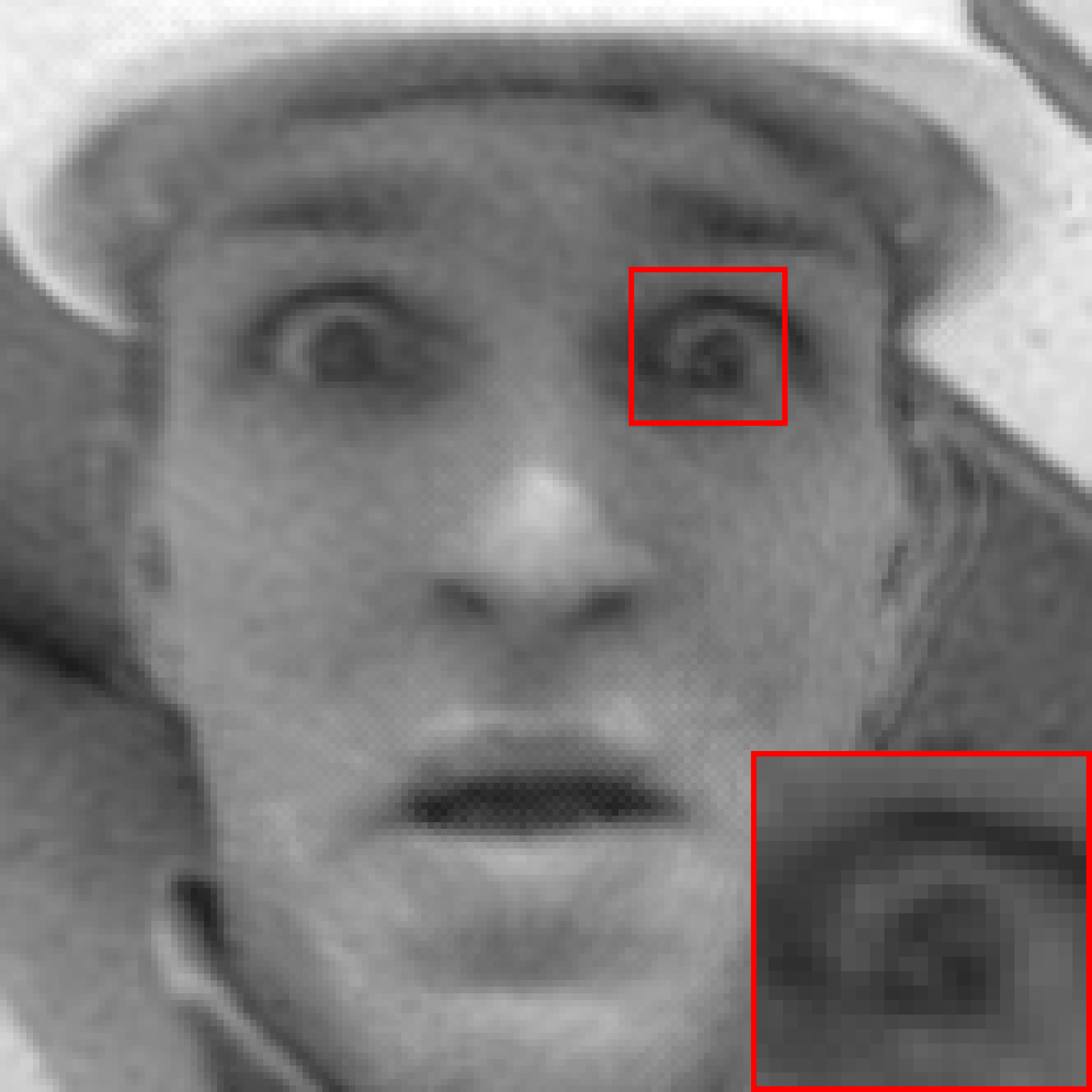}}
  \centerline{(f)}\smallskip
\end{minipage} \\
\vspace{-0.7cm}
\caption{Sample of the 93th reconstructed frame of the \textit{Foreman} sequence (with large innovation's level). (a) Original image. (b) Bicubic interpolation (MSE=17.43dB, SSIM=0.886). (c) LMS (MSE=18.91dB, SSIM=0.810). (d) R-LMS (MSE=17.71dB, SSIM=0.856). (e) TSR-LMS (MSE=17.00dB, SSIM=0.886). (f) LTSR-LMS (MSE=17.21dB, SSIM=0.887).}
\label{fig:resRealMoreInn}
\end{figure}

\begin{figure*}[htb]
\begin{minipage}[b]{0.16\textwidth}
  \centering
  \centerline{\includegraphics[width=0.9\linewidth]{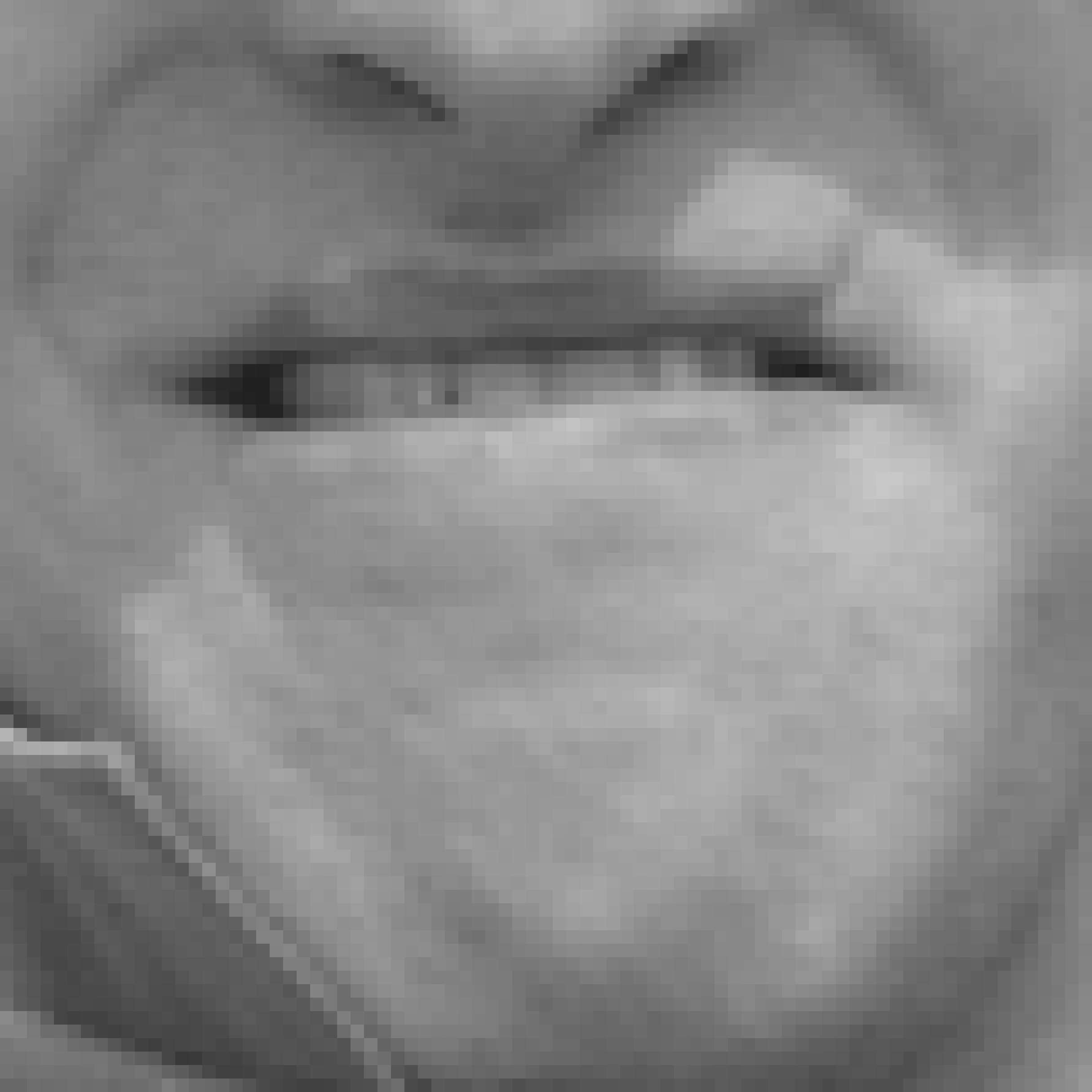}}
  \centerline{(a)}\smallskip
\end{minipage}
%\hfill
\begin{minipage}[b]{0.16\textwidth}
  \centering
  \centerline{\includegraphics[width=0.9\linewidth]{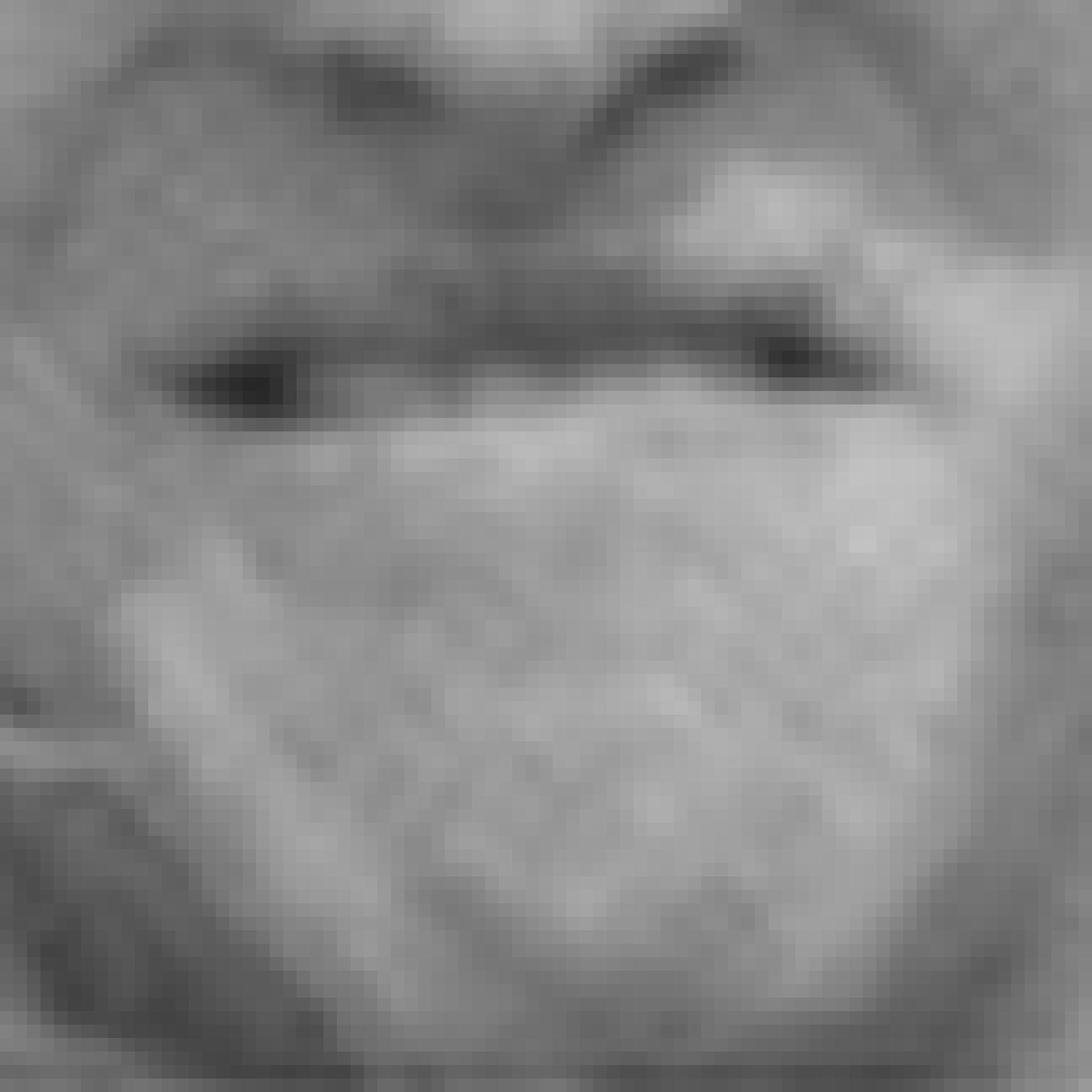}}
  \centerline{(b)}\smallskip
\end{minipage}
\begin{minipage}[b]{0.16\textwidth}
  \centering
  \centerline{\includegraphics[width=0.9\linewidth]{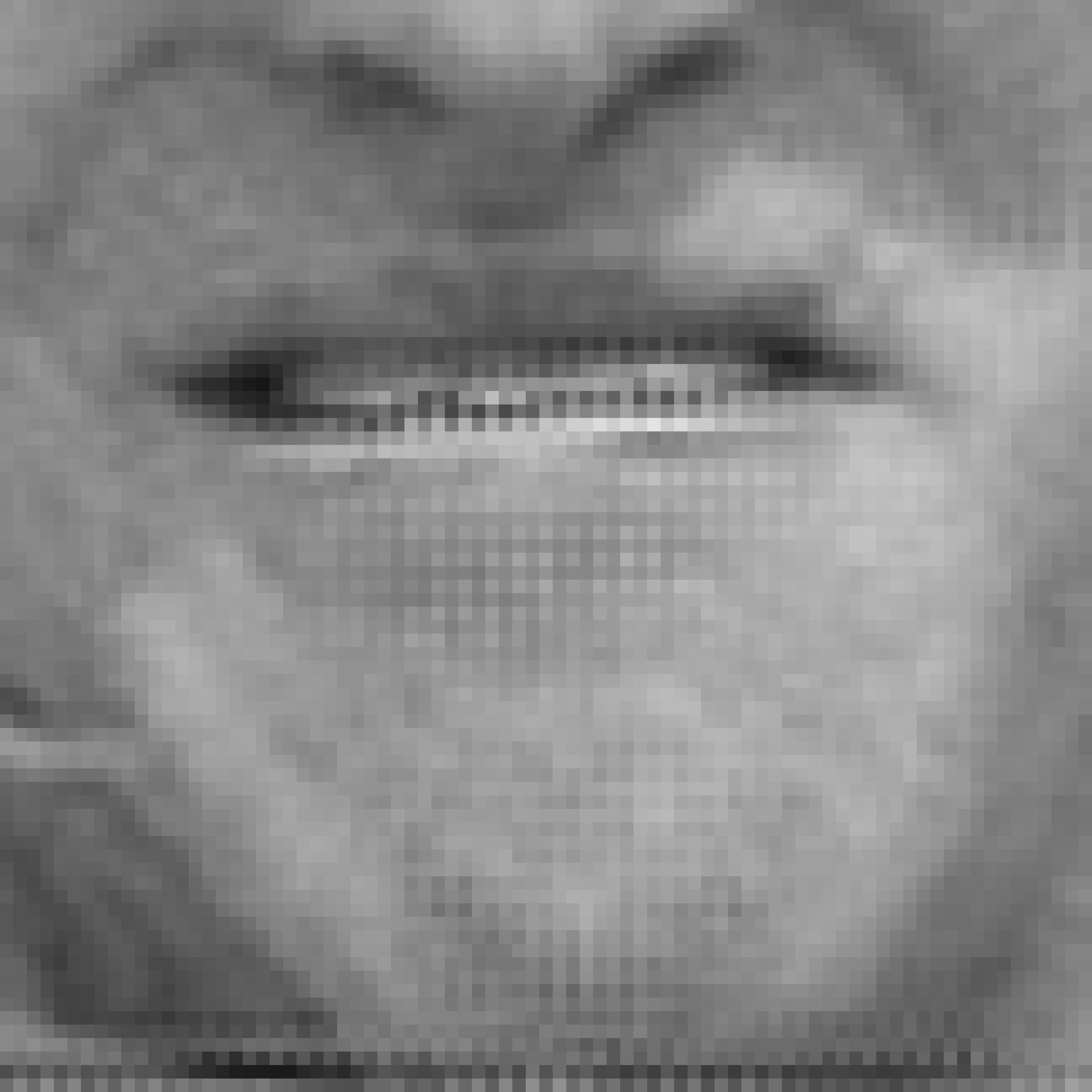}}
  \centerline{(c)}\smallskip
\end{minipage}
%\hfill
%
\begin{minipage}[b]{0.16\textwidth}
  \centering
  \centerline{\includegraphics[width=0.9\linewidth]{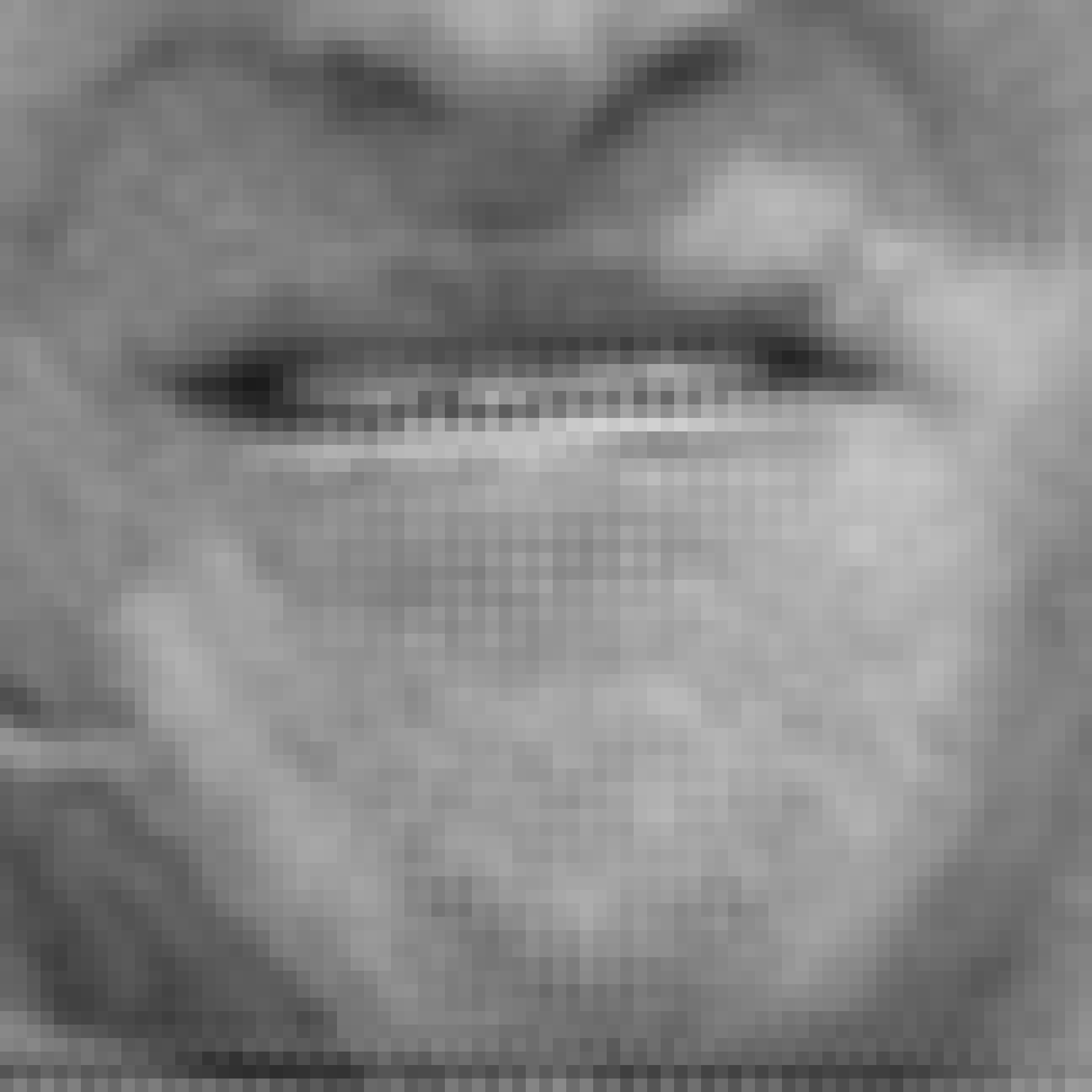}}
  \centerline{(d)}\smallskip
\end{minipage}
% \hfill
%
%
\begin{minipage}[b]{0.16\textwidth}
  \centering
  \centerline{\includegraphics[width=0.9\linewidth]{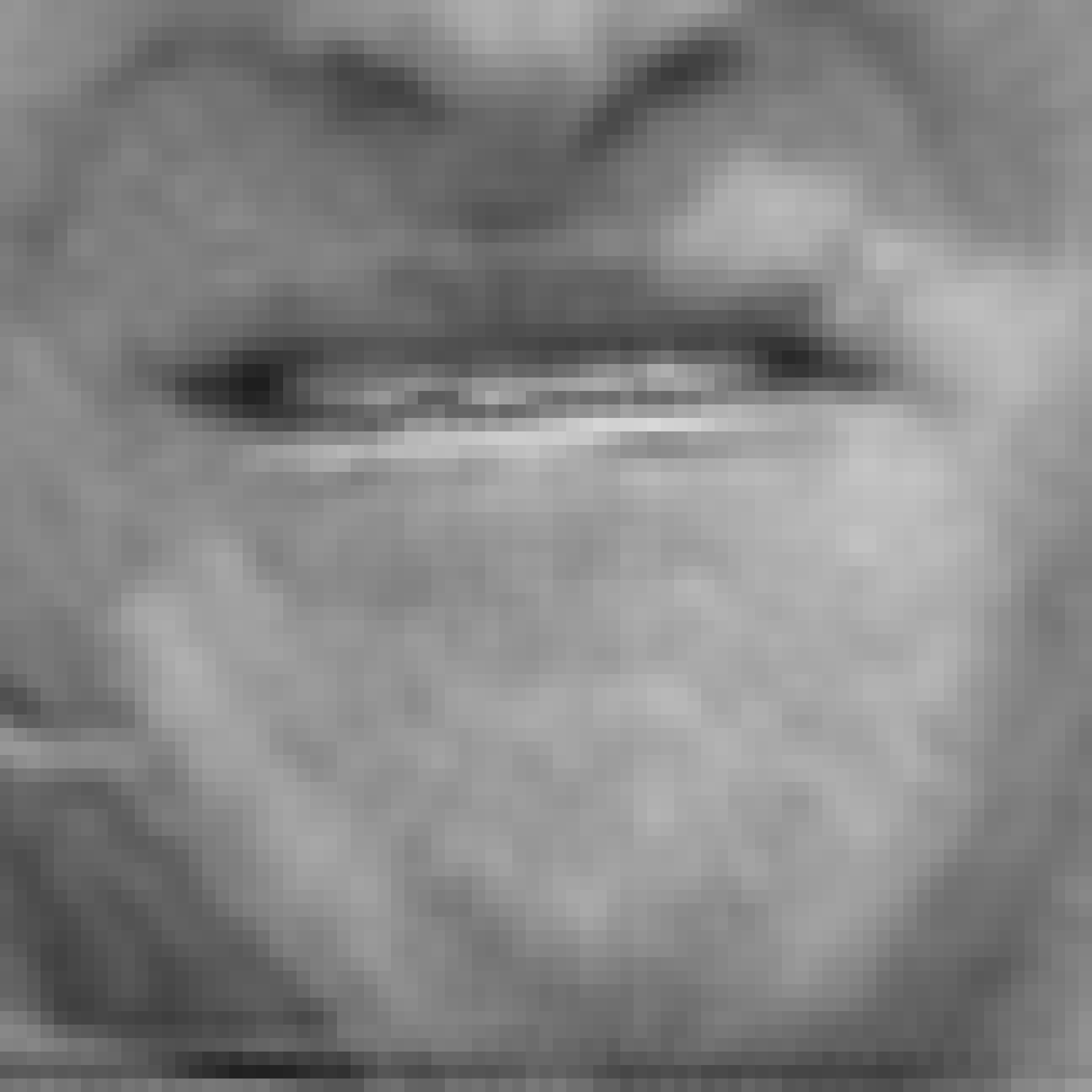}}
  \centerline{(e)}\smallskip
\end{minipage}
% \hfill
\begin{minipage}[b]{0.16\textwidth}
  \centering
  \centerline{\includegraphics[width=0.9\linewidth]{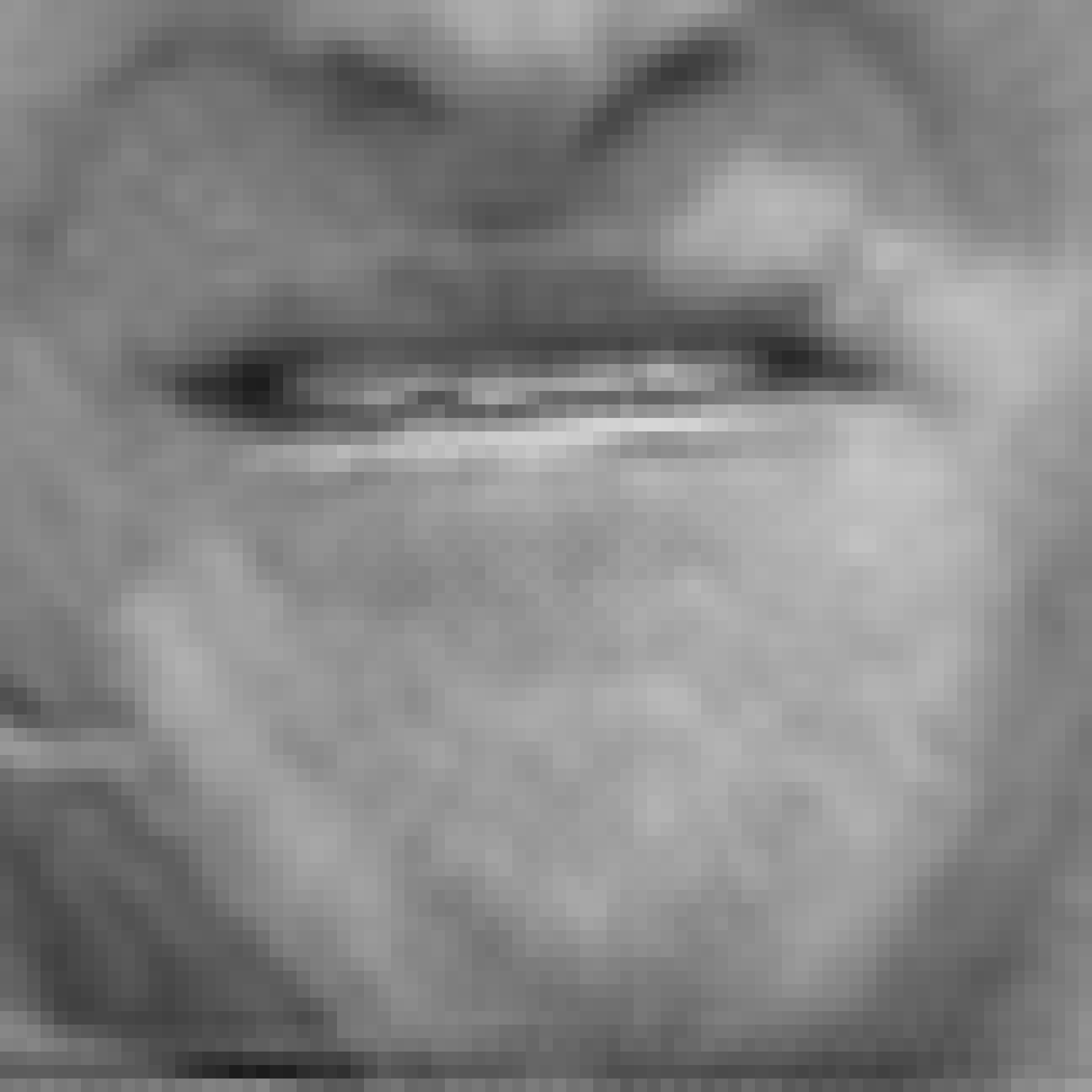}}
  \centerline{(f)}\smallskip
\end{minipage}
\vspace{-0.3cm}
\caption{Sample of the 33th reconstructed frame of the \textit{Foreman} sequence (with small innovation level). (a) Original image. (b) Bicubic interpolation (MSE=18.00dB, SSIM=0.868). (c) LMS (MSE=17.71dB, SSIM=0.849). (d) R-LMS (MSE=17.43dB, SSIM=0.868). (e) TSR-LMS (MSE=17.55dB, SSIM=0.879). (f) LTSR-LMS (MSE=17.56dB, SSIM=0.884).}
\label{fig:resRealLessInn}
\end{figure*}

Figure~\ref{fig:MSEreal} shows the MSE evolutions, which indicate a better performance for the TSR-LMS and LTSR-LMS methods, both of which performed similarly. The improvement offered by the proposed methods can be most clearly observed when a high degree of innovations is present in the scene. It can be noted that the reconstruction error exhibits a more regular behavior across the entire sequence, without being considerably influenced by the outliers, which cause significant spikes in the MSE of the LMS and R-LMS algorithms. To illustrate this scenario, Figure~\ref{fig:resRealMoreInn} shows the 93rd super-resolved frame of the \textit{Foreman} sequence. The advantage of the proposed methods becomes apparent through a more clear reconstruction result, as opposed to a vast amount of artifacts found in the images reconstructed by the LMS and R-LMS algorithms, mainly in the regions where innovations are present. The MSE performances of the four methods are less discernible at the time intervals where the amount of innovations is less significant. Nevertheless, the difference in the perceptual quality of the reconstructed images is still noticeable. For instance, Figure~\ref{fig:resRealLessInn} shows the reconstruction results of part of the 33rd frame of the \textit{Foreman} sequence. Although for this frame the MSE differences between the results of the four algorithms are small, the images super-resolved by the proposed methods still offer a better perceptual quality with reduced artifacts where small localized motion occur (e.g. close to the man's mouth).

% \begin{table} [htb]
% \small
% \caption{\cred{Average processing time per frame for the videos in Example~4.}} % CNN = 81,7146 per frame for large imgs
% \vspace{-0.3cm}
% \begin{center}
% \renewcommand{\arraystretch}{1.2}
% \begin{tabular}{c||ccccccc}
% \hline
% & {Time (in seconds)}  \\
% \hline
% Bicubic~\cite{blu2001moms} & 1.56 \\
% LMS & 0.46 \\
% %
% R-LMS & 0.53 \\
% %
% TSR-LMS & 0.60 \\
% %
% LTSR-LMS & 0.62 \\
% %
% Bayesian~\cite{liu2014bayesianVideoSRR} & 49.68  \\
% %
% CNN~\cite{tao2017detailRevealingSRRneuralNets} & 81.72  \\
% \hline
% \end{tabular}
% \end{center}
% \label{tab:algs_processing_time}
% \end{table}

\begin{table} [htb]
\small
\caption{Average processing time per frame for the videos in Example~4.}
\vspace{-0.2cm}
\centering
% \begin{center}
\renewcommand{\arraystretch}{1.1}
\setlength\tabcolsep{3.5pt}
\resizebox{\linewidth}{!}{%
\begin{tabular}{ccccccc}
\hline
% & Time (in seconds)  \\
Bicubic & LMS & R-LMS & TSR-LMS & LTSR-LMS & Bayesian~\cite{liu2014bayesianVideoSRR} & CNN~\cite{tao2017detailRevealingSRRneuralNets} \\
\hline
1.56\,s & 0.46\,s & 0.53\,s & 0.60\,s & 0.62\,s & 49.68\,s & 81.72\,s  \\
\hline
\end{tabular}}
% \end{center}
\label{tab:algs_processing_time}
\end{table}

\begin{table} [htb]
\small
\caption{Average PSNR (in decibels) for the videos in Example~4.}
\vspace{-0.3cm}
\centering
% \begin{center}
\renewcommand{\arraystretch}{1.1}
\setlength\tabcolsep{3.5pt}
\resizebox{\linewidth}{!}{%
\begin{tabular}{l|cccccc|c}
\hline
% & Time (in seconds)  \\
 & Bear & Bus & Elephant & Horse & Paragliding & Sheep & Mean \\ %& Train \\
\hline
Bicubic  & 28.43 & 32.66 & 30.56 & 29.91 & 35.58 & 25.02 & 30.36 \\ %& 31.15 \\
LMS      & 32.78 & 31.06 & 32.51 & 28.73 & 33.24 & 30.46 & 31.46 \\ %& 33.15 \\
R-LMS    & 33.38 & 33.09 & 33.57 & 30.89 & 35.10 & 30.37 & 32.73 \\ %& 34.58 \\
TSR-LMS  & 34.00 & 34.46 & 34.16 & 32.61 & 36.48 & 30.99 & 33.78 \\ %& 35.56 \\
LTSR-LMS & 33.94 & 34.89 & 34.25 & 33.26 & 36.59 & 30.75 & 33.95 \\ %& 35.63 \\
Bayesian%~\cite{liu2014bayesianVideoSRR} 
         & 30.52 & 33.42 & 32.08 & 31.72 & 34.87 & 29.60 & 32.04 \\ %& 33.08 \\
CNN%~\cite{tao2017detailRevealingSRRneuralNets} 
         & 32.21 & 33.89 & 32.96 & 32.75 & 34.54 & 29.66 & 32.67 \\ %& 33.63 \\
\hline
\end{tabular}}
% \end{center}
\label{tab:algs_PSNR_ex4}
\end{table}

% \begin{table} [htb]
% \small
% \caption{\cred{Average MSE (in decibels) for the videos in Example~4.}}
% \vspace{-0.65cm}
% \begin{center}
% \renewcommand{\arraystretch}{1.1}
% \resizebox{\linewidth}{!}{%
% \begin{tabular}{l|ccccccc}
% \hline
% % & Time (in seconds)  \\
%  & Bear & Bus & Elephant & Horse & Paragliding & Sheep \\ %& Train \\
% \hline
% Bicubic  & 19.70 & 15.47 & 17.57 & 18.23 & 12.55 & 23.11 \\ %& 16.98 \\
% LMS      & 15.35 & 17.07 & 15.62 & 19.40 & 14.90 & 17.67 \\ %& 14.98 \\
% R-LMS    & 14.75 & 15.04 & 14.56 & 17.24 & 13.03 & 17.76 \\ %& 13.55 \\
% TSR-LMS  & 14.13 & 13.68 & 13.97 & 15.52 & 11.66 & 17.14 \\ %& 12.57 \\
% LTSR-LMS & 14.19 & 13.24 & 13.88 & 14.87 & 11.54 & 17.39 \\ %& 12.50 \\
% Bayesian~\cite{liu2014bayesianVideoSRR} 
%          & 17.62 & 14.71 & 16.06 & 16.41 & 13.26 & 18.53 \\ %& 15.06 \\
% CNN~\cite{tao2017detailRevealingSRRneuralNets} 
%          & 15.92 & 14.25 & 15.17 & 15.39 & 13.59 & 18.47 \\ %& 14.50 \\
% \hline
% \end{tabular}}
% \end{center}
% \label{tab:algs_MSE_ex4}
% \end{table}

% =======================================================================
\subsection{Example 4}
\label{sec:examples_real_ii}

% \cred{This example extends Example~3 in order to compare the proposed solutions when super-resolving more recent and challenging video sequences.}

% In this example, we compare the proposed method with the state of the art, .

This example illustrates the performance of the algorithms using recent and challenging video sequences taken from the DAVIS dataset~\cite{pont2017davisDatasetSegmentation}.
We extracted six reference HR sequences from videos with resolution of $1080\times1920$ pixels, and generated the degraded LR sequences from them as in Example~3.

To compare performances with recent state-of-the-art algorithms, we super-resolved these sequences using interpolation, LMS, R-LMS and two more recent video SRR algorithms, namely, the adaptive Bayesian method from~\cite{liu2014bayesianVideoSRR} and the Convolutional Neural Network (CNN) from~\cite{tao2017detailRevealingSRRneuralNets}. For the method of~\cite{liu2014bayesianVideoSRR} we fixed the solution to the blur matrix estimation of $\mH$ at the optimal value and used the same registration algorithm of~\cite{Sun10}, which was also used for the other methods. The CNN has an embedded registration algorithm which could not be modified. The parameters used for LMS, R-LMS, TSR-LMS and LTSR-LMS methods were the same as in Table~\ref{tab:parametersOutlier} (which were determined based on the simulations with synthetic sequences in Example~2).
The CNN~\cite{tao2017detailRevealingSRRneuralNets} was implemented in Python using TensorFlow. The other methods were implemented in Matlab$^\copyright$. Codes for methods from~\cite{liu2014bayesianVideoSRR} and~\cite{tao2017detailRevealingSRRneuralNets} were provided by the respective authors. All simulations were executed on a desktop computer with an Intel Core I7 processor with 4.2Ghz and 16Gb of RAM.

\begin{figure}[htb]
\begin{minipage}[b]{.49\linewidth}
  \centering
  \centerline{\includegraphics[width=0.95\linewidth]{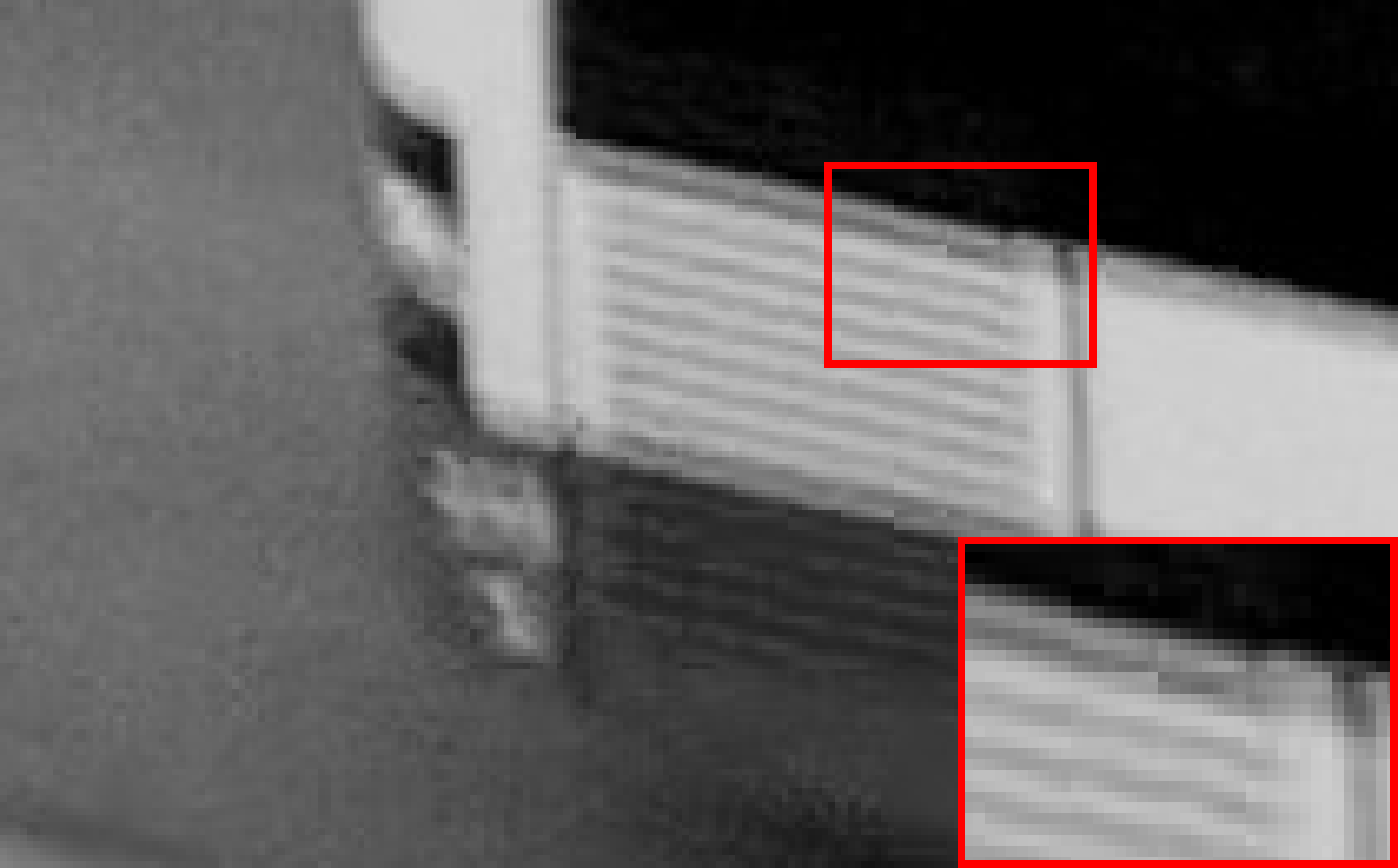}}
  \centerline{(a)}\smallskip
\end{minipage}
%\hfill
\begin{minipage}[b]{0.49\linewidth}
  \centering
  \centerline{\includegraphics[width=0.95\linewidth]{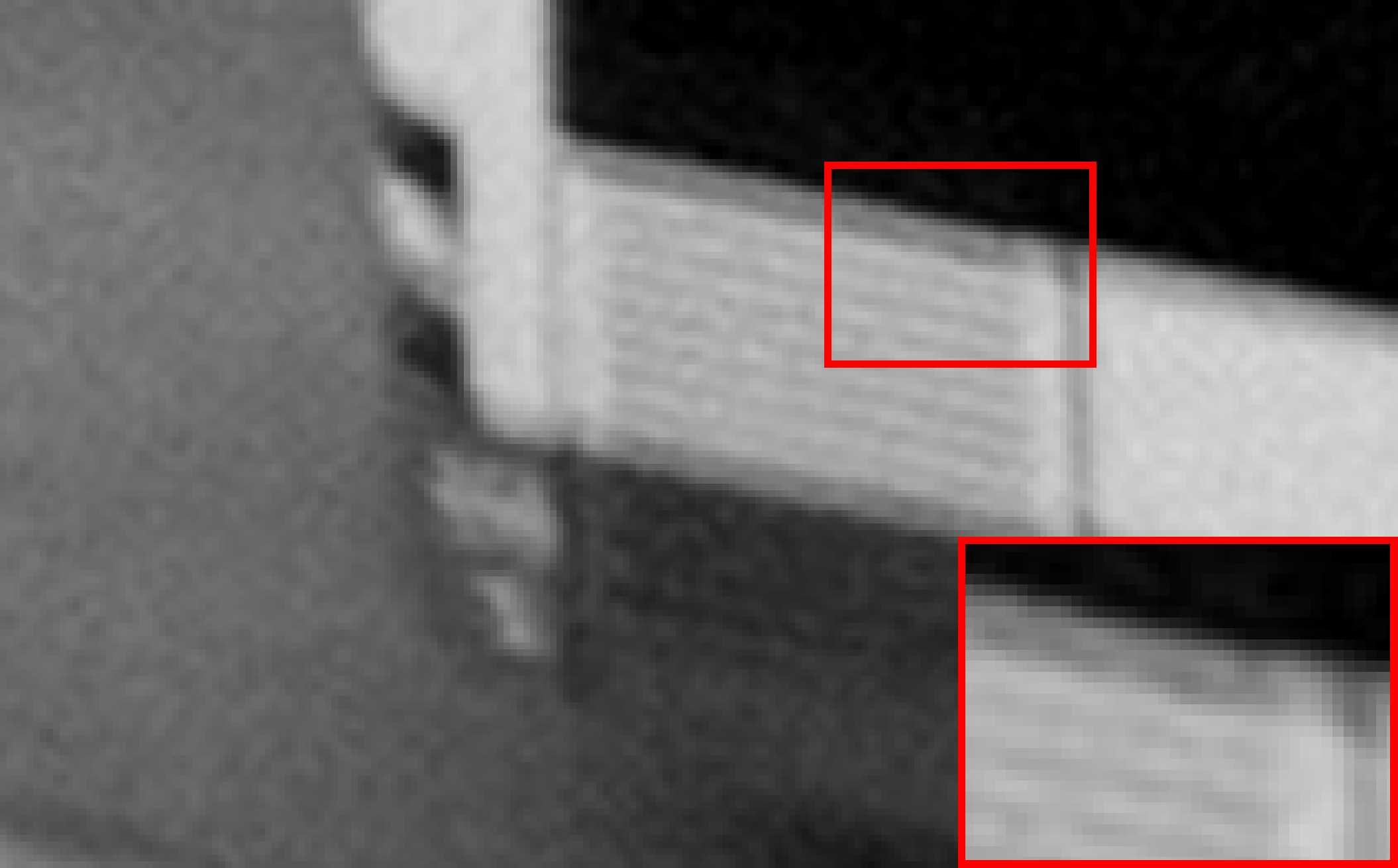}}
  \centerline{(b)}\smallskip
\end{minipage}
\begin{minipage}[b]{.49\linewidth}
  \centering
  \centerline{\includegraphics[width=0.95\linewidth]{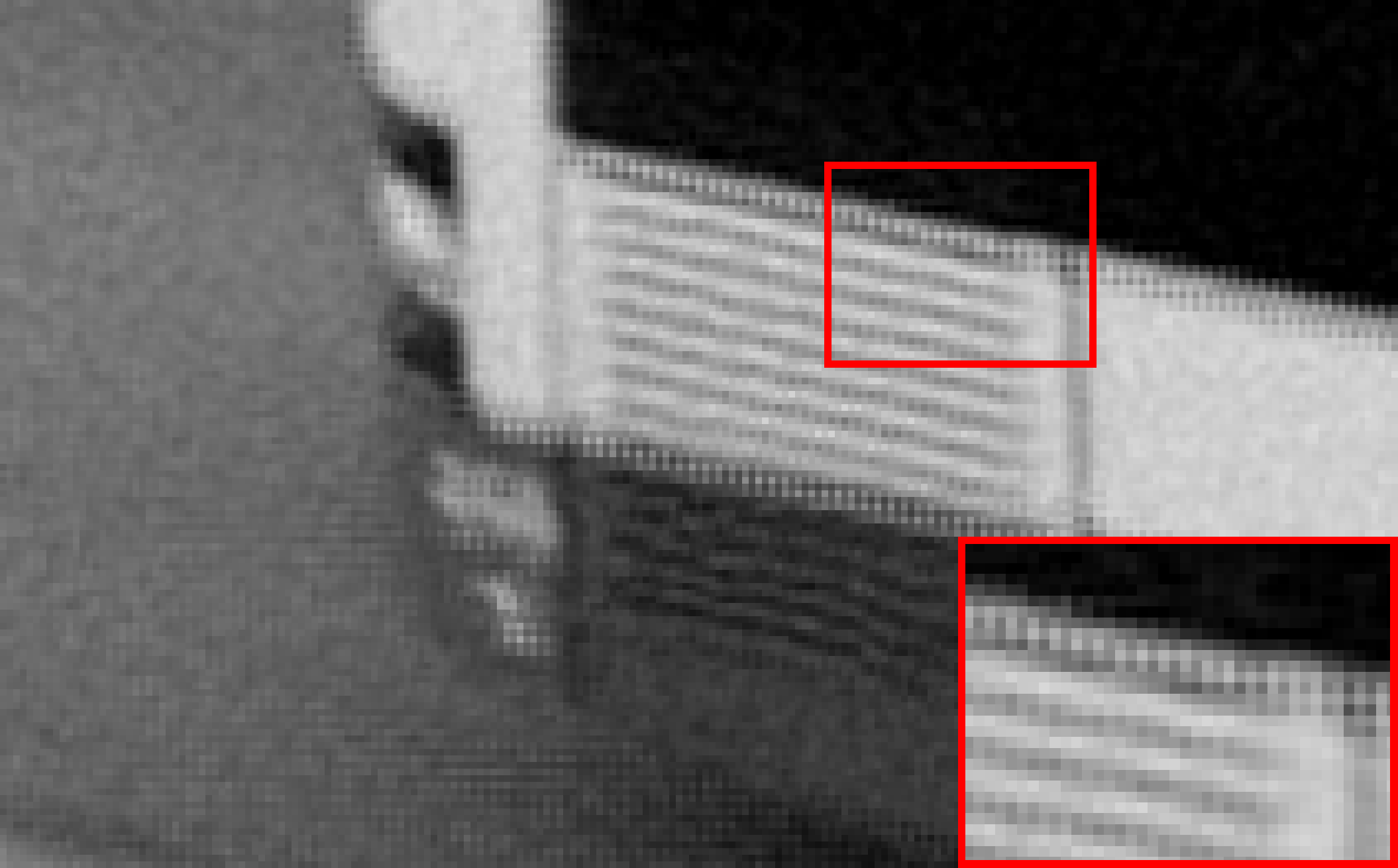}}
  \centerline{(c)}\smallskip
\end{minipage}
%\hfill
\begin{minipage}[b]{0.49\linewidth}
  \centering
  \centerline{\includegraphics[width=0.95\linewidth]{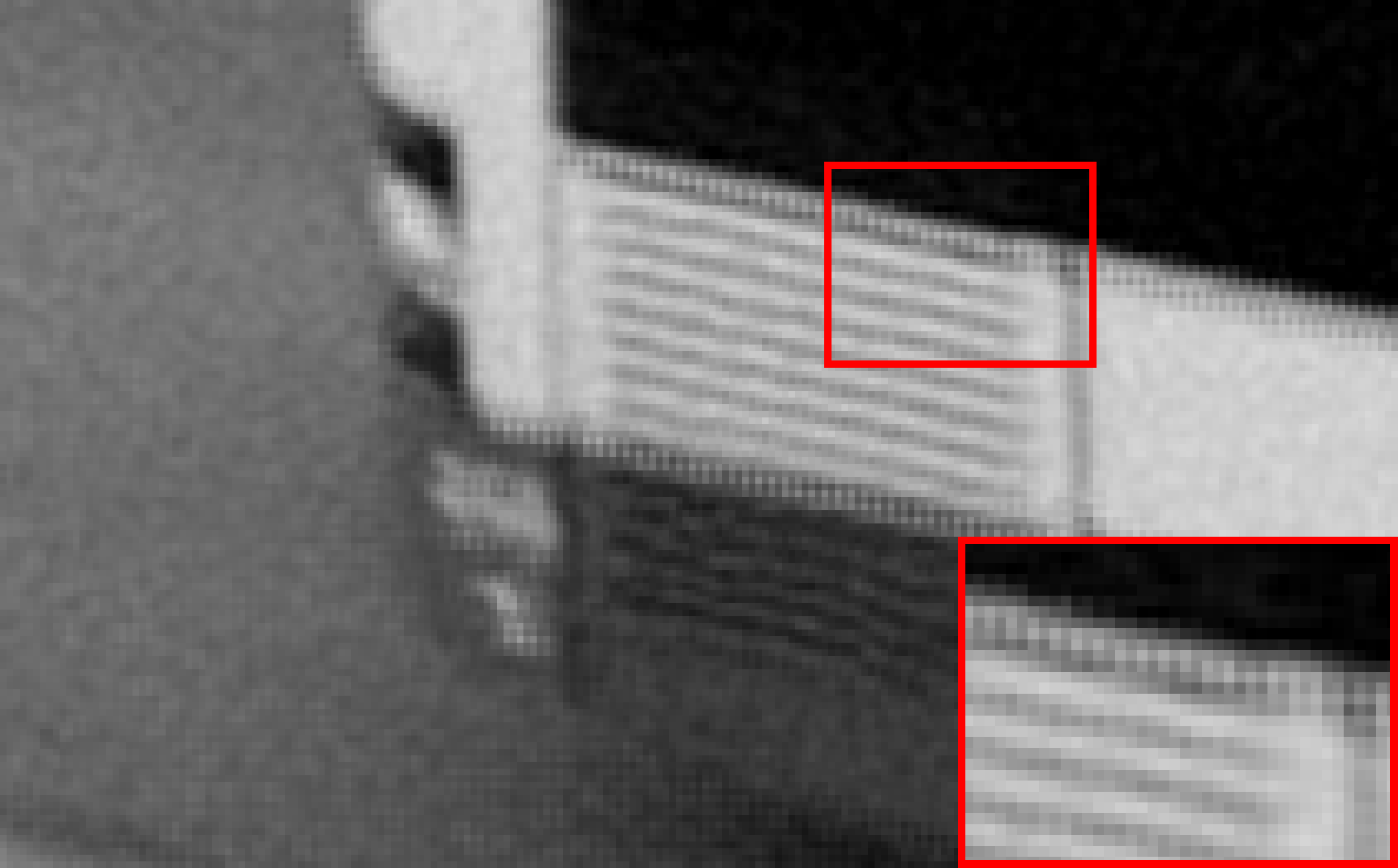}}
  \centerline{(d)}\smallskip
\end{minipage}
\begin{minipage}[b]{.49\linewidth}
  \centering
  \centerline{\includegraphics[width=0.95\linewidth]{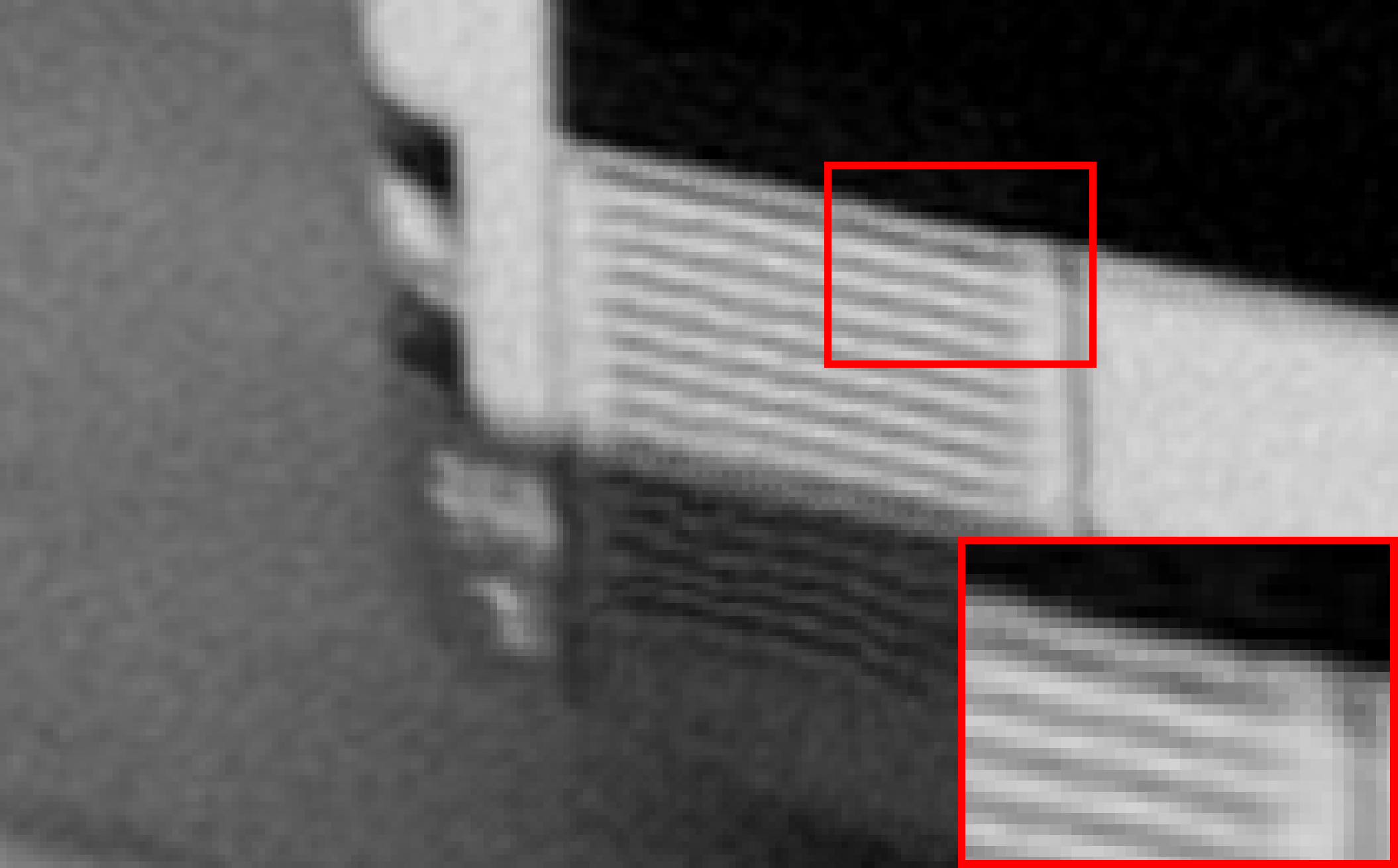}}
  \centerline{(e)}\smallskip
\end{minipage}
\hfill
\begin{minipage}[b]{0.49\linewidth}
  \centering
  \centerline{\includegraphics[width=0.95\linewidth]{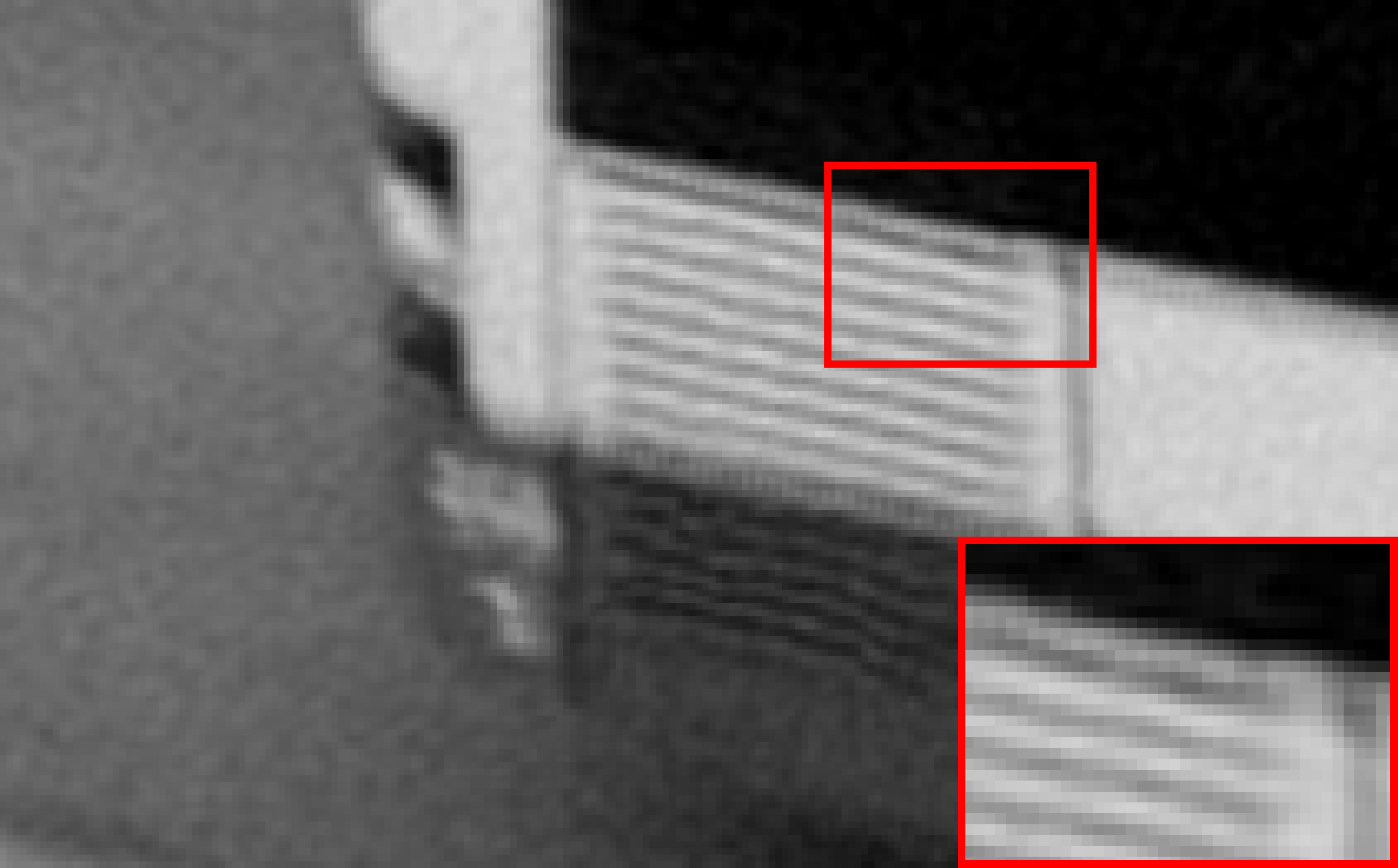}}
  \centerline{(f)}\smallskip
\end{minipage} 
\begin{minipage}[b]{.49\linewidth}
  \centering
  \centerline{\includegraphics[width=0.95\linewidth]{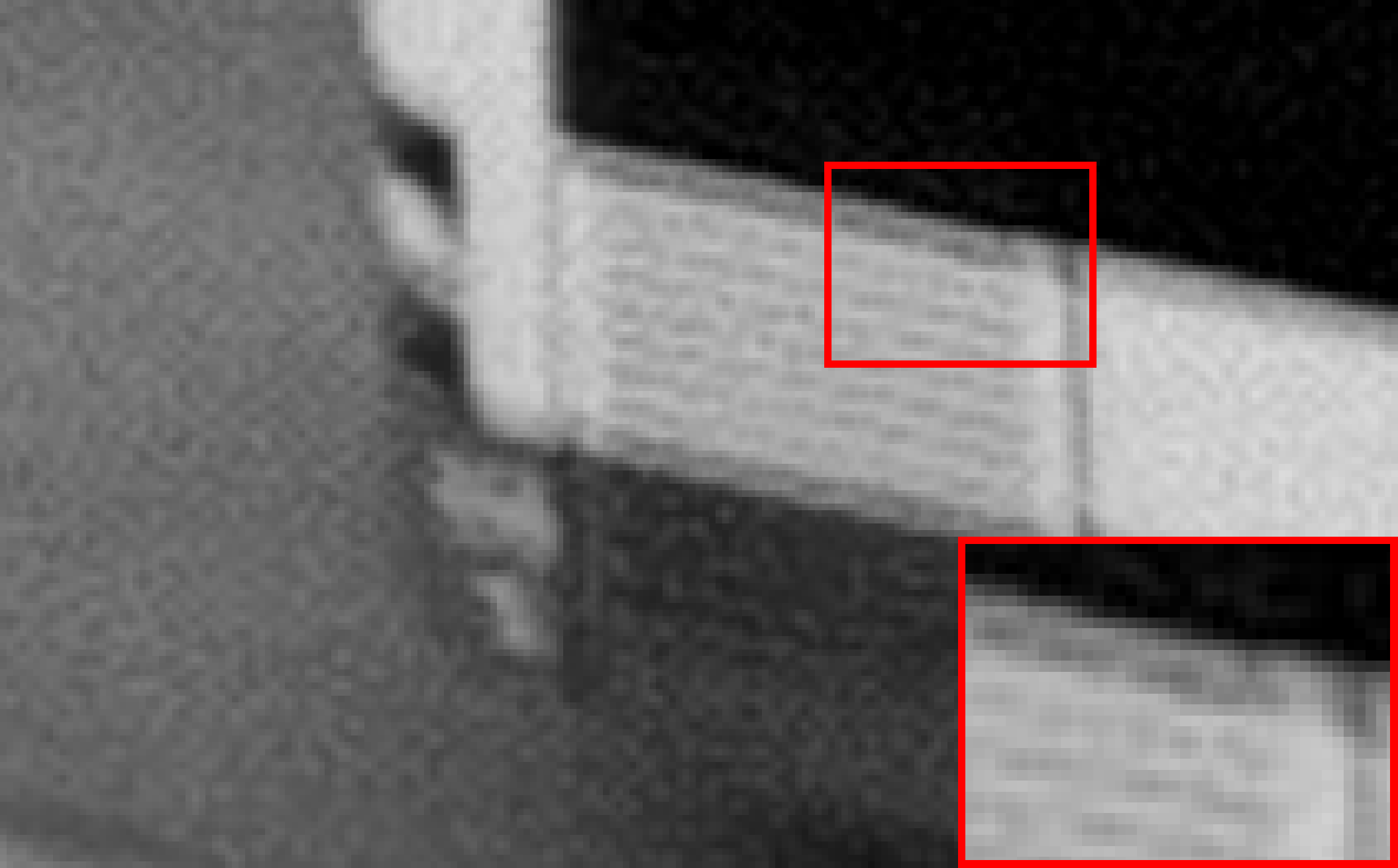}}
  \centerline{(g)}\smallskip
\end{minipage}
\hfill
\begin{minipage}[b]{0.49\linewidth}
  \centering
  \centerline{\includegraphics[width=0.95\linewidth]{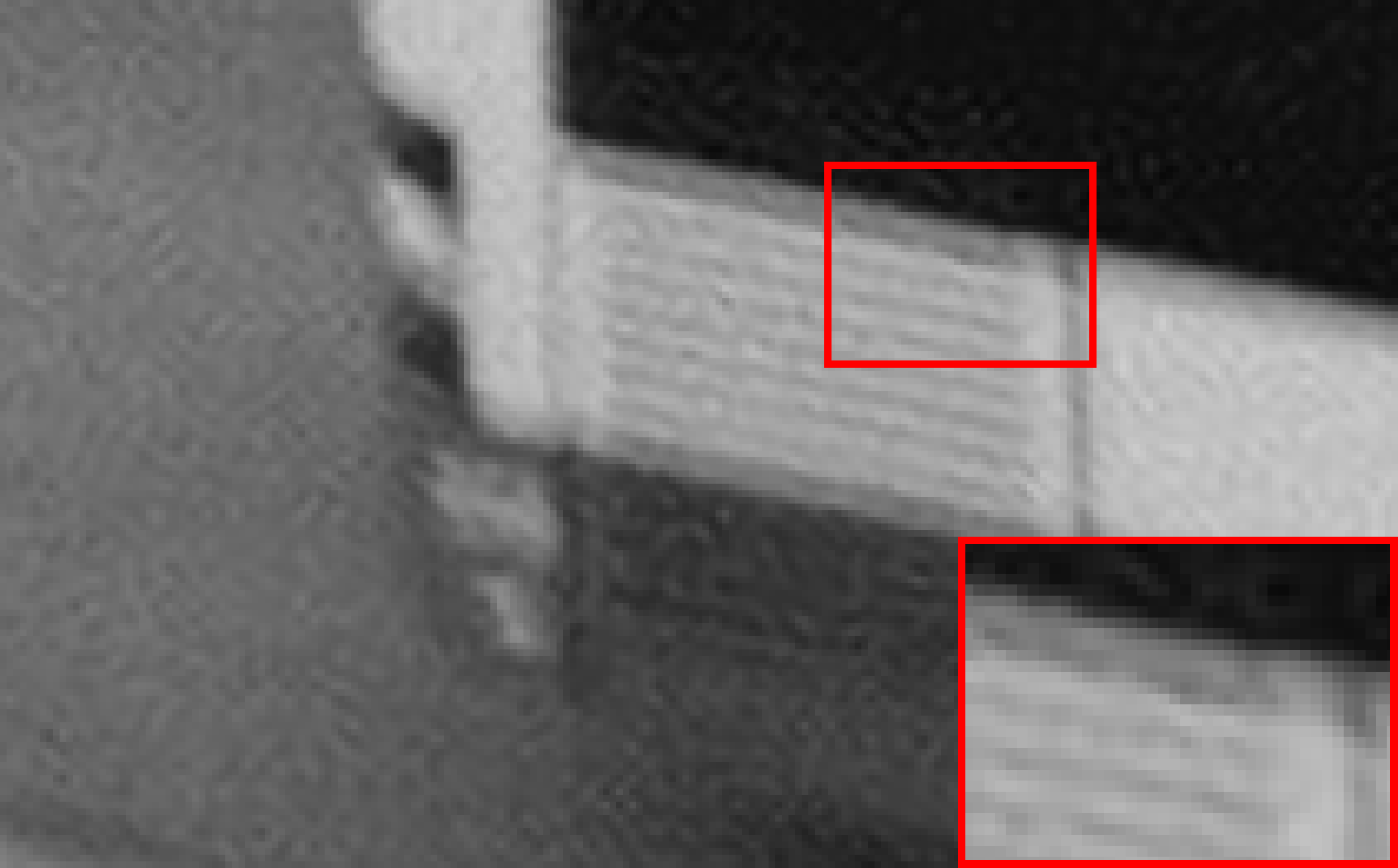}}
  \centerline{(h)}\smallskip
\end{minipage} \\
\vspace{-0.7cm}
\caption{Sample of the 49th frame of the \textit{bus} sequence. (a) Original image. (b) Bicubic interpolation. (c) LMS. (d) R-LMS. (e) TSR-LMS. (f) LTSR-LMS. (g) Bayesian method~\cite{liu2014bayesianVideoSRR}. (h) CNN~\cite{tao2017detailRevealingSRRneuralNets}.}
\label{fig:resNewvids1}
\end{figure}

\begin{figure}[htb]
\begin{minipage}[b]{.49\linewidth}
  \centering
  \centerline{\includegraphics[width=0.95\linewidth]{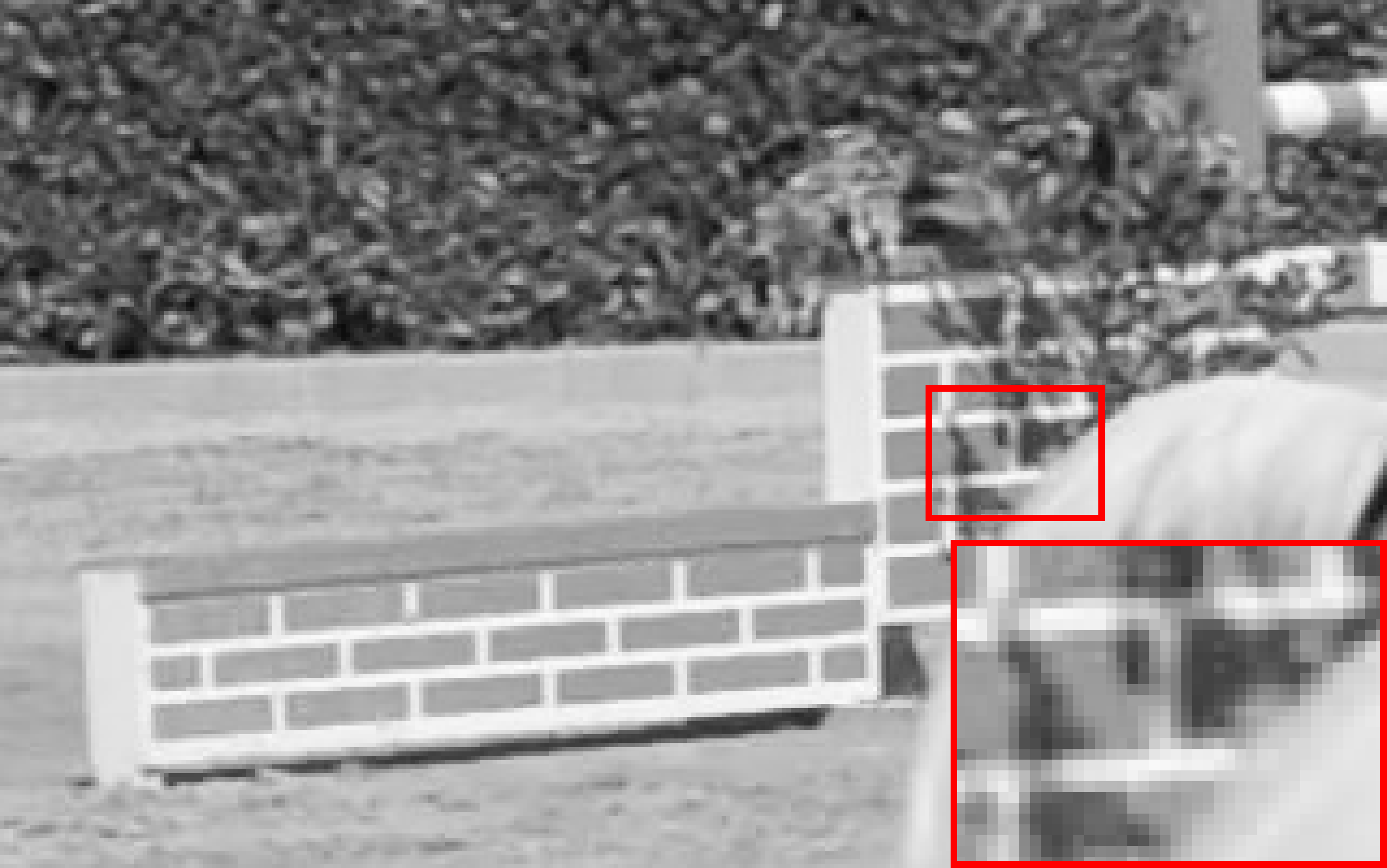}}
  \centerline{(a)}\smallskip
\end{minipage}
%\hfill
\begin{minipage}[b]{0.49\linewidth}
  \centering
  \centerline{\includegraphics[width=0.95\linewidth]{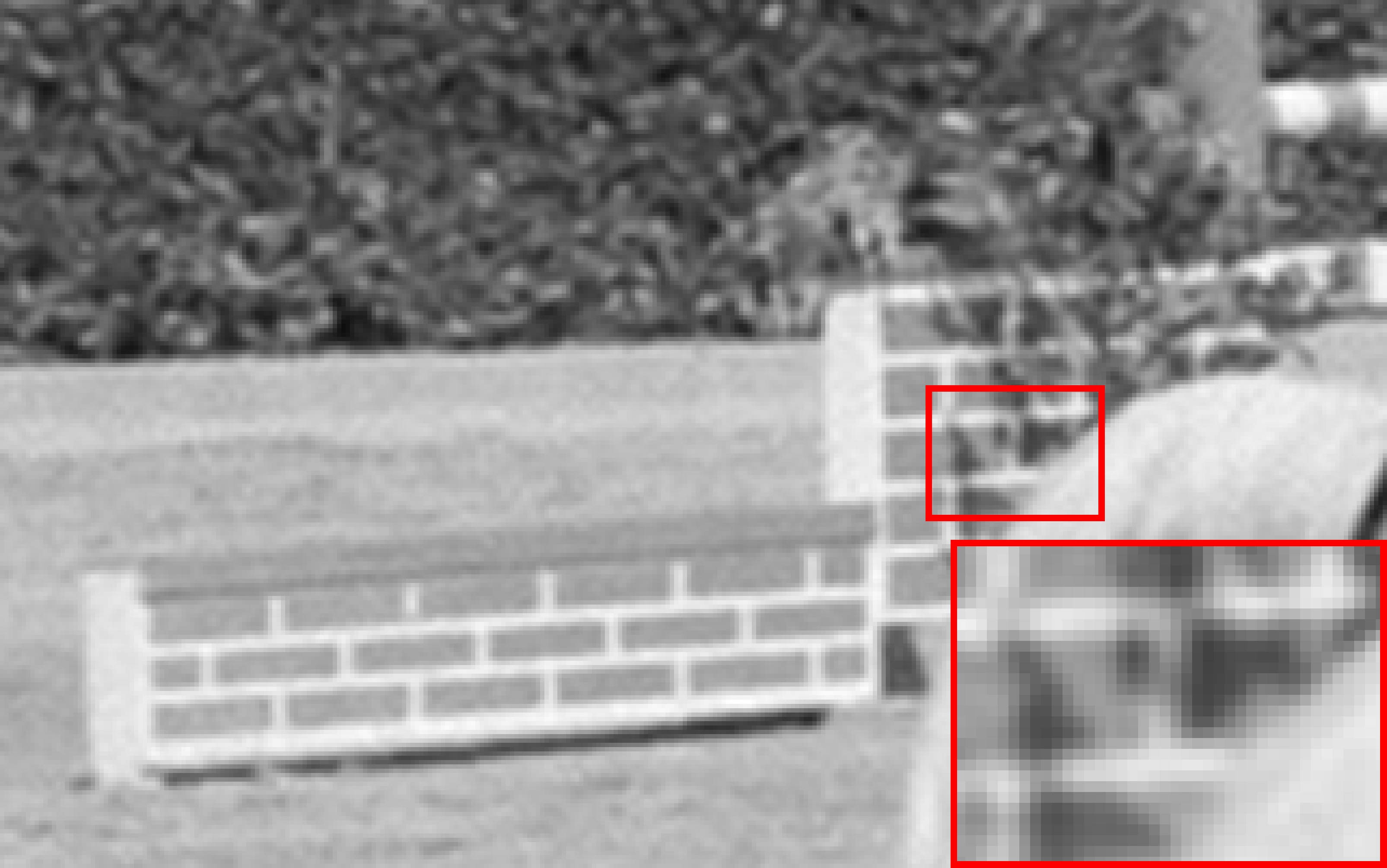}}
  \centerline{(b)}\smallskip
\end{minipage}
\begin{minipage}[b]{.49\linewidth}
  \centering
  \centerline{\includegraphics[width=0.95\linewidth]{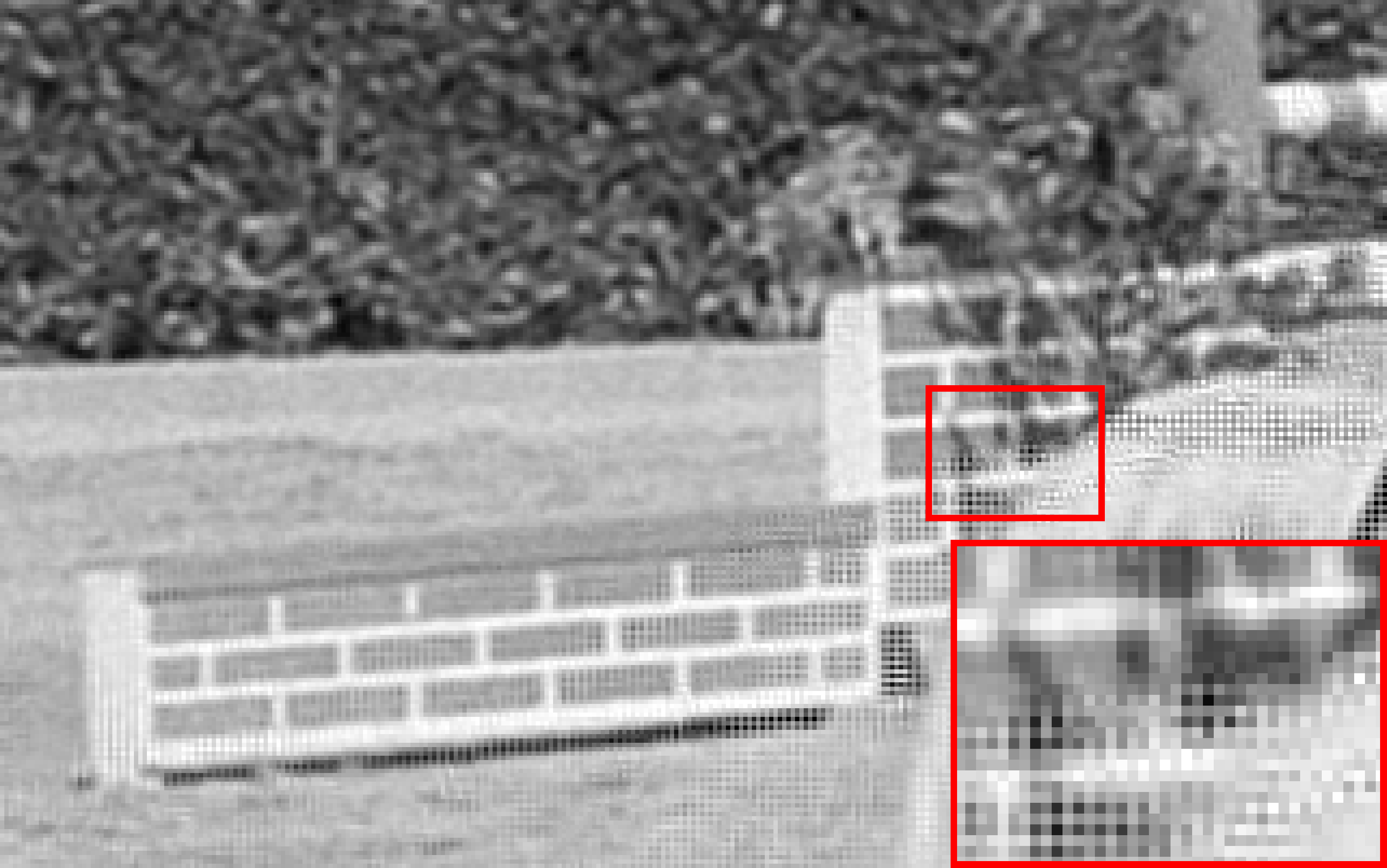}}
  \centerline{(c)}\smallskip
\end{minipage}
%\hfill
\begin{minipage}[b]{0.49\linewidth}
  \centering
  \centerline{\includegraphics[width=0.95\linewidth]{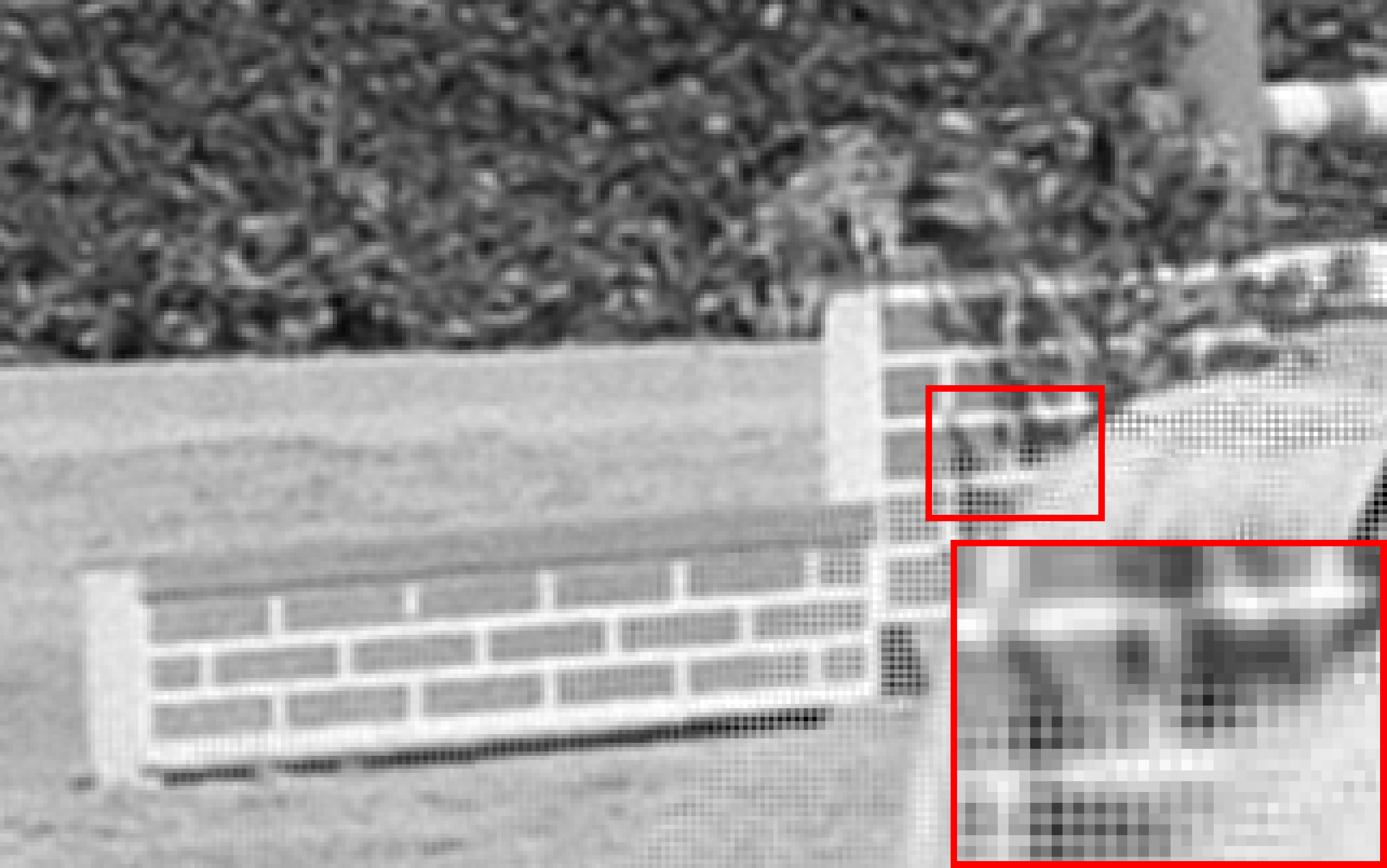}}
  \centerline{(d)}\smallskip
\end{minipage}
\begin{minipage}[b]{.49\linewidth}
  \centering
  \centerline{\includegraphics[width=0.95\linewidth]{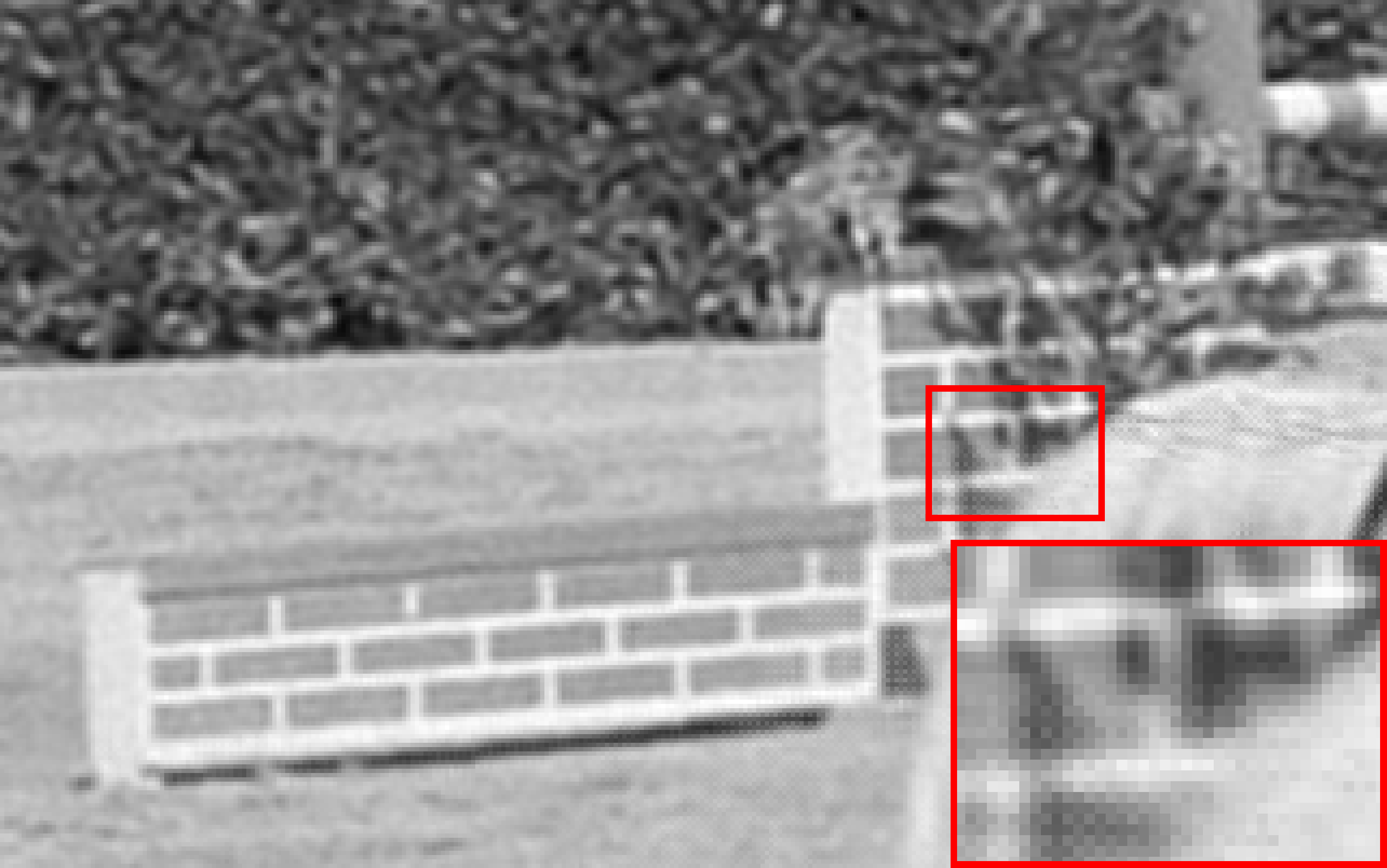}}
  \centerline{(e)}\smallskip
\end{minipage}
\hfill
\begin{minipage}[b]{0.49\linewidth}
  \centering
  \centerline{\includegraphics[width=0.95\linewidth]{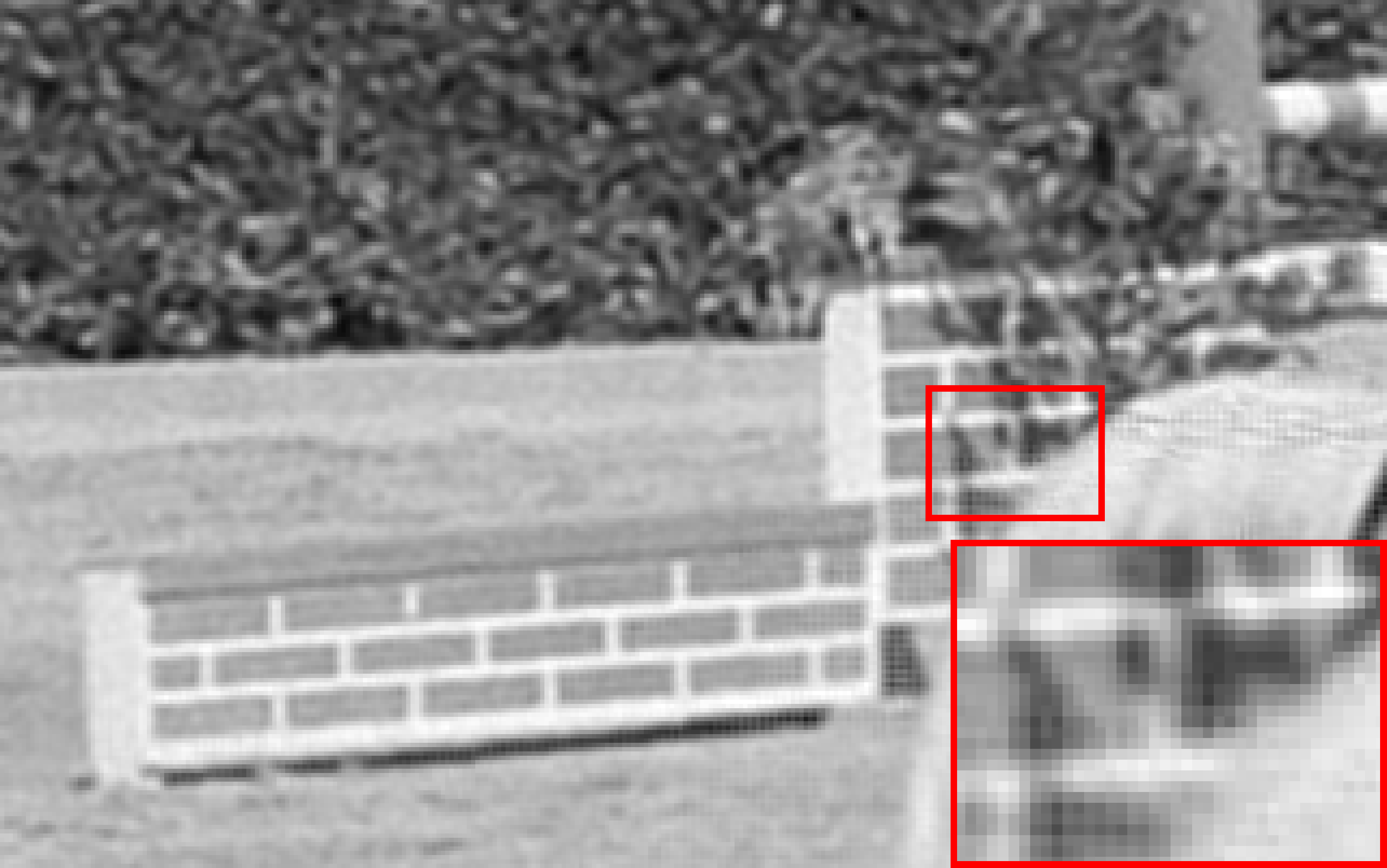}}
  \centerline{(f)}\smallskip
\end{minipage} 
\begin{minipage}[b]{.49\linewidth}
  \centering
  \centerline{\includegraphics[width=0.95\linewidth]{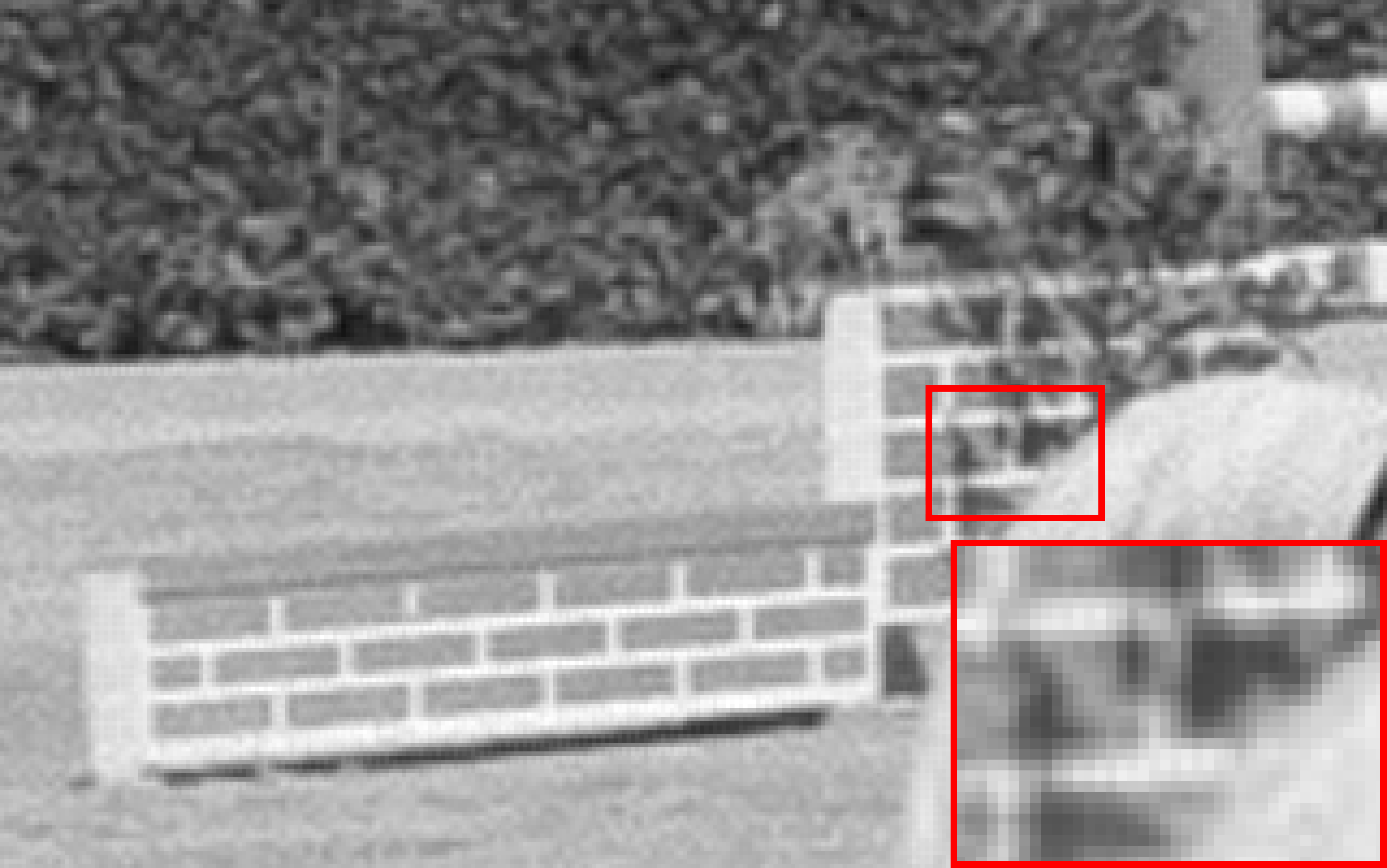}}
  \centerline{(g)}\smallskip
\end{minipage}
\hfill
\begin{minipage}[b]{0.49\linewidth}
  \centering
  \centerline{\includegraphics[width=0.95\linewidth]{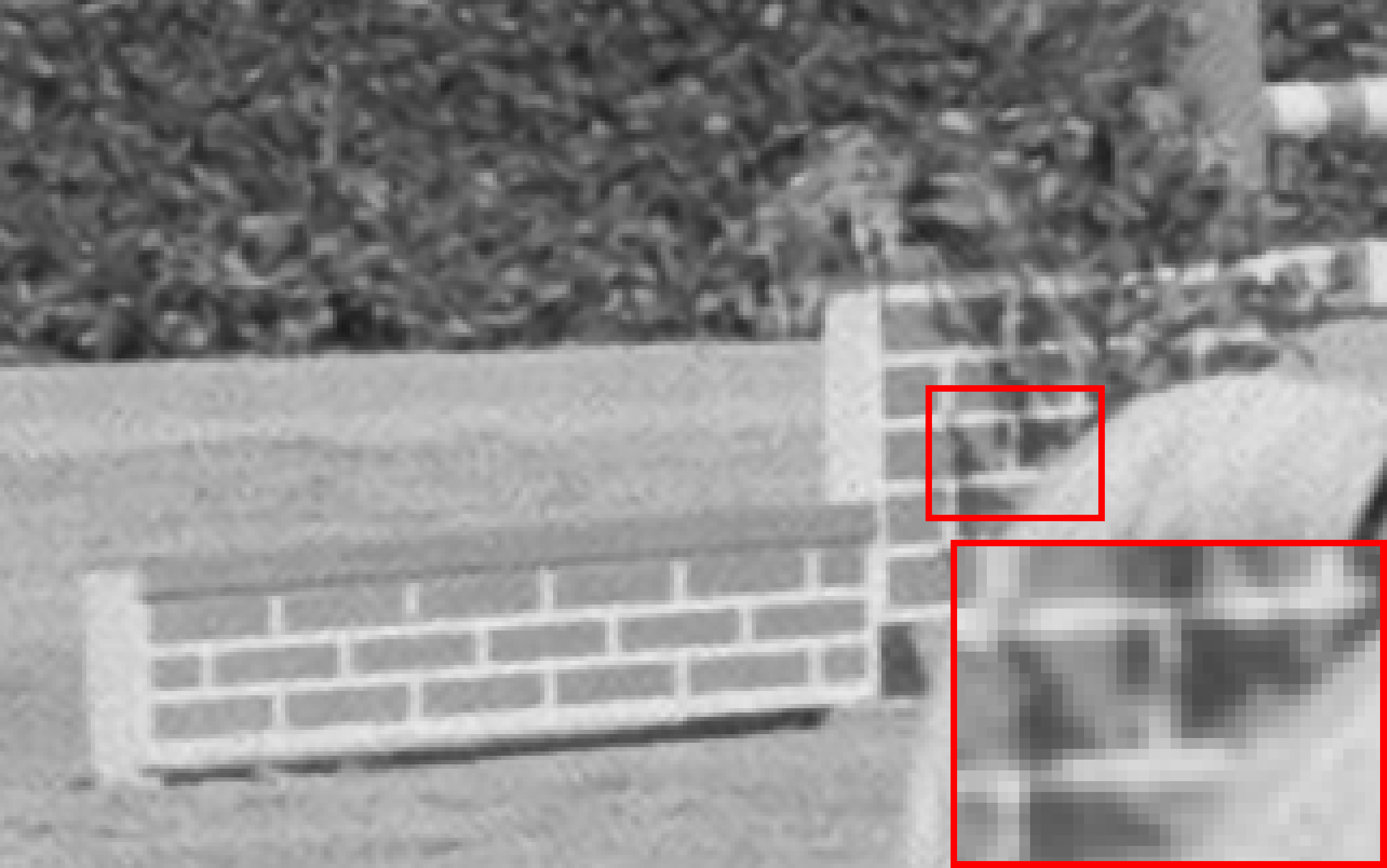}}
  \centerline{(h)}\smallskip
\end{minipage} \\
\vspace{-0.7cm}
\caption{Sample of the 50th frame of the \textit{horsejump-low} sequence. (a) Original image. (b) Bicubic interpolation. (c) LMS. (d) R-LMS. (e) TSR-LMS. (f) LTSR-LMS. (g) Bayesian method~\cite{liu2014bayesianVideoSRR}. (h) CNN~\cite{tao2017detailRevealingSRRneuralNets}.}
\label{fig:resNewvids3}
\end{figure}

We assess the performances of the different methods both quantitatively and visually. The peak signal to noise ratio (PSNR) for all algorithms and video sequences is presented in Table~\ref{tab:algs_PSNR_ex4}. It can be verified that the proposed methods usually led to an image quality that is at least comparable with (usually better than) that resulting from using the competing algorithms.
Excerpts from reconstruction results of two sequences\footnote{More extensive results are available in the supplementary document. Illustrative video excerpts are also included as supplementary material on~{https://www.dropbox.com/sh/hmyp5nks2sbj6to/AADnSqRE5YpTDJheLzDlTUI3a?dl=0}.}, shown in Figs.~\ref{fig:resNewvids1} and~\ref{fig:resNewvids3}, also support this conclusion.
%
% In several experiments we realized, the proposed methods usually lead to an image quality that is at least comparable with that resulting from using the competing algorithms.
Fig.~\ref{fig:resNewvids1} shows that the proposed methods yield a significantly better resolution than the algorithms in~\cite{liu2014bayesianVideoSRR} and~\cite{tao2017detailRevealingSRRneuralNets}.
%
% Fig.~\ref{fig:resNewvids2} also shows a better reconstruction result for the proposed methods, especially near the clock tower. %
% Fig.~\ref{fig:resNewvids3} is an example in which the competing methods performed better, specially the CNN~\cite{tao2017detailRevealingSRRneuralNets} as can be noticed in the brick wall. Still, the proposed methods provided good reconstruction quality and show less influence of noise.
Fig.~\ref{fig:resNewvids3} is an example in which the competing methods performed well, specially the CNN~\cite{tao2017detailRevealingSRRneuralNets} as can be noticed in the brick wall. Still, the proposed methods provided good reconstruction quality and show less influence of noise.

The methods of~\cite{liu2014bayesianVideoSRR} and~\cite{tao2017detailRevealingSRRneuralNets} also show robustness to innovation outliers, as can be noticed in the horse's tail in Fig.~\ref{fig:resNewvids3} where the reconstructed images do not exhibit artifacts like the LMS and R-LMS algorithms.
The proposed methods also show significantly less artifacts than the LMS and R-LMS algorithms, albeit not being as clear as the methods in~\cite{liu2014bayesianVideoSRR} and~\cite{tao2017detailRevealingSRRneuralNets}.

%\cred{
%Since techniques from different families are being compared and the Convolutional Neural Network does not directly minimizes a quantitative metric, the MSE and SSIM did not show a good agreement with the perceptual quality of the reconstruction (this occurs specially if the images are not very similar, due to small spatial displacements for example~\cite{kim2014problemMSEalignement}). Therefore we resort to visual comparisons in order to assess the quality of the reconstructions.
%}

Table~\ref{tab:algs_processing_time} shows the execution times\footnote{The execution time of the CNN~\cite{tao2017detailRevealingSRRneuralNets} includes image registration, since the SRR time cannot be measured separately in the provided implementation.} for Example~4. Algorithms~\cite{liu2014bayesianVideoSRR} and~\cite{tao2017detailRevealingSRRneuralNets} were approximately two orders of magnitude slower than the remaining methods (except for bicubic interpolation). These results and the comparable reconstruction quality clearly show that the proposed algorithms are competitive in term of quality and robustness at a much lower computational cost.

The implementations of the proposed algorithms were not optimized, and thus they could not be tested in real-time. However, real-time implementation is perfectly within reach using existing devices. For instance, the LTSR-LMS algorithm needs approximately $0.137$ billion floating point operations (GFLOPS) to process each frame in the current example. Consider using a fast image registration algorithm such as~\cite{caner2006LMSadaptiveOF}, which has a computational complexity of $\kappa^2 M^2+g_{\max}^2 M^2+M^2$, with $\kappa$ being a small image window and $g_{\max}$ the maximum displacement amplitude. For typical values $\kappa=3$ and $g_{\max}=10$, the image registration cost is $0.228$ GFLOPS per frame. Now, for real-time performance (at 30 frames per second), the aggregate cost of image registration and SRR becomes $10.95$ GFLOPS/second, which is well within the capability of graphical processing units released almost a decade ago~\cite{surkov2010parallelGPUcalacity}.

%--------------------------------------------------------
%--------------------------------------------------------
%--------------------------------------------------------
\section{Conclusions}
\label{sec:concl}

This work proposed a new super-resolution reconstruction method aimed at an increased robustness to innovation outliers in real-time operation. An intuitive interpretation was proposed for the proximal-point cost function representation of the R-LMS gradient descent algorithm. A new regularization was then devised using statistical information on the innovations. This new  regularization allowed for faster convergence of the solution in the subspace related to the innovations, while preserving previously estimated details.
Two new algorithms were derived which present an increased robustness to outliers when compared to the R-LMS, with only a modest increase in the resulting computational cost. 
Computer simulations both with synthetic and real video sequences illustrated the effectiveness of the proposed methods.

\bibliographystyle{IEEEtran}
\bibliography{references}

\vspace{-3ex}
\begin{IEEEbiography}[{\includegraphics[width=1in,height=1.25in,clip,keepaspectratio]{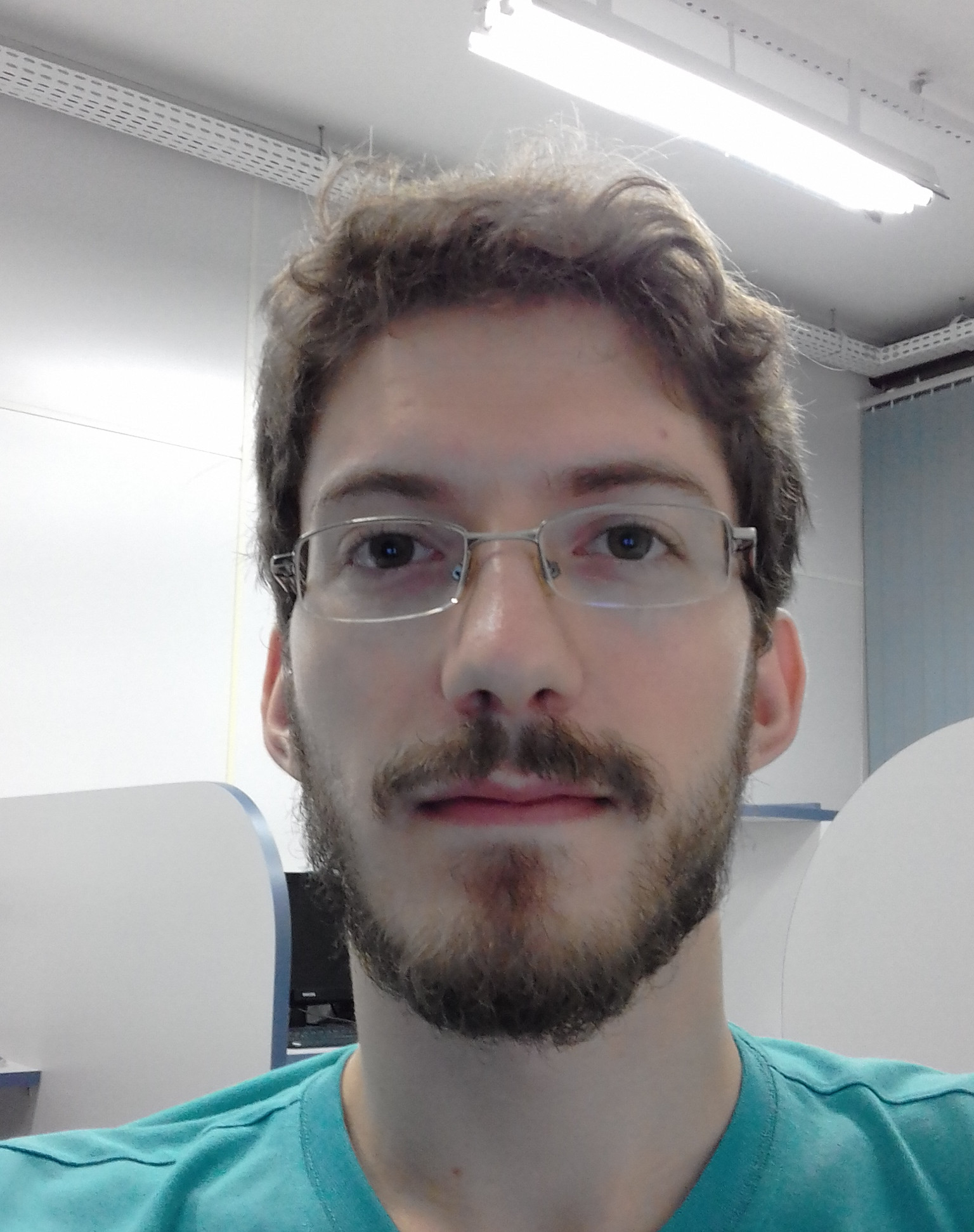}}]{Ricardo A. Borsoi}
% \begin{IEEEbiography}{Ricardo Augusto Borsoi}
received the BE degree on control and automation engineering from University of Caxias do Sul (UCS), Caxias do Sul, Brazil, in 2014, and received the MSc degree in electrical engineering from Federal University of Santa Catarina (UFSC), Florian\'opolis, Brazil, in 2016. He is currently working towards his doctoral degree. His current research interests include image processing, adaptive filtering and hyperspectral image analysis.
\end{IEEEbiography}

\vspace{-3ex}
\begin{IEEEbiography}[{\includegraphics[width=1in,height=1.25in,clip,keepaspectratio]{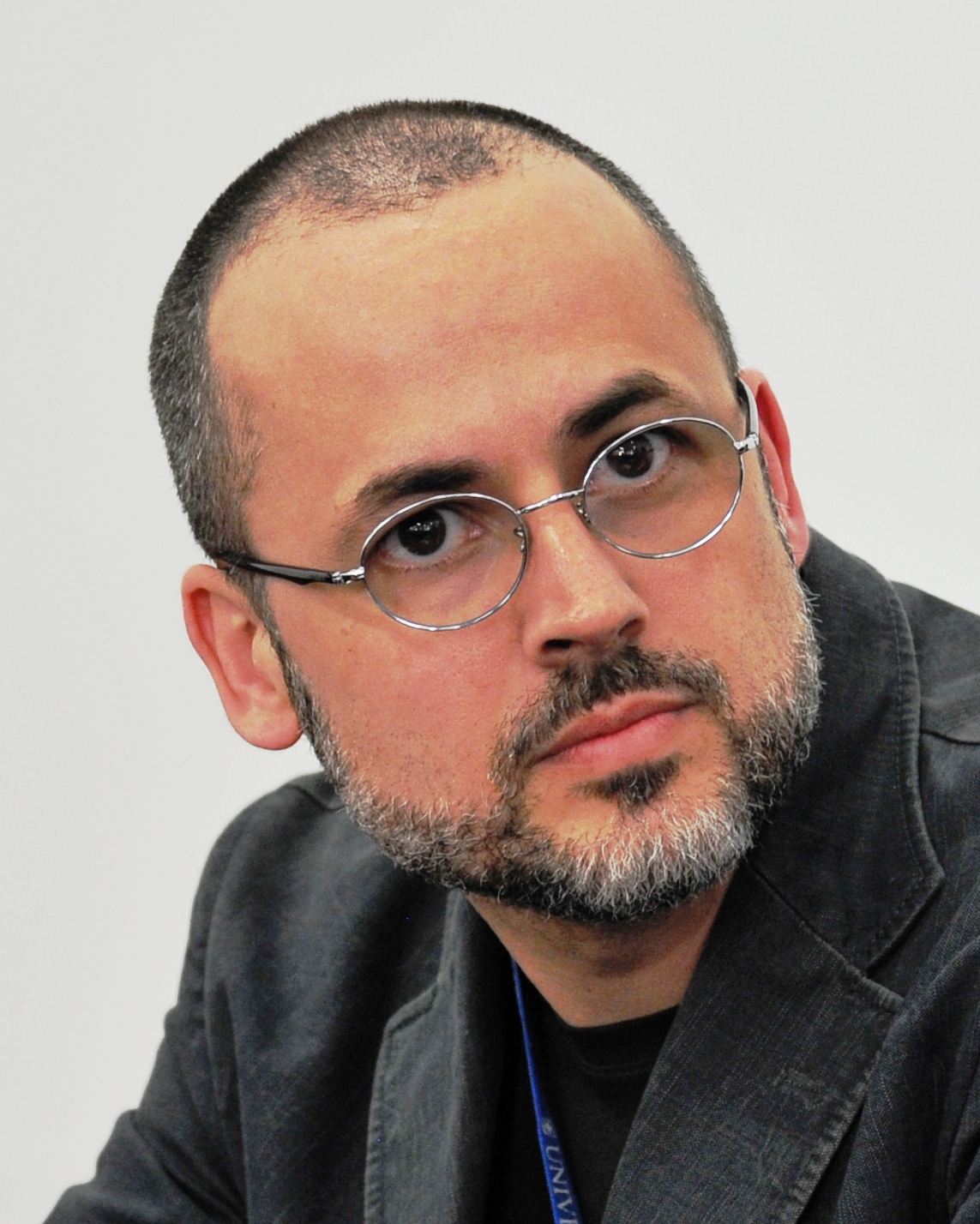}}]{Guilherme H. Costa}
% \begin{IEEEbiography}{Guilherme Holsbach Costa}
% (S05-M08) received the BEE degree in electrical engineering from Federal University of Rio Grande do Sul (UFRGS), Porto Alegre, Brazil, in 2001, and received the MSc and Doctoral degrees in electrical engineering from Federal University of Santa Catarina (UFSC), Florian\'opolis, Brazil, in 2003 and 2007, respectively. He was born in Pelotas, Brazil, in 1975. Currently, he is a full professor at University of Caxias do Sul (UCS), Caxias do Sul, Brazil. From 2015 to 2017 he worked as IEEE Signal Processing Society South Brazil (South Region) chapter chair. His current research interests include image super-resolution, computer vision and luminous efficiency. He is a member of the IEEE.
(S05-M08) received the BEE degree in electrical engineering from Federal University of Rio Grande do Sul (UFRGS), Porto Alegre, Brazil, in 2001, and received the MSc and Doctoral degrees in electrical engineering from Federal University of Santa Catarina (UFSC), Florian\'opolis, Brazil, in 2003 and 2007, respectively. Currently, he is a full professor at University of Caxias do Sul (UCS), Caxias do Sul, Brazil. From 2015 to 2017 he worked as IEEE Signal Processing Society South Brazil (South Region) chapter chair. His current research interests include image super-resolution, computer vision and luminous efficiency. 
\end{IEEEbiography}

\vspace{-3ex}
\begin{IEEEbiography}[{\includegraphics[width=1in,height=1.25in,clip,keepaspectratio]{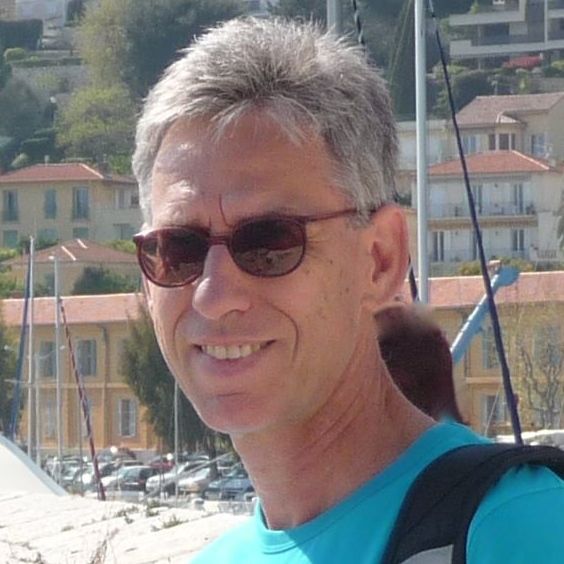}}]{Jos\'{e} C. M. Bermudez}  (M'85--SM'02) 
% \begin{IEEEbiography}{Jos\'{e} C. M. Bermudez}  
(M'85--SM'02) received the B.E.E. degree from the Federal University of Rio de Janeiro (UFRJ), Rio de Janeiro, Brazil, the M.Sc. degree in electrical engineering from COPPE/UFRJ, and the Ph.D. degree in electrical engineering from Concordia University, Montreal, Canada, in 1978, 1981, and 1985, respectively. He joined the Department of Electrical Engineering, Federal University of Santa Catarina (UFSC), Florianopolis, Brazil, in 1985, where he is a Professor. His research interests are in statistical signal processing, including adaptive filtering, image processing, hyperspectral image processing and machine learning. He served as an Associate Editor of the IEEE TRANSACTIONS ON SIGNAL PROCESSING  from 1994 to 1996 and from 1999 to 2001, and as an Associate Editor of the EURASIP Journal of Advances on Signal Processing from 2006 to 2010. He is a Senior Area Editor of the IEEE TRANSACTIONS ON SIGNAL PROCESSING and Associated Editor of the GRETSI journal Traitement du Signal. 
\end{IEEEbiography}

% received the B.E.E. degree from the Federal University of Rio de Janeiro (UFRJ), Rio de Janeiro, Brazil, the M.Sc. degree in electrical engineering from COPPE/UFRJ, and the Ph.D. degree in electrical engineering from Concordia University, Montreal, Canada, in 1978, 1981, and 1985, respectively. He joined the Department of Electrical Engineering, Federal University of Santa Catarina (UFSC), Florianopolis, Brazil, in 1985. He is currently a Professor of Electrical Engineering. His recent research interests are in statistical signal processing, including linear and nonlinear adaptive filtering, image processing, hyperspectral image processing and machine learning. He served as an Associate Editor of the IEEE TRANSACTIONS ON SIGNAL PROCESSING in the area of adaptive filtering from 1994 to 1996 and from 1999 to 2001. He also served as an Associate Editor of the EURASIP Journal of Advances on Signal Processing from 2006 to 2010. He is presently a Senior Area Editor of the IEEE TRANSACTIONS ON SIGNAL PROCESSING since 2015, and Associated Editor of the GRETSI journal Traitement du Signal.  

% He was a Visiting Researcher with the Concordia University (1992), with the University of California, Irvine (UCI) (1994),  with Institut National Polytechnique de Toulouse, France, in 2007, 2009, 2011 and 2016, with Universit\'e Nice Sophia-Antipolis, France, in 2010, 2014 (sabbatical year) 2014, 2016 and 2017.

\end{document}